\documentclass[preprint,12pt]{elsarticle}
\usepackage[T1]{fontenc}

\usepackage{multirow}
\usepackage{multicol}
\usepackage{lscape}
\usepackage{adjustbox}
\renewcommand{\arraystretch}{2}
\usepackage{makecell}
\usepackage[justification=raggedright]{caption}
\usepackage{geometry}
\usepackage{longtable}
\usepackage{lscape}
\usepackage{pdflscape}
\usepackage{afterpage}
\usepackage{rotating}
\usepackage{ltablex}
\usepackage{array,longtable}
\usepackage{pdflscape}
\usepackage{xltabular}
\usepackage{url}
\usepackage{graphicx}
\usepackage{float}
\usepackage{threeparttable}
\usepackage{subcaption}
\usepackage{algorithm}
\usepackage{algorithmic}
\newcommand{\highlight}[1]{\textcolor{black}{#1}}
\usepackage{pifont}
\newcommand{\cmark}{\ding{51}}  
\newcommand{\xmark}{\textcolor{red}{\ding{55}}}  
\makeatletter
\renewcommand{\paragraph}{\@startsection{paragraph}{4}{\z@}{1ex}{-1em}{\normalfont\normalsize}}
\makeatother
\usepackage{tikz}
\usetikzlibrary{shapes.geometric, arrows,positioning}
\usepackage{titlesec}

\titleformat{\subsubsection}[runin]{\normalfont\itshape}{\thesubsubsection}{0.3em}{}[]
\titlespacing*{\section}{0pt}{0.1\baselineskip}{0.1\baselineskip}
\titlespacing*{\subsection}{0pt}{0.1\baselineskip}{0.1\baselineskip}
\usepackage{etoolbox}
\makeatletter
\def\ifGm@preamble#1{\@firstofone}
\appto\restoregeometry{%
  \pdfpagewidth=\paperwidth
  \pdfpageheight=\paperheight}
\apptocmd\newgeometry{%
  \pdfpagewidth=\paperwidth
  \pdfpageheight=\paperheight}{}{}
\usepackage{ltablex}
\makeatother

\usepackage{caption} 
\usepackage[final]{pdfpages}
\usepackage{amssymb}
\usepackage{amsmath}

\journal{Neurocomputing}

\begin{document}

\begin{frontmatter}



\title{Vision Transformers on the Edge: A Comprehensive Survey of Model Compression and Acceleration Strategies}


\author{Shaibal Saha\corref{cor1}\fnref{label1}}
\ead{shaibalsaha@oakland.edu}
\author{Lanyu Xu\fnref{label1}}
\affiliation[label1]{organization={Department of Computer Science and Engineering, Oakland University},
            addressline={}, 
            city={Rochester Hills},
            postcode={48309}, 
            state={Michigan},
            country={USA}}
\cortext[cor1]{Corresponding author}



\begin{abstract}
In recent years, vision transformers (ViTs) have emerged as powerful and promising techniques for computer vision tasks such as image classification, object detection, and segmentation. Unlike convolutional neural networks (CNNs), which rely on hierarchical feature extraction, ViTs treat images as sequences of patches and leverage self-attention mechanisms. However, their high computational complexity and memory demands pose significant challenges for deployment on resource-constrained edge devices. To address these limitations, extensive research has focused on model compression techniques and hardware-aware acceleration strategies. Nonetheless, a comprehensive review that systematically categorizes these techniques and their trade-offs in accuracy, efficiency, and hardware adaptability for edge deployment remains lacking. This survey bridges this gap by providing a structured analysis of model compression techniques, software tools for inference on edge, and hardware acceleration strategies for ViTs. We discuss their impact on accuracy, efficiency, and hardware adaptability, highlighting key challenges and emerging research directions to advance ViT deployment on edge platforms, including graphics processing units (GPUs), \highlight{application-specific integrated circuit (ASICs)}, and field-programmable gate arrays (FPGAs). The goal is to inspire further research with a contemporary guide on optimizing ViTs for efficient deployment on edge devices.
\end{abstract}




\end{frontmatter}
\section{Introduction}
Deep learning architectures have evolved significantly in recent years, with transformers emerging as one of the most transformative breakthroughs. Transformers initially introduced for natural language processing (NLP) by Vaswani et al.~\cite{vaswani2017attention} in 2017 replaced recurrent models such as long short-term memory (LSTMs)~\cite{hochreiter1997long} and gated recurrent~\cite{chung2014empirical}, leveraging self-attention mechanisms to capture long-range dependencies in sequential data efficiently.

\begin{figure}[]
  \centering
  \includegraphics[scale=0.28]{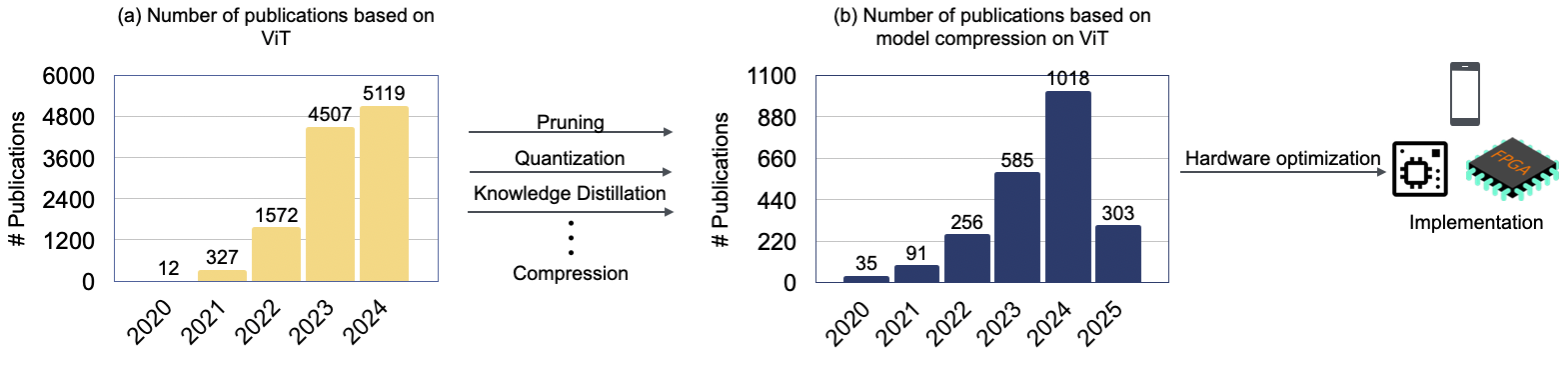}
  \caption{(a) The prevalence of transformer-based models in computer vision has led to a substantial increase in research publications. (b) Given their high computational complexity, model compression techniques are critical for reducing redundancy and improving efficiency. These advancements are essential for optimizing ViTs for hardware acceleration and real-world deployment on resource-constrained platforms~\cite{dimensions_data}.}
  \label{fig:current_trends}
\end{figure}

Following the tremendous success of transformers in NLP, researchers adapted their architecture for computer vision (CV), leading to the development of vision transformers (ViTs)~\cite{dosovitskiy2020image}. Unlike convolutional neural networks (CNNs)~\cite{krizhevsky2012imagenet}, which rely on hierarchical feature extraction, \highlight{ViT} model processes images as sequences of patch embeddings, enabling global context modeling via self-attention. 
Since the introduction of ViTs~\cite{dosovitskiy2020image}, research interest in ViT-based models has grown exponentially, as reflected in the increasing number of publications each year~(\highlight{see} Figure~\ref{fig:current_trends}a). This surge in publications highlights ViTs' dominance in CV tasks, driven by their state-of-the-art (SOTA) performance across various tasks, including image classification~\cite{dosovitskiy2020image,liu2021swin}, object detection~\cite{carion2020end,zhang2021vit}, and segmentation~\cite{hatamizadeh2022unetr,li2023lvit}.

While ViT-based models have demonstrated significant capabilities, the substantial size of these models presents major challenges for practical deployment. For instance, ViT-Huge includes over 632M parameters~\cite{dosovitskiy2020image} and \highlight{has recently been} extended to 22B parameters~\cite{dehghani2023scaling}, demanding extensive computational resources. These memory and processing requirements make direct deployment on resource-constrained edge devices impractical without optimization. To overcome these limitations, researchers have explored various model compression techniques to reduce computational overhead while preserving performance. As ViTs continue to gain prominence in CV tasks (Figure~\ref{fig:current_trends}a), there has been a parallel increase in research focused on optimizing their efficiency through compression techniques (Figure~\ref{fig:current_trends}b). Techniques such as pruning~\cite{liang2021pruning,bai2022dynamically}, quantization~\cite{lin2021fq,li2023repq}, and knowledge distillation (KD)~\cite{gou2021knowledge,lin2022knowledge} on ViT have gained traction, offering solutions to reduce model size, improve inference speed, and lower power consumption without significantly compromising accuracy.
However, compression techniques alone are often insufficient to meet real-time applications' stringent latency and throughput requirements on edge devices.

\highlight{Although model compression techniques reduce computational overhead, they are often insufficient to ensure real-time performance on edge devices due to hardware-specific constraints. Effective deployment of ViT requires tailored acceleration strategies that align with the architectural characteristics of each hardware platform. These strategies optimize execution by addressing ViTs' core computational bottlenecks—such as \highlight{self-attention's quadratic complexity and patch embedding inefficiencies}—through hardware-aware optimizations, efficient non-linear operations, and resource-efficient computation.} These approaches address the inherent on-device computational bottlenecks of ViTs, such as the quadratic complexity of self-attention and the inefficiencies in processing patch embeddings. Recent advancements in accelerating techniques, including the use of specialized or custom accelerators and \highlight{optimized libraries (e.g., TensorRT), have further expanded the possibilities for accelerating ViTs on edge devices (e.g., graphics processing units (GPUs), application-specific integrated circuits (ASICs), and field-programmable gate arrays (FPGAs))}. By bridging the gap between model-level optimizations and hardware-specific execution, software-hardware (SW-HW) co-design also plays a pivotal role in deploying ViTs on \highlight{edge} devices~\cite{fan2022m3vit,dong2023heatvit,wang2022via}. Compression techniques, optimized software tools, and hardware-aware acceleration strategies~\cite{nag2023vita, wang2022via,you2023vitcod} provide a pathway toward efficient, low-latency ViT inference, unlocking new possibilities \highlight{for different application domains such as autonomous driving, mobile vision applications, and medical imaging on edge devices.}

This survey provides a comprehensive review of both model compression and acceleration strategies tailored for ViT, \highlight{focusing} on their applicability to edge devices such as GPUs, central processing units (CPUs), FPGAs, and \highlight{ASICs}. We systematically categorize and analyze the latest advancements in pruning, quantization, KD, and hardware-aware optimizations. Furthermore, we explore emerging acceleration techniques, which aim to reduce latency and improve energy efficiency. By synthesizing insights from a broad range of studies, this survey serves as a valuable resource for researchers and practitioners seeking to deploy ViTs on edge devices.
\subsection{Motivations and Contribution}
ViTs have revolutionized CV tasks, achieving SOTA performance across tasks such as image classification, object detection, and segmentation. However, their high computational cost, memory footprint, and energy consumption present significant challenges for deployment on resource-constrained edge devices. \highlight{While various optimization techniques exist for ViTs, a systematic survey that collectively examines ViT-focused model compression, software tools, evaluation metrics, and hardware acceleration strategies for edge deployment remains limited. Existing surveys often treat these aspects independently, lacking a single resource that presents them together in a structured manner tailored to edge deployment.} Table~\ref{tab:current_survey} compares existing surveys on ViT model compression and acceleration techniques.
\renewcommand{\arraystretch}{1.2}
\begin{table}[]
\centering
\caption{Comparison of existing surveys on model compression and acceleration techniques for ViTs. \textbf{\cmark\cmark} indicates a comprehensive discussion, while \textbf{\cmark} denotes a limited discussion.}
\label{tab:current_survey}
\resizebox{0.9\columnwidth}{!}{%
\begin{tabular}{c|c|cccc}
\hline
\multirow{2}{*}{\textbf{Survey}} &
  \multirow{2}{*}{\textbf{Year}} &
  \multicolumn{4}{c}{\textbf{Scope}} \\ \cline{3-6} 
 &
   &
  \multicolumn{1}{c|}{\textbf{Model Compression}} &
  \multicolumn{1}{c|}{\textbf{Software Tools}} &
  \multicolumn{1}{c|}{\textbf{Evaluation Metrics}} &
  \textbf{Hardware Accelerators} \\ \hline
\cite{chen2022lightweight} &
  2022 &
  \multicolumn{1}{c|}{\cmark} &
  \multicolumn{1}{c|}{\xmark} &
  \multicolumn{1}{c|}{\xmark} &
  \xmark \\ \hline
\cite{huang2022hardware} &
  2022 &
  \multicolumn{1}{c|}{\cmark\cmark} &
  \multicolumn{1}{c|}{\xmark} &
  \multicolumn{1}{c|}{\xmark} &
  \cmark \\ \hline
\cite{papa2024survey} &
  2024 &
  \multicolumn{1}{c|}{\cmark\cmark} &
  \multicolumn{1}{c|}{\xmark} &
  \multicolumn{1}{c|}{\cmark} &
  \xmark \\ \hline
\cite{du2024model} &
  2024 &
  \multicolumn{1}{c|}{\cmark\cmark} &
  \multicolumn{1}{c|}{\xmark} &
  \multicolumn{1}{c|}{\xmark} &
  \cmark \\ \hline
Our Survey &
   2025 &
  \multicolumn{1}{c|}{\cmark\cmark} &
  \multicolumn{1}{c|}{\cmark\cmark} &
  \multicolumn{1}{c|}{\cmark\cmark} &
  \cmark\cmark \\ \hline
\end{tabular}%
}
\end{table}

This survey addresses this gap by systematically analyzing model compression techniques (pruning, quantization, KD) and hardware-aware acceleration strategies (efficient attention mechanisms, SW-HW co-design, FPGA optimizations, etc.). By analyzing insights from a broad range of studies, this survey serves as a valuable resource for researchers seeking to deploy ViTs on edge devices. The main contributions of our survey are as follows:
\begin{enumerate}
    \item We systematically categorize and analyze \highlight{model compression techniques such as} pruning, quantization, and KD to optimize ViTs for deploying on edge devices.
    \item We investigate current \highlight{software} tools for efficient inference and hardware-aware accelerating techniques to enhance inference efficiency across edge platforms like GPUs, FPGAs, and ASICs \highlight{for ViT models}.
    \item By \highlight{p}roviding a structured roadmap for integrating compression and acceleration techniques, we offer comparative analyses and identify challenges and future research directions for real-time, low-power ViT applications.
\end{enumerate}
\subsection{Literature Collection and Organizations}\label{question}
\begin{figure}[htb]
  \centering
  \includegraphics[scale=0.50]{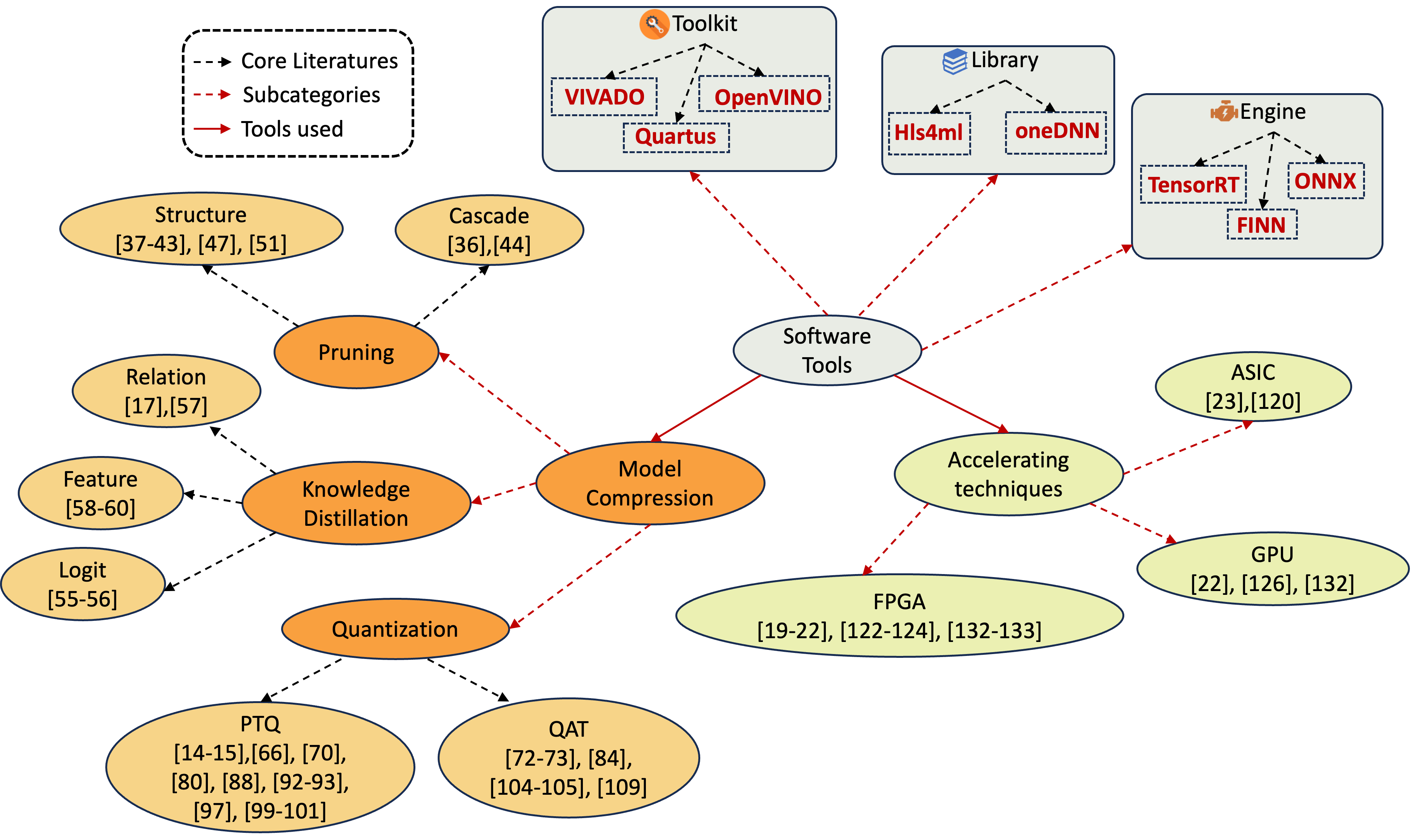}
  \caption{\highlight{Core components and tools are covered in this survey. This survey focuses on three main parts—model compression techniques (pruning, KD, and quantization), accelerating techniques (using FPGA, GPU, and ASIC), and associated software tools (toolkits, libraries, and inference engines)—to provide a comprehensive understanding of efficient ViT deployment.}}
  \label{fig:key_concepts}
\end{figure}

Our literature search was conducted across major academic databases, including \textbf{Google Scholar, IEEE Xplore, arXiv, and the ACM Digital Library}, to ensure comprehensive coverage of relevant research. We utilized targeted search queries with keywords such as \textbf{vision transformer, acceleration techniques, edge devices, software-hardware co-design, pruning, quantization, and knowledge distillation} to identify studies relevant to this survey. \highlight{A total of 170 papers were collected, with works published up to January 2025 considered for inclusion. Figure~\ref{fig:key_concepts} presents a structured overview of the selected literature, categorizing each referenced work according to the core research themes and enabling tools addressed in this survey. Papers were included based on how well they aligned with our key research questions. A paper was selected if it met one or more of the following criteria:}
\begin{enumerate}
    \item Does the paper propose a compression technique for improving ViT efficiency in terms of computational cost or energy consumption?
    \item Does the paper provide a comparative analysis of ViT acceleration techniques or benchmark performance across different hardware platforms?
    \item Does the paper explore the integration of ViTs with hardware-aware optimizations, including \highlight{SW-HW} co-design strategies?
\end{enumerate}
The remainder of the survey is organized as follows. Section~\ref{model_com} presents an in-depth discussion on model compression techniques, including pruning, quantization, and KD, which enhance ViT efficiency while preserving performance across various \highlight{CV} tasks. Following this, Section~\ref{tools_for_edge} explores an overview of current software tools, optimization frameworks, and evaluation metrics designed for efficient edge inference across different edge devices. \highlight{Section~\ref{acce_tech} further explores hardware-aware acceleration techniques, focusing on optimizations for non-linear operations (e.g., softmax, GELU, and LayerNorm) and recent SOTA SW-HW co-design strategies, along with a comparative analysis of their performance.} Furthermore, Section~\ref{cha_fu} discusses key challenges and future research directions, identifying multiple avenues for advancing ViT acceleration and deployment on edge devices. Finally, we conclude this
paper in Section~\ref{conclusion}.
\section{Model Compression}\label{model_com}
Model compression is a key technique for deploying a model on edge devices with limited computational power and memory while maintaining \highlight{accuracy. It enables lowering inference latency as well as reducing memory and energy consumption.} However, model compression on ViT still needs extensive exploration due to \highlight{its} complex architecture and high resource usage tendencies. In this section, we will discuss prominent compression techniques for ViT models: pruning (Section~\ref{prun}), KD (Section~\ref{know}), and quantization (Section~\ref{quan}). 

\subsection{Pruning}\label{prun}
Pruning is used \highlight{to reduce both memory and bandwidth}. Most of the initial pruning techniques \highlight{based} on biased weight decay~\cite{hanson1988comparing},
second-order derivatives~\cite{hassibi1992second}, and channels~\cite{molchanov2016pruning}. Early days pruning techniques reduce the number of connections based on the hessian of the loss function~\cite{cheng2017survey,wu2021evolutionary}. In general, pruning removes redundant parameters that do not significantly contribute to the accuracy of results. The pruned model has fewer edges/connections than the original model. Most early pruning techniques are like brute force pruning, where one needs to manually check which weights do not cause any accuracy loss. Pruning techniques in deep learning became prominent post-2000 as neural networks (NNs) grew in size and complexity. The following subsections will discuss different pruning types and recent techniques applied to ViT-based models.

\subsubsection{Types}\hfill\\ 
In recent studies, various pruning techniques have been utilized to optimize ViT models, categorizing them into different methods based on their approach and application timing. For example, unstructured pruning targets individual weights for removal, whereas structured pruning removes components at a broader scale, like layers or channels. On the one hand, static pruning is predetermined and fixed, which is ideal for environments with known constraints. On the other hand, dynamic pruning offers real-time adaptability, potentially enhancing model efficiency without sacrificing accuracy. Another pruning method, cascade pruning, is highlighted as a hybrid approach, integrating the iterative adaptability of dynamic pruning with the structured approach of static pruning. The following subsections will be a detailed discussion of different pruning types.\\

\noindent \textbf{Unstructured vs Structured Pruning} Unstructured pruning removes individual weights or parameters from the network based on certain criteria. Unstructured pruning can result in highly sparse networks. However, it often does not lead to computational efficiency as the sparsity is not aligned with the memory access patterns or computational primitives of hardware accelerators. However, structured pruning applies to the specific components of the network, especially in layers, neurons, or channels. Structured pruning leads to more hardware-friendly sparsity patterns but often at the cost of higher accuracy loss. Cai et al.~\cite{10.1145/3543622.3573044} proposed a two-stage coarse-grained/fine-grained structured pruning method based on top-K sparsification, reducing 60\% overall computation in the embedded NNs. In a recent survey~\cite{he2023structured}, He et al. discussed a range of SOTA structured pruning techniques, covering topics such as filter ranking methods, regularization methods, dynamic execution, neural architecture search (NAS), the lottery ticket hypothesis, and the applications of pruning.\\

\begin{figure*}[h]
  \centering
  \includegraphics[scale=0.30]{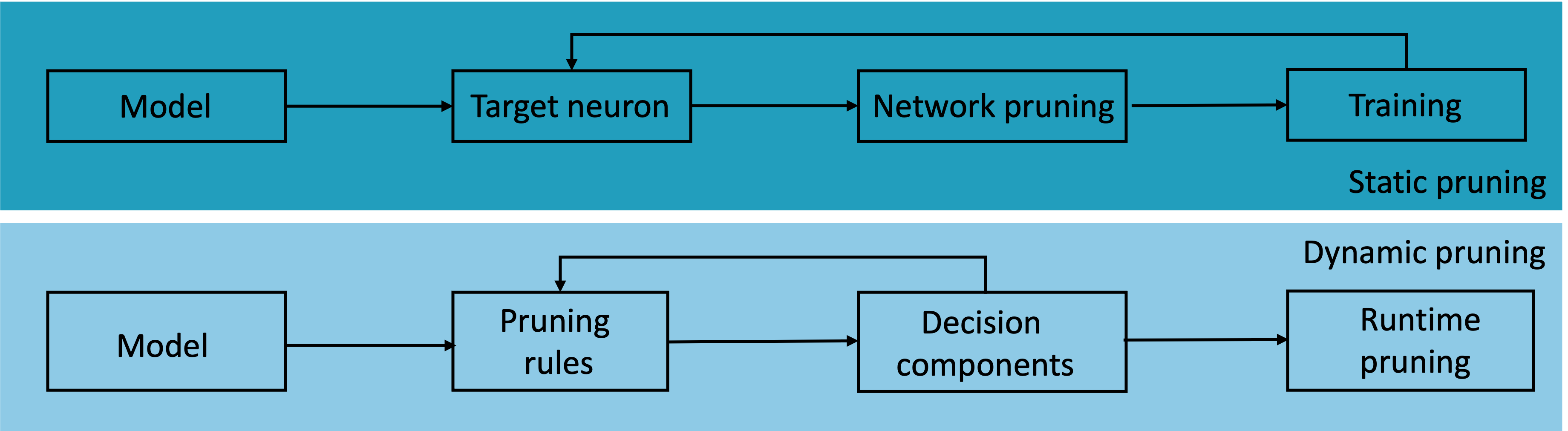}
  \caption{Static Pruning vs Dynamic Pruning}
  \label{fig:staticvsdynamic}
\end{figure*}

\noindent \textbf{Static vs. Dynamic Pruning} Figure~\ref{fig:staticvsdynamic} demonstrates the workflow of the static and dynamic pruning. Static pruning works at the offline inference level, whereas dynamic pruning performs at the runtime level. Moreover, static pruning applies during training, where a fixed portion of a NN's components, such as neurons, channels, or weights, is removed or pruned. In static pruning, the decision on which components to prune is typically made before the training begins, and the pruning schedule remains constant throughout training. Static pruning is helpful in scenarios where the hardware constraints are well-defined. One of the most used static pruning techniques is Magnitude-based pruning~\cite{han2015learning}. In magnitude-based pruning, given a pruning rate \textbf{r}, weights whose absolute value is among the smallest \textbf{r\%} are pruned. In other words, for a weight matrix \textbf{W} of a layer, weights \textbf{w} in \textbf{W} are pruned if \textbf{\textbar w\textbar $\leq$ threshold}, where the \textbf{threshold} is determined such that the proportion of \textbf{\textbar w\textbar $\leq$ threshold} is \textbf{r}.
In contrast, dynamic pruning applies during runtime based on specific criteria, such as the importance of neurons or weights. One of the significant drawbacks of static pruning is that it relies on a fixed pruning schedule and rate, which is determined before training begins. That means that static pruning does not adjust to the network's learning progress or the changing importance of neurons or weights during training. Dynamic pruning evaluates and adjusts the pruning criteria during training based on real-time importance assessments to overcome the limitation of static pruning. Most of the recent pruning techniques use dynamic pruning techniques~\cite{song2022cp,bai2022dynamically,kong2022spvit} to get the accuracy without loss of any information.
\newline\hfill\break
\noindent \textbf{Cascade Pruning} Cascade pruning combines the iterative nature of dynamic pruning with predefined aspects resembling static pruning. Cascade pruning operates through multiple sequential iterations, also known as stages. Each stage selects a predefined portion of the network's components, such as neurons, channels, or weights, for pruning. The criteria for choosing which components to prune can vary between iterations. The ability to adjust pruning criteria between iterations makes cascade pruning adaptable to evolving training data, tasks, or hardware constraints.
\renewcommand{\arraystretch}{1.2}
\begin{table}[H]
\centering
\caption{\highlight{Categorization of ViT pruning methods across key perspectives. Many methods span multiple categories.}}
\label{tab:pruning}
\resizebox{0.8\columnwidth}{!}{%
\begin{tabular}{c|c|c|c|c}
\hline
\textbf{Method} & \textbf{Importance-Based} & \textbf{Token Pruning} & \textbf{Structured} & \textbf{Cascade} \\ \hline
VTP~\cite{zhu2021vision}            & \checkmark & \xmark     & \xmark     & \xmark     \\ \hline
Patch Slimming~\cite{tang2022patch} & \checkmark & \checkmark & \xmark     & Top-down   \\ \hline
WDPruning~\cite{yu2022width}        & \checkmark & \xmark     & \checkmark & \xmark     \\ \hline
SP-ViT~\cite{kong2022spvit}         & \xmark     & \checkmark & \checkmark & \xmark     \\ \hline
UVC~\cite{yu2022unified}            & \checkmark & \xmark     & \checkmark & \xmark     \\ \hline
CP-ViT~\cite{song2022cp}            & \checkmark & \checkmark & \checkmark & \checkmark \\ \hline
NViT~\cite{yang2023global}          & \checkmark & \xmark     & \checkmark & \xmark     \\ \hline
S$^2$ViT~\cite{chen2021chasing}     & \checkmark & \checkmark & \checkmark & \xmark     \\ \hline
VTC-LFC~\cite{wang2022vtc}          & \checkmark & \xmark     & \checkmark & \checkmark \\ \hline
\end{tabular}%
}
\end{table}
\subsubsection{Pruning Techniques for Vision Transformer}\label{pruning_in_vit}
\hfill\\
Pruning methods for ViT-based models remain an underexplored area, with only a handful of studies in recent years. \highlight{Current ViT pruning methods can be mainly divided into importance-based, token pruning, structure pruning, cascading pruning (combining multiple pruning techniques for each module), and miscellaneous methods. Note that many of these pruning approaches overlap across categories. For instance, patch slimming~\cite{tang2022patch} and CP-ViT~\cite{song2022cp} apply both importance-based and structured token pruning, while CP-ViT~\cite{song2022cp} and VTC-LFC~\cite{wang2022vtc} also follow a cascade strategy. Table~\ref{tab:pruning} summarizes the categorization across different pruning approaches.} This section provides a brief overview of the current SOTA pruning techniques for ViTs.\\ 

\noindent \textbf{Structure Pruning} Recently, many studies on ViT pruning have embraced structure pruning techniques to optimize model efficiency. As discussed in the previous subsections, importance-based pruning and token pruning can also be considered as structured pruning. Additionally, Yu et al.~\cite{yu2022unified} proposed a structure pruning in a ViT named UVC, where they pruned the number of heads and dimensions inside each layer. Experiments of this paper were conducted in various ViT models (e.g., DeiT-Tiny and T2T-ViT-14) on ImageNet-1k~\cite{5206848} datasets. DeiT-Tiny~\cite{touvron2021training} cut down to 50\% of the original floating-point operations per second (FLOPS) while not dropping accuracy much in this study. Another study~\cite{zheng2022savit} proposed structure pruning on multi-head self-attention (MSA)~\cite{vaswani2017attention} and feedforward neural network (FFN)~\cite{vaswani2017attention} by removing unnecessary parameter groups. Other studies on structure pruning named NViT~\cite{yang2023global} proposed hessian-based \highlight{iterative} structure pruning criteria comparable across all layers and structures. Moreover, it incorporated latency-aware regularization techniques to reduce latency directly. Another study on the structure pruning in ViT called S\textsuperscript{2}ViT~\cite{chen2021chasing} removed submodules like self-attention heads by manipulating the weight, activations \& gradients. Additionally, current structure pruning methods on ViTs utilize either importance-based or learnable token selector pruning.\\

\begin{figure}[htb]
  \centering
  \includegraphics[scale=0.4]{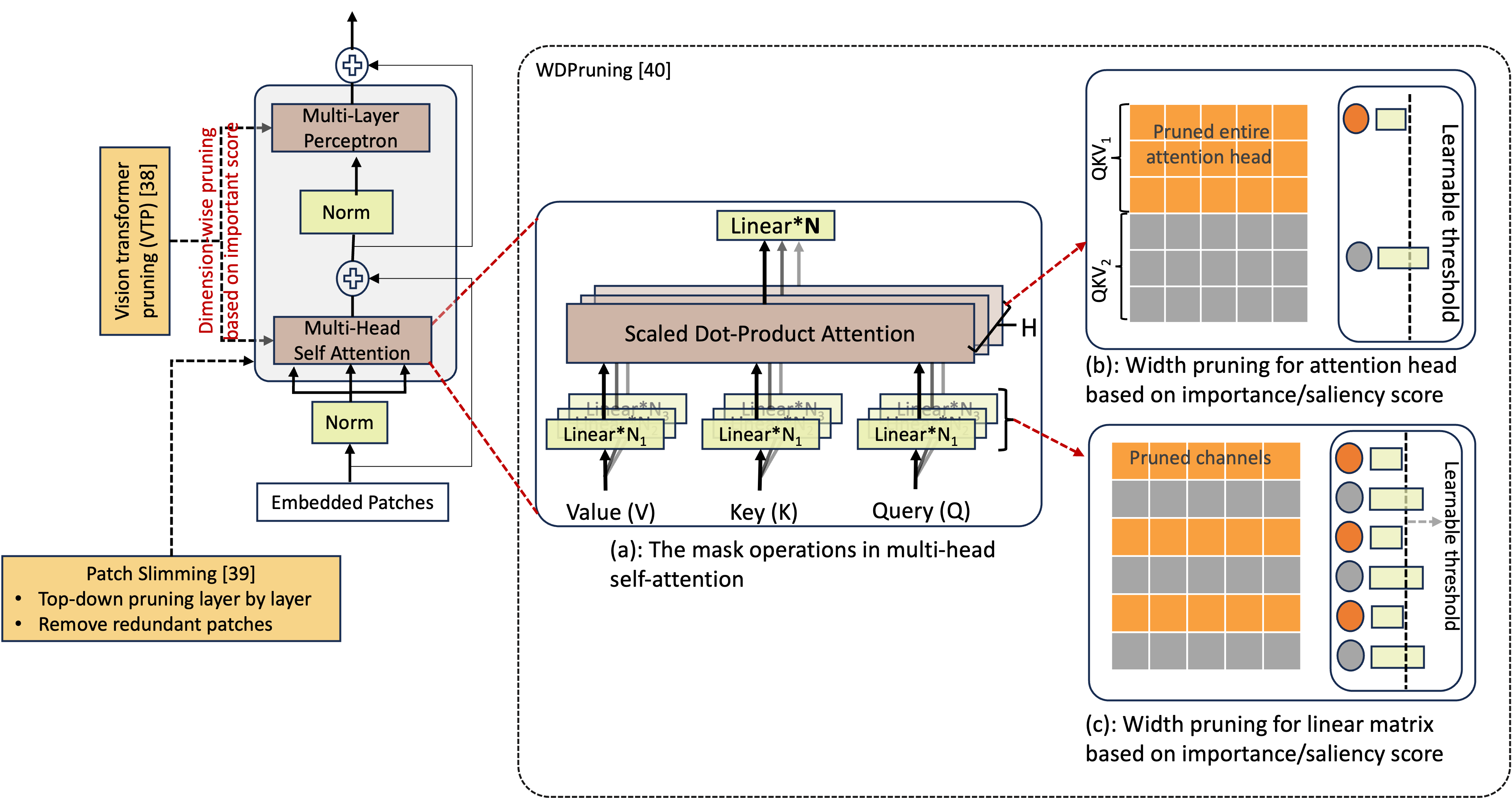}
  \caption{\highlight{The overview of importance-based pruning techniques for ViTs. VTP~\cite{zhu2021vision} prunes dimensionality in the multi-head self-attention (MSA) and multi-layer perception (MLP) modules using important scores. WDPruning~\cite{yu2022width} applies a binary mask ($N$) to the MSA module (a), followed by pruning of (b) attention heads and (c) linear projection channels. Patch Slimming~\cite{tang2022patch} removes redundant patches in a top-down, layer-wise manner.}}
  \label{fig:important_based}
\end{figure}
\noindent \textbf{\textit{\highlight{Importance-based Pruning}}} \highlight{One of the significant challenges in importance-based pruning for ViTs lies in determining effective thresholds to identify which components—such as channels, attention heads, blocks, or input patches—should be pruned. Different pruning strategies address this by designing tailored scoring criteria and thresholding mechanisms, often customized based on the targeted layer type or pruning granularity. Since a significant portion of ViT’s computational burden stems from the MSA and multi-layer perceptron (MLP) modules~\cite{vaswani2017attention}, targeted pruning of these modules can yield substantial computational savings. Figure~\ref{fig:important_based} summarizes the current pruning techniques on ViT based on different important score measurements. Zhu et al.~\cite{zhu2021vision} introduced vision transformer pruning (VTP), one of the first dedicated pruning methods for ViTs that removes dimensions with lower importance scores, achieving a high pruning ratio without sacrificing accuracy.} VTP operates in three key steps: 1) L1 sparse regularization is applied during training to identify less significant channels, 2) channel pruning eliminates redundant computations \highlight{based on thresholds derived from a predefined pruning rate}, and (3) finetuning. \highlight{However, VTP ignores network depth, which is an important dimension for pruning the model. To address this issue, another study~\cite{yu2022width} (see Figure~\ref{fig:important_based}(a--c)) explored width and depth pruning on the MSA module in the transformer models. Block a in Figure~\ref{fig:important_based} illustrates the application of binary masking $N$ within the MSA module. Additionally, Blocks b and c in Figure~\ref{fig:important_based} depict width pruning applied to the linear projection matrices and attention head pruning, respectively.} Moreover, leveraging the learnable saliency score approach enables non-uniform sparsity across layers and facilitates progressive block removal, allowing the model to be pruned sequentially during a single training epoch. Furthermore, Tang et al.~\cite{tang2022patch} \highlight{identified that many input patches exhibit redundancy across transformer layers. Based on this insight, they proposed a layer-by-layer patch-slimming approach that reduces redundant patches in a top-down manner, offering a more adaptive alternative to conventional structured pruning techniques.} The authors calculated their importance scores for the final classification feature to identify unimportant patches. The proposed method started by identifying important patches in the last transformer layer \highlight{and recursively utilized them back to patches in earlier layers.}\\

\begin{figure}[]
  \centering
  \includegraphics[scale=0.5]{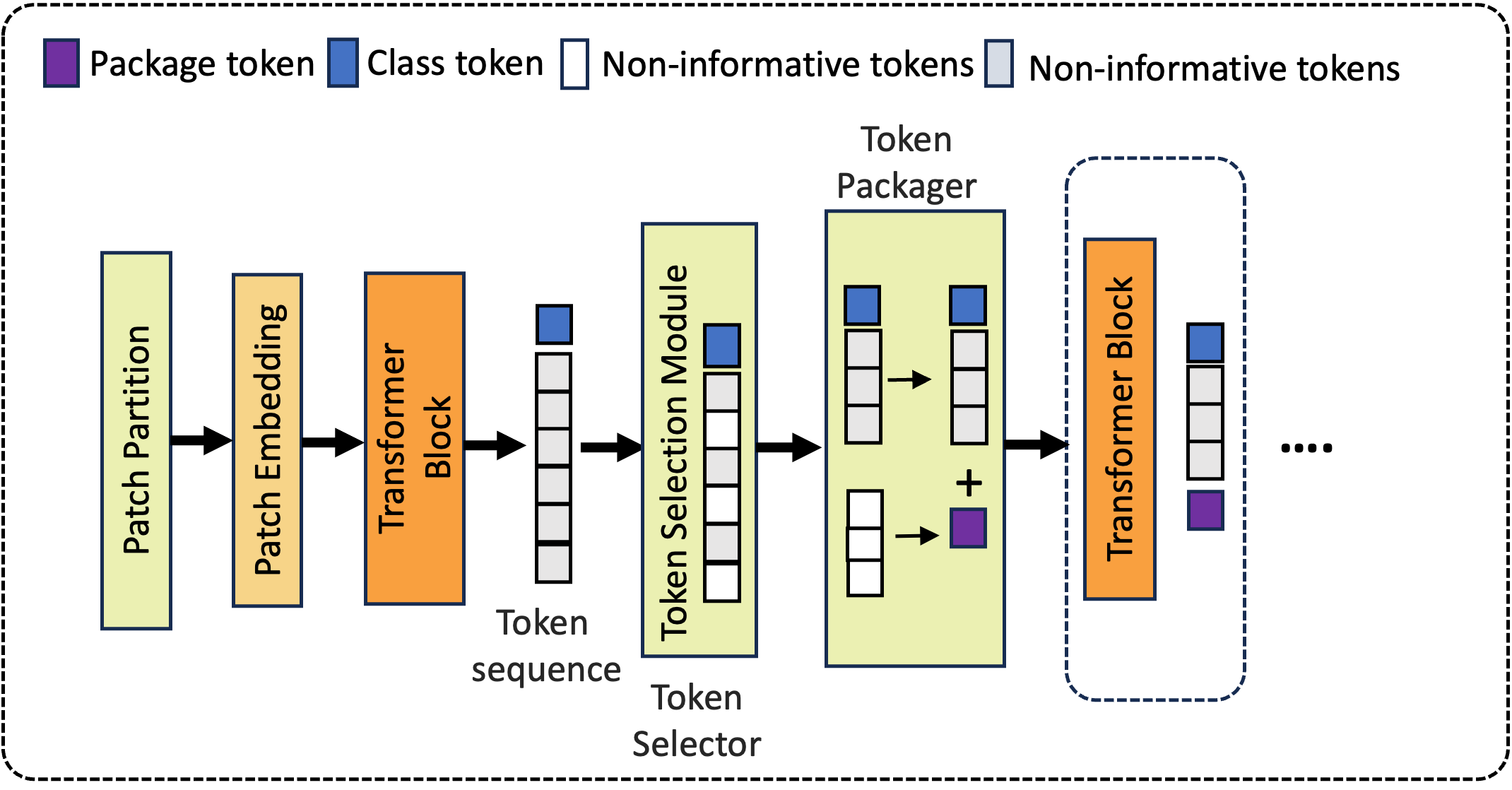}
  \caption{\highlight{The overview of token pruning based on SP-ViT~\cite{kong2022spvit}.}}
  \label{fig:token}
\end{figure} 
\noindent \textbf{\textit{\highlight{Token Pruning}}} Kong et al.~\cite{kong2022spvit} introduced a latency-aware soft token pruning framework, SP-ViT. \highlight{Figure~\ref{fig:token} illustrates the workflow of SP-ViT token pruning on ViT models.} The framework was implemented on vanilla transformers such as data-efficient image transformers (DeiT)~\cite{touvron2021training} and swin transformers~\cite{liu2021swin}. The authors proposed a dynamic attention-based multi-head token selector for adaptive instance-wise token selection. Subsequently, they incorporated a soft pruning method that consolidated less informative tokens into a package token instead of entirely discarding them, identified by the selector module. \highlight{The authors evaluated their proposed method on ImageNet-1k~\cite{5206848} dataset using Swin-S, Swin-S/T~\cite{liu2021swin}, PiT-Xs/S~\cite{zheng2023pit} models. SP-ViT was subsequently adopted by FPGA-based accelerating frameworks such as HeatViT~\cite{dong2023heatvit}. While SP-ViT~\cite{kong2022spvit} was initially developed and evaluated for image classification, recent studies have extended token pruning techniques to more complex tasks such as object detection~\cite{sah2024token} and semantic segmentation~\cite{tang2023dynamic} using lightweight, background-aware ViTs.} \\ 
\renewcommand{\arraystretch}{1.2}
\begin{table}[htb]
\centering
\caption{\highlight{Top-1 (\%) accuracy comparison of pruning techniques across different ViT models. '\textbf{$\downarrow$}' denotes accuracy loss from the baseline ViT model, and '$\uparrow$' denotes the accuracy gain after applying the pruning technique. All results are evaluated on the ImageNet-1k~\cite{5206848} dataset.}}
\label{tab:pruning_accuracy}
\resizebox{0.8\columnwidth}{!}{%
\begin{tabular}{c|c|ccccc}
\hline
\multirow{2}{*}{\textbf{Algorithm}} &
  \multirow{2}{*}{\textbf{Method}} &
  \multicolumn{5}{c}{\textbf{Top-1 Accuracy (\%) on ViT Models}} \\ \cline{3-7} 
 &
   &
  \multicolumn{1}{c|}{\textbf{DeiT-T}} &
  \multicolumn{1}{c|}{\textbf{DeiT-S}} &
  \multicolumn{1}{c|}{\textbf{DeiT-B}} &
  \multicolumn{1}{c|}{\textbf{Swin-S}} &
  \textbf{T2T-ViT-14} \\ \hline
\textbf{Baseline} &
  - &
  \multicolumn{1}{c|}{72.2} &
  \multicolumn{1}{c|}{79.8} &
  \multicolumn{1}{c|}{81.8} &
  \multicolumn{1}{c|}{83.2} &
  81.5 \\ \hline
VTP~\cite{zhu2021vision} &
  Dimension-based &
  \multicolumn{1}{c|}{-} &
  \multicolumn{1}{c|}{-} &
  \multicolumn{1}{c|}{\begin{tabular}[c]{@{}c@{}}80.7\\ ($\downarrow$ 1.34\%)\end{tabular}} &
  \multicolumn{1}{c|}{-} &
  - \\ \hline
WDPruning~\cite{yu2022width} &
  Width-depth pruning &
  \multicolumn{1}{c|}{\begin{tabular}[c]{@{}c@{}}70.34\\ ($\downarrow$ 2.58\%)\end{tabular}} &
  \multicolumn{1}{c|}{\begin{tabular}[c]{@{}c@{}}78.3\\ ($\downarrow$ 1.88\%)\end{tabular}} &
  \multicolumn{1}{c|}{\begin{tabular}[c]{@{}c@{}}80.8\\ ($\downarrow$ 1.22\%)\end{tabular}} &
  \multicolumn{1}{c|}{\begin{tabular}[c]{@{}c@{}}81.8\\ ($\downarrow$ 1.68\%)\end{tabular}} &
  - \\ \hline
MD~\cite{hou2022multi} &
  \begin{tabular}[c]{@{}c@{}}Dependency based pruning \\ by Gaussian-search\end{tabular} &
  \multicolumn{1}{c|}{-} &
  \multicolumn{1}{c|}{\begin{tabular}[c]{@{}c@{}}79.9\\ ($\uparrow$ 0.13\%)\end{tabular}} &
  \multicolumn{1}{c|}{\begin{tabular}[c]{@{}c@{}}82.3\\ ($\downarrow$ 0.61\%)\end{tabular}} &
  \multicolumn{1}{c|}{-} &
  \textbf{\begin{tabular}[c]{@{}c@{}}81.7\\ ($\uparrow$ 0.24\%)\end{tabular}} \\ \hline
SPViT~\cite{kong2022spvit} &
  \begin{tabular}[c]{@{}c@{}}Pruning based on\\ token score\end{tabular} &
  \multicolumn{1}{c|}{-} &
  \multicolumn{1}{c|}{-} &
  \multicolumn{1}{c|}{-} &
  \multicolumn{1}{c|}{\begin{tabular}[c]{@{}c@{}}82.7\\ ($\downarrow$ 0.60\%)\end{tabular}} &
  - \\ \hline
PS-ViT~\cite{tang2022patch} &
  \begin{tabular}[c]{@{}c@{}}Layer-wise top \\ down pruning\end{tabular} &
  \multicolumn{1}{c|}{\begin{tabular}[c]{@{}c@{}}72.0 \\ ($\downarrow$ 0.28\%)\end{tabular}} &
  \multicolumn{1}{c|}{\begin{tabular}[c]{@{}c@{}}79.4 \\ ($\downarrow$ 0.50\%)\end{tabular}} &
  \multicolumn{1}{c|}{\begin{tabular}[c]{@{}c@{}}81.5\\ ($\downarrow$ 0.30\%)\end{tabular}} &
  \multicolumn{1}{c|}{-} &
  \begin{tabular}[c]{@{}c@{}}81.1\\ ($\downarrow$ 0.49\%)\end{tabular} \\ \hline
UVC~\cite{yu2022unified} &
  Head/dimension pruning &
  \multicolumn{1}{c|}{\begin{tabular}[c]{@{}c@{}}70.6\\ ($\downarrow$ 2.08\%)\end{tabular}} &
  \multicolumn{1}{c|}{\begin{tabular}[c]{@{}c@{}}78.8\\ ($\downarrow$ 1.25\%)\end{tabular}} &
  \multicolumn{1}{c|}{\begin{tabular}[c]{@{}c@{}}80.6 \\ ($\downarrow$ 1.47\%)\end{tabular}} &
  \multicolumn{1}{c|}{-} &
  \begin{tabular}[c]{@{}c@{}}78.9\\ ($\downarrow$ 3.19\%)\end{tabular} \\ \hline
NViT~\cite{yang2023global} &
  \begin{tabular}[c]{@{}c@{}}Latency-aware, Hessian-\\ based pruning\end{tabular} &
  \multicolumn{1}{c|}{\textbf{\begin{tabular}[c]{@{}c@{}}76.2\\ ($\uparrow$ 5.55\%)\end{tabular}}} &
  \multicolumn{1}{c|}{\textbf{\begin{tabular}[c]{@{}c@{}}82.2\\ ($\uparrow$ 3.01\%)\end{tabular}}} &
  \multicolumn{1}{c|}{\textbf{\begin{tabular}[c]{@{}c@{}}83.3\\ ($\uparrow$ 1.83\%)\end{tabular}}} &
  \multicolumn{1}{c|}{\textbf{\begin{tabular}[c]{@{}c@{}}83.0\\ ($\downarrow$ 0.24\%)\end{tabular}}} &
  - \\ \hline
SAViT~\cite{zheng2022savit} &
  \begin{tabular}[c]{@{}c@{}}Structural pruning on \\ MSA attention \& FFN\end{tabular} &
  \multicolumn{1}{c|}{\begin{tabular}[c]{@{}c@{}}70.7\\ ($\downarrow$ 2.08\%)\end{tabular}} &
  \multicolumn{1}{c|}{\begin{tabular}[c]{@{}c@{}}80.1\\ ($\uparrow$ 0.37\%)\end{tabular}} &
  \multicolumn{1}{c|}{\begin{tabular}[c]{@{}c@{}}81.7\\ ($\uparrow$ 0.12\%)\end{tabular}} &
  \multicolumn{1}{c|}{-} &
  - \\ \hline
S$^2$ViTE~\cite{chen2021chasing} &
  \begin{tabular}[c]{@{}c@{}}Removing sub-modules \\ by manipulating weight,\\ activation,\& gradient\end{tabular} &
  \multicolumn{1}{c|}{\begin{tabular}[c]{@{}c@{}}70.1\\ ($\downarrow$ 2.91\%)\end{tabular}} &
  \multicolumn{1}{c|}{\begin{tabular}[c]{@{}c@{}}79.2 \\ ($\downarrow$ 0.75\%)\end{tabular}} &
  \multicolumn{1}{c|}{\begin{tabular}[c]{@{}c@{}}82.2\\ ($\uparrow$ 0.49\%)\end{tabular}} &
  \multicolumn{1}{c|}{-} &
  - \\ \hline
VTC-LFC~\cite{wang2022vtc} &
  \begin{tabular}[c]{@{}c@{}}Token \& channel pruning \\ using a hyperparameter\end{tabular} &
  \multicolumn{1}{c|}{\begin{tabular}[c]{@{}c@{}}71.0\\ ($\downarrow$ 1.66\%)\end{tabular}} &
  \multicolumn{1}{c|}{\begin{tabular}[c]{@{}c@{}}79.6\\ ($\downarrow$ 0.25\%)\end{tabular}} &
  \multicolumn{1}{c|}{\begin{tabular}[c]{@{}c@{}}81.6\\ ($\downarrow$ 0.24\%)\end{tabular}} &
  \multicolumn{1}{c|}{-} &
  - \\ \hline
CP-ViT~\cite{song2022cp} &
  \begin{tabular}[c]{@{}c@{}}Prune PH-regions in\\ MSA \& FFN \\ dynamically\end{tabular} &
  \multicolumn{1}{c|}{\begin{tabular}[c]{@{}c@{}}71.2\\ ($\downarrow$ 1.39\%)\end{tabular}} &
  \multicolumn{1}{c|}{\begin{tabular}[c]{@{}c@{}}79.0\\ ($\downarrow$ 1.00\%)\end{tabular}} &
  \multicolumn{1}{c|}{\begin{tabular}[c]{@{}c@{}}81.1\\ ($\downarrow$ 0.86\%)\end{tabular}} &
  \multicolumn{1}{c|}{-} &
  - \\ \hline
\end{tabular}%
}

\begin{tablenotes}
\footnotesize
\item \textit{PH-region: Informative \textbf{P}atches and attention \textbf{H}eads identified during cascade pruning.}
\end{tablenotes}
\end{table}

\noindent \textbf{Cascade Pruning} Cascade pruning combines multiple pruning techniques to reduce parameters and giga FLOPS (GFLOPS) while preserving accuracy. A standout method named VTC-LFC~\cite{wang2022vtc} aimed to improve the identification of informative channels and tokens in a model, leading to better accuracy preservation. This approach introduced a bottom-up cascade (BCP) pruning strategy that gradually prunes tokens and channels, starting from the first block and advancing to the last. The pruning process is controlled by a hyperparameter called a \textbf{global allowable drop}, ensuring the performance drop remains within an acceptable range. Additionally, BCP guarantees efficient compression without sacrificing model performance by pruning each block and immediately stopping the compression process when the performance drop reaches a predefined threshold. Another cascade ViT pruning ~\cite{song2022cp} utilized the sparsity for pruning PH-regions (containing patches and heads) in the MSA \& FFN progressively and dynamically. The authors conducted experiments on three different types of datasets: ImageNet-1k~\cite{5206848}, CIFAR-10~\cite{cifar10}, and CIFAR-100~\cite{cifar10}.\\
\renewcommand{\arraystretch}{1.2}
\begin{table}[htb]
\centering
\caption{\highlight{GFLOPS reduction comparison of pruning techniques across different ViT models. '\textbf{$\downarrow$}' denotes the reduction rate from the baseline ViT model.}}
\label{tab:pruning_gflops}
\resizebox{0.6\columnwidth}{!}{%
\begin{tabular}{c|ccccc}
\hline
\multirow{2}{*}{\textbf{Algorithm}} &
  \multicolumn{5}{c}{\textbf{GFLOPS ($\downarrow$)}} \\ \cline{2-6} 
 &
  \multicolumn{1}{c|}{\textbf{DeiT-T}} &
  \multicolumn{1}{c|}{\textbf{DeiT-S}} &
  \multicolumn{1}{c|}{\textbf{DeiT-B}} &
  \multicolumn{1}{c|}{\textbf{Swin-S}} &
  \textbf{T2T-ViT-14} \\ \hline
\textbf{Baseline} &
  \multicolumn{1}{c|}{1.3} &
  \multicolumn{1}{c|}{4.6} &
  \multicolumn{1}{c|}{17.6} &
  \multicolumn{1}{c|}{8.7} &
  4.8 \\ \hline
VTP~\cite{zhu2021vision} &
  \multicolumn{1}{c|}{-} &
  \multicolumn{1}{c|}{-} &
  \multicolumn{1}{c|}{\begin{tabular}[c]{@{}c@{}}10.0\\ ($\downarrow$ 43.2\%)\end{tabular}} &
  \multicolumn{1}{c|}{-} &
  - \\ \hline
WDPruning~\cite{yu2022width} &
  \multicolumn{1}{c|}{\begin{tabular}[c]{@{}c@{}}0.7\\ ($\downarrow$ 46.2\%)\end{tabular}} &
  \multicolumn{1}{c|}{\begin{tabular}[c]{@{}c@{}}2.6\\ ($\downarrow$ 43.5\%)\end{tabular}} &
  \multicolumn{1}{c|}{\begin{tabular}[c]{@{}c@{}}9.9\\ ($\downarrow$ 43.8\%)\end{tabular}} &
  \multicolumn{1}{c|}{\begin{tabular}[c]{@{}c@{}}6.3\\ ($\downarrow$ 27.6\%)\end{tabular}} &
  - \\ \hline
MD~\cite{hou2022multi} &
  \multicolumn{1}{c|}{-} &
  \multicolumn{1}{c|}{\begin{tabular}[c]{@{}c@{}}2.9\\ ($\downarrow$ 37.0\%)\end{tabular}} &
  \multicolumn{1}{c|}{\begin{tabular}[c]{@{}c@{}}11.2\\ ($\downarrow$ 36.4\%)\end{tabular}} &
  \multicolumn{1}{c|}{-} &
  \begin{tabular}[c]{@{}c@{}}2.9\\ ($\downarrow$ 39.6\%)\end{tabular} \\ \hline
SPViT~\cite{kong2022spvit} &
  \multicolumn{1}{c|}{-} &
  \multicolumn{1}{c|}{-} &
  \multicolumn{1}{c|}{-} &
  \multicolumn{1}{c|}{\begin{tabular}[c]{@{}c@{}}6.4 \\ ($\downarrow$ 26.4\%)\end{tabular}} &
  - \\ \hline
PS-ViT~\cite{tang2022patch} &
  \multicolumn{1}{c|}{\begin{tabular}[c]{@{}c@{}}0.7 \\ ($\downarrow$ 46.2\%)\end{tabular}} &
  \multicolumn{1}{c|}{\begin{tabular}[c]{@{}c@{}}2.6 \\ ($\downarrow$ 43.5\%)\end{tabular}} &
  \multicolumn{1}{c|}{\begin{tabular}[c]{@{}c@{}}9.8 \\ ($\downarrow$ 44.3\%)\end{tabular}} &
  \multicolumn{1}{c|}{-} &
  \begin{tabular}[c]{@{}c@{}}3.1 \\ ($\downarrow$ 35.4\%)\end{tabular} \\ \hline
UVC~\cite{yu2022unified} &
  \multicolumn{1}{c|}{\textbf{\begin{tabular}[c]{@{}c@{}}0.5\\ ($\downarrow$ 61.5\%)\end{tabular}}} &
  \multicolumn{1}{c|}{\textbf{\begin{tabular}[c]{@{}c@{}}2.3\\ ($\downarrow$ 50.0\%)\end{tabular}}} &
  \multicolumn{1}{c|}{\begin{tabular}[c]{@{}c@{}}8.0\\ ($\downarrow$ 54.5\%)\end{tabular}} &
  \multicolumn{1}{c|}{-} &
  \textbf{\begin{tabular}[c]{@{}c@{}}2.1 \\ ($\downarrow$ 56.0\%)\end{tabular}} \\ \hline
NViT~\cite{yang2023global} &
  \multicolumn{1}{c|}{\begin{tabular}[c]{@{}c@{}}1.3 \\ ($\downarrow$ 0.0\%)\end{tabular}} &
  \multicolumn{1}{c|}{\begin{tabular}[c]{@{}c@{}}4.2 \\ ($\downarrow$ 8.7\%)\end{tabular}} &
  \multicolumn{1}{c|}{\textbf{\begin{tabular}[c]{@{}c@{}}6.8 \\ ($\downarrow$ 61.4\%)\end{tabular}}} &
  \multicolumn{1}{c|}{\textbf{\begin{tabular}[c]{@{}c@{}}6.2 \\ ($\downarrow$ 28.7\%)\end{tabular}}} &
  - \\ \hline
SAViT~\cite{zheng2022savit} &
  \multicolumn{1}{c|}{\begin{tabular}[c]{@{}c@{}}0.9 \\ ($\downarrow$ 30.8\%)\end{tabular}} &
  \multicolumn{1}{c|}{\begin{tabular}[c]{@{}c@{}}3.1 \\ ($\downarrow$ 32.6\%)\end{tabular}} &
  \multicolumn{1}{c|}{\begin{tabular}[c]{@{}c@{}}10.6 \\ ($\downarrow$ 39.8\%)\end{tabular}} &
  \multicolumn{1}{c|}{-} &
  - \\ \hline
S$^2$ViTE~\cite{chen2021chasing} &
  \multicolumn{1}{c|}{\begin{tabular}[c]{@{}c@{}}0.9 \\ ($\downarrow$ 30.8\%)\end{tabular}} &
  \multicolumn{1}{c|}{\begin{tabular}[c]{@{}c@{}}3.1\\ ($\downarrow$ 32.6\%)\end{tabular}} &
  \multicolumn{1}{c|}{\begin{tabular}[c]{@{}c@{}}11.8 \\ ($\downarrow$ 33.0\%)\end{tabular}} &
  \multicolumn{1}{c|}{-} &
  - \\ \hline
VTC-LFC~\cite{wang2022vtc} &
  \multicolumn{1}{c|}{\begin{tabular}[c]{@{}c@{}}0.76\\ ($\downarrow$ 41.7\%)\end{tabular}} &
  \multicolumn{1}{c|}{\begin{tabular}[c]{@{}c@{}}2.4\\ ($\downarrow$ 47.1\%)\end{tabular}} &
  \multicolumn{1}{c|}{\begin{tabular}[c]{@{}c@{}}8.03\\ ($\downarrow$ 54.4\%)\end{tabular}} &
  \multicolumn{1}{c|}{-} &
  - \\ \hline
CP-ViT~\cite{song2022cp} &
  \multicolumn{1}{c|}{\begin{tabular}[c]{@{}c@{}}0.74\\ ($\downarrow$ 43.3\%)\end{tabular}} &
  \multicolumn{1}{c|}{\begin{tabular}[c]{@{}c@{}}2.7\\ ($\downarrow$ 42.2\%)\end{tabular}} &
  \multicolumn{1}{c|}{\begin{tabular}[c]{@{}c@{}}10.28\\ ($\downarrow$ 41.6\%)\end{tabular}} &
  \multicolumn{1}{c|}{} &
   \\ \hline
\end{tabular}%
}
\end{table}

\noindent \textbf{Miscellaneous Approaches in Pruning} Hou et al.~\cite{hou2022multi} introduced a multi-dimensional pruning strategy for ViTs, leveraging a statistical dependence-based criterion to identify and remove redundant components across different dimensions. Beyond this, several pruning techniques have been developed to accelerate ViTs, particularly for edge devices, including column balanced block pruning ~\cite{9424344}, end-to-end exploration ~\cite{chen2021chasing}, gradient-based learned runtime pruning ~\cite{li2022accelerating}. These techniques have shown stability in applying pruning on ViT models without compromising accuracy.\\

\renewcommand{\arraystretch}{1.2}
\begin{table}[]
\centering
\caption{\highlight{Parameter (Million) reduction comparison of pruning techniques across different ViT models. '\textbf{$\downarrow$}' denotes parameter reduction rate from the baseline ViT model.}}
\label{tab:pruning_params}
\resizebox{0.5\columnwidth}{!}{%
\begin{tabular}{c|cccc}
\hline
\multirow{2}{*}{\textbf{Algorithm}} &
  \multicolumn{4}{c}{\textbf{Params (M $\downarrow$)}} \\ \cline{2-5} 
 &
  \multicolumn{1}{c|}{\textbf{DeiT-T}} &
  \multicolumn{1}{c|}{\textbf{DeiT-S}} &
  \multicolumn{1}{c|}{\textbf{DeiT-B}} &
  \textbf{Swin-S} \\ \hline
\textbf{Baseline} &
  \multicolumn{1}{c|}{5.6} &
  \multicolumn{1}{c|}{22} &
  \multicolumn{1}{c|}{86} &
  50 \\ \hline
\begin{tabular}[c]{@{}c@{}}VTP\\ ~\cite{zhu2021vision}\end{tabular} &
  \multicolumn{1}{c|}{-} &
  \multicolumn{1}{c|}{-} &
  \multicolumn{1}{c|}{\begin{tabular}[c]{@{}c@{}}47.3\\ ($\downarrow$ 45.0\%)\end{tabular}} &
  - \\ \hline
\begin{tabular}[c]{@{}c@{}}WDPruning\\ ~\cite{yu2022width}\end{tabular} &
  \multicolumn{1}{c|}{\textbf{\begin{tabular}[c]{@{}c@{}}3.5 \\ ($\downarrow$ 37.5\%)\end{tabular}}} &
  \multicolumn{1}{c|}{\begin{tabular}[c]{@{}c@{}}13.3\\ ($\downarrow$ 39.5\%)\end{tabular}} &
  \multicolumn{1}{c|}{\begin{tabular}[c]{@{}c@{}}55.3 \\ ($\downarrow$ 35.7\%)\end{tabular}} &
  \begin{tabular}[c]{@{}c@{}}32.8\\ ($\downarrow$ 34.4\%)\end{tabular} \\ \hline
NViT~\cite{yang2023global} &
  \multicolumn{1}{c|}{\textbf{\begin{tabular}[c]{@{}c@{}}3.5 \\ ($\downarrow$ 37.5\%)\end{tabular}}} &
  \multicolumn{1}{c|}{\textbf{\begin{tabular}[c]{@{}c@{}}10.5 \\ ($\downarrow$ 52.3\%)\end{tabular}}} &
  \multicolumn{1}{c|}{\textbf{\begin{tabular}[c]{@{}c@{}}17.0\\ ($\downarrow$ 80.3\%)\end{tabular}}} &
  \textbf{\begin{tabular}[c]{@{}c@{}}15.0\\ ($\downarrow$ 70.0\%)\end{tabular}} \\ \hline
SAViT~\cite{zheng2022savit} &
  \multicolumn{1}{c|}{\begin{tabular}[c]{@{}c@{}}4.2\\ ($\downarrow$ 25.0\%)\end{tabular}} &
  \multicolumn{1}{c|}{\begin{tabular}[c]{@{}c@{}}14.7\\ ($\downarrow$ 33.2\%)\end{tabular}} &
  \multicolumn{1}{c|}{\begin{tabular}[c]{@{}c@{}}51.9\\ ($\downarrow$ 39.6\%)\end{tabular}} &
  - \\ \hline
S$^2$ViTE~\cite{chen2021chasing} &
  \multicolumn{1}{c|}{\begin{tabular}[c]{@{}c@{}}4.2\\ ($\downarrow$ 25.0\%)\end{tabular}} &
  \multicolumn{1}{c|}{\begin{tabular}[c]{@{}c@{}}14.6\\ ($\downarrow$ 33.6\%)\end{tabular}} &
  \multicolumn{1}{c|}{\begin{tabular}[c]{@{}c@{}}56.8\\ ($\downarrow$ 34.0\%)\end{tabular}} &
  - \\ \hline
VTC-LFC~\cite{wang2022vtc} &
  \multicolumn{1}{c|}{\begin{tabular}[c]{@{}c@{}}4.2\\ ($\downarrow$ 25.0\%)\end{tabular}} &
  \multicolumn{1}{c|}{\begin{tabular}[c]{@{}c@{}}15.3\\ ($\downarrow$ 30.4\%)\end{tabular}} &
  \multicolumn{1}{c|}{\begin{tabular}[c]{@{}c@{}}56.8\\ ($\downarrow$ 34.0\%)\end{tabular}} &
  - \\ \hline
\end{tabular}%
}
\end{table}

\subsubsection{Discussion}\hfill\\
Pruning is utilized as a fundamental way to reduce the computation of the pre-trained ViT models. For ViT, the development of the pruning methods has systematically covered each perspective of model design, making the current pruning methods more flexible and well-organized for ViT models. \highlight{We provide a comparative summary of these pruning methods in terms of Top-1 accuracy (Table~\ref{tab:pruning_accuracy}), GFLOPS reduction (Table~\ref{tab:pruning_gflops}), and parameter reduction in millions (Table~\ref{tab:pruning_params}), comparable to baseline models. We inherit the baseline results from the official implementations of the respective models, as detailed in the tables, and compute the corresponding accuracy drop, parameter reduction, and GFLOPs reduction rates to evaluate the impact of pruning. Most pruning methods evaluated in Tables~\ref{tab:pruning_accuracy}–\ref{tab:pruning_params} focus on the ImageNet-1K dataset~\cite{5206848}, with CP-ViT~\cite{song2022cp} being a notable exception, tested on CIFAR~\cite{cifar10} dataset. As shown in Table~\ref{tab:pruning_accuracy}, NViT~\cite{yang2023global} consistently achieves competitive top-1 (\%) accuracy across DeiT-Tiny (T), Small (S), Base (B), and Swin-S, with a minimal drop of only 5.55\% from the baseline. Interestingly, few studies(e.g, SPViT, VTP) have explored pruning across diverse ViT models such as Swin or T2T-ViT-14, highlighting a gap in the generalizability of current techniques. In terms of computational savings, UVC~\cite{yu2022unified} achieves the highest GFLOPs reduction for DeiT-T/S variants. Notably, NViT also delivers substantial reductions in GFLOPs for more complex models, such as DeiT-B, while improving Top-1 accuracy—a noteworthy exception to the typical trade-off where aggressive pruning often leads to performance loss. Additionally, both WDPruning~\cite{yu2022width} and NViT~\cite{yang2023global} achieved comparable parameter sizes (see Table~\ref{tab:pruning_params}) when applied to DeiT-T. However, NViT demonstrated a more substantial reduction in the number of parameters than other ViT models.} While these results demonstrate notable progress in pruning ViTs, the training and finetuning costs remain a major limitation, particularly for deployment on resource-constrained edge devices. Therefore, training-efficient or finetuned free pruning techniques need more attention in the near future for efficient deployment on the edge. This necessitates a more precise estimation of parameter or block sensitivity using limited data, as well as a deeper exploration of the information embedded within hidden features during training and inference.
\begin{figure}[htbp]
    \centering

    \begin{subfigure}[b]{0.45\textwidth}
        \centering
        \includegraphics[width=\textwidth]{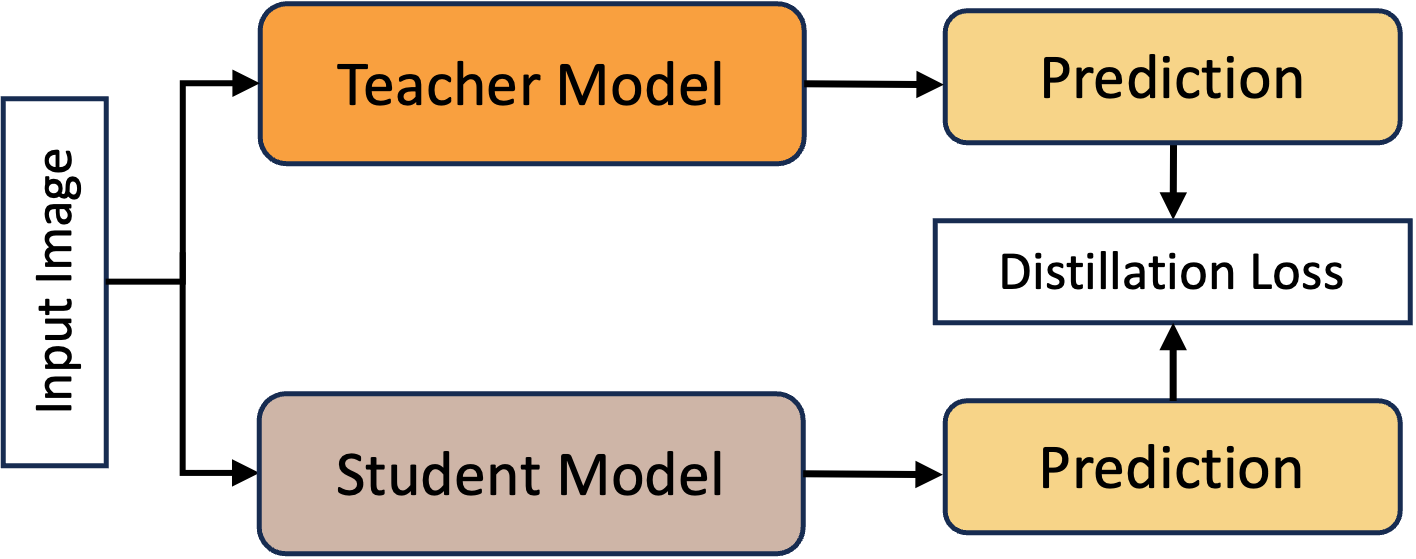}
        \caption{A schematic illustration of the logit-based KD technique.}
        \label{fig:logit}
    \end{subfigure}
    \hfill
    \begin{subfigure}[b]{0.45\textwidth}
        \centering
        \includegraphics[width=\textwidth]{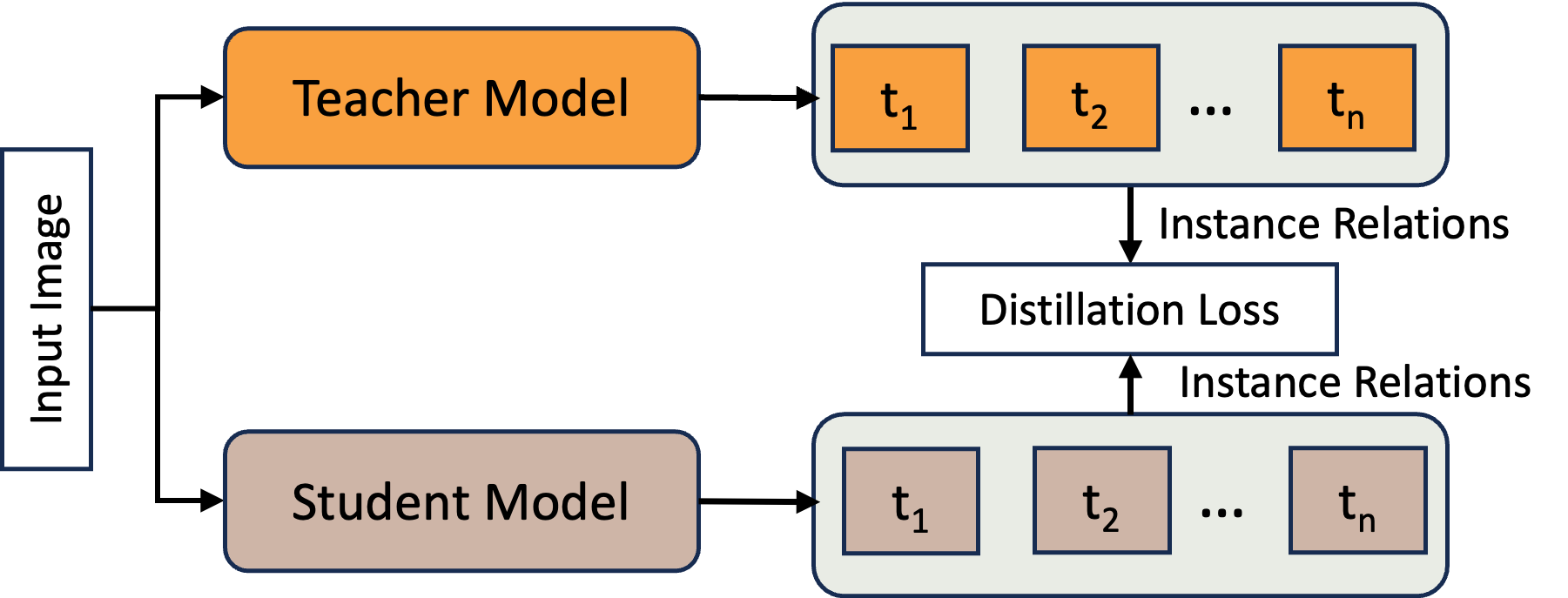}
        \caption{A schematic illustration of the relation-based KD technique.}
        \label{fig:relation}
    \end{subfigure}

    \vspace{0.5cm} 

    \begin{subfigure}[b]{0.45\textwidth}
        \centering
        \includegraphics[width=\textwidth]{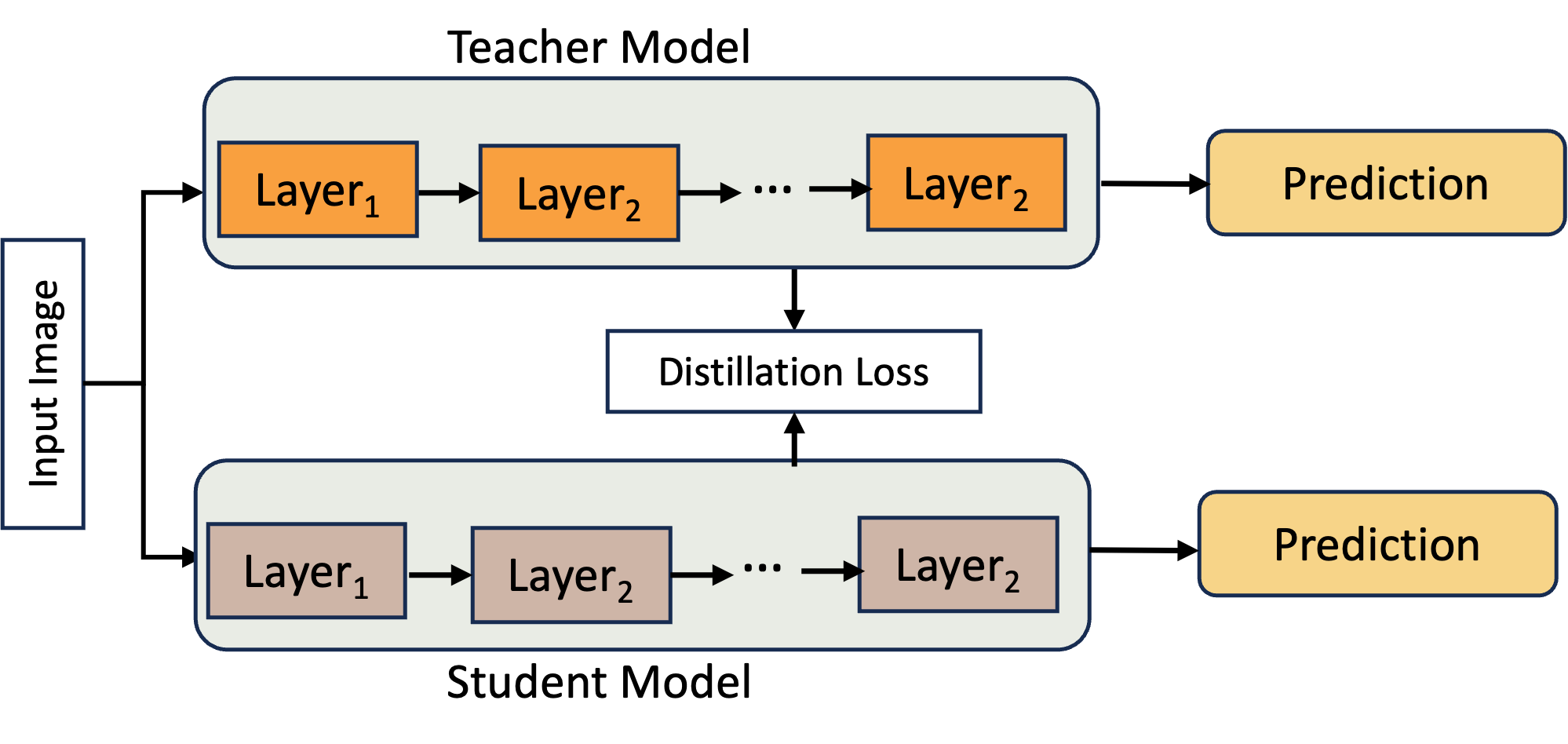}
        \caption{A schematic illustration of the feature-based KD technique.}
        \label{fig:feature}
    \end{subfigure}

    \caption{\highlight{An illustration depicting the three types of KD: logit-based, relation-based, and feature-based between teacher and student models.}}
    \label{fig:kd_types}
\end{figure}

\subsection{Knowledge Distillation}\label{know}
Knowledge distillation (KD) is another model compression technique in machine learning where a smaller model (the "student") is trained to reproduce the behavior of a larger model (the "teacher"). The purpose is to transfer the "knowledge" from the larger model to the smaller one, thereby reducing computational resources without significantly losing accuracy. Pruning is a direct way to reduce the complexity of the original model, whereas KD involves training a new, more compact model that is easy to deploy. By using KD during finetuning, the pruned model can benefit from the insights and information captured by the larger teacher model. KD helps to regain the original performance and compensate for the accuracy loss during the other compression techniques (e.g., pruning, quantization).

\subsubsection{KD techniques for \highlight{V}ision \highlight{T}ransformer}\label{kd_in_vit}\hfill\\
\highlight{Current KD techniques on ViT models can be broadly classified into three types: logit-based~\cite{wu2022tinyvit,chen2022dearkd}, relation-based~\cite{ren2022co,lin2022knowledge}, and feature-based~\cite{hao2022learning,zhang2022minivit,9747484} methods. Feature-based KD enables the student model to capture richer semantic representations by aligning intermediate features across hidden layers. Both logit-based and feature-based KD typically focus on transferring knowledge from individual data samples. In contrast, relation-based KD extends this idea by modeling relational knowledge across the whole dataset, aiming to capture inter-sample relationships. Figure~\ref{fig:kd_types} illustrates a schematic overview of these KD types, providing a visual summary of their key differences. However, A detailed analysis of these three categories has also been conducted by Yang et al.\cite{yang2023categories}.  In the following subsections, we provide a detailed discussion of the current proposed techniques on ViT models of each category.}\\

\renewcommand{\arraystretch}{1.2}
\begin{table}[htb]
\centering
\caption{\highlight{Results of KD techniques applied to ViT models for classification tasks.} MSE refers to Mean Squared Error loss. \highlight{'$\uparrow$' in the student model indicates the performance improvement over the baseline student model.}}
\label{tab:KD_result_classification}
\resizebox{\columnwidth}{!}{%
\begin{tabular}{c|c|c|c|c|cc|cc}
\hline
\multirow{2}{*}{\textbf{Type}} &
  \multirow{2}{*}{\textbf{Algorithm}} &
  \multirow{2}{*}{\textbf{Method}} &
  \multirow{2}{*}{\textbf{Loss function}} &
  \multirow{2}{*}{\textbf{Dataset}} &
  \multicolumn{2}{c|}{\textbf{Teachers}} &
  \multicolumn{2}{c}{\textbf{Students}} \\ \cline{6-9} 
 &
   &
   &
   &
   &
  \multicolumn{1}{c|}{\textbf{Models}} &
  \textbf{\begin{tabular}[c]{@{}c@{}}Top-1\\ (\%)\end{tabular}} &
  \multicolumn{1}{c|}{\textbf{Models}} &
  \textbf{\begin{tabular}[c]{@{}c@{}}Top-1\\ (\%)\end{tabular}} \\ \hline
\multirow{6}{*}{\begin{tabular}[c]{@{}c@{}}Relation\end{tabular}} &
  \multirow{2}{*}{TAT~\cite{lin2022knowledge}} &
  \multirow{2}{*}{\begin{tabular}[c]{@{}c@{}}One-to-all spatial\\ matching KD\end{tabular}} &
  \multirow{2}{*}{\begin{tabular}[c]{@{}c@{}}Vanilla distillation \\ + L\textsubscript2 loss\end{tabular}} &
  \multirow{2}{*}{\begin{tabular}[c]{@{}c@{}}ImageNet-1k~\cite{5206848}\end{tabular}} &
  \multicolumn{1}{c|}{\multirow{2}{*}{ResNet34}} &
  \multirow{2}{*}{72.4\%} &
  \multicolumn{1}{c|}{\multirow{2}{*}{ResNet18}} &
  \multirow{2}{*}{\begin{tabular}[c]{@{}c@{}}72.1\\ ($\uparrow$ 2.0\%)\end{tabular}} \\
 &
   &
   &
   &
   &
  \multicolumn{1}{c|}{} &
   &
  \multicolumn{1}{c|}{} &
   \\ \cline{2-9} 
 &
  \multirow{4}{*}{Co-advise~\cite{ren2022co}} &
  \multirow{4}{*}{\begin{tabular}[c]{@{}c@{}}Co-advised by \\ lightweight teachers\end{tabular}} &
  \multirow{4}{*}{\begin{tabular}[c]{@{}c@{}}Kull back divergence\\  + Cross entropy loss\end{tabular}} &
  \multirow{4}{*}{\begin{tabular}[c]{@{}c@{}}ImageNet-1k~\cite{5206848}\end{tabular}} &
  \multicolumn{1}{c|}{\multirow{4}{*}{ResNet18}} &
  \multirow{4}{*}{83.4\%} &
  \multicolumn{1}{c|}{\multirow{4}{*}{CiT-Ti}} &
  \multirow{4}{*}{\begin{tabular}[c]{@{}c@{}}88.0\\ ($\uparrow$ 1.5\%)\end{tabular}} \\
 &
   &
   &
   &
   &
  \multicolumn{1}{c|}{} &
   &
  \multicolumn{1}{c|}{} &
   \\
 &
   &
   &
   &
   &
  \multicolumn{1}{c|}{} &
   &
  \multicolumn{1}{c|}{} &
   \\
 &
   &
   &
   &
   &
  \multicolumn{1}{c|}{} &
   &
  \multicolumn{1}{c|}{} &
   \\ \hline
\multirow{9}{*}{\begin{tabular}[c]{@{}c@{}}Feature\end{tabular}} &
  \multirow{3}{*}{Manifold ~\cite{hao2022learning}} &
  \multirow{3}{*}{\begin{tabular}[c]{@{}c@{}}Patch-level manifold\\ space method\end{tabular}} &
  \multirow{3}{*}{\begin{tabular}[c]{@{}c@{}}Manifold \\distillation loss\end{tabular}} &
  \multirow{3}{*}{\begin{tabular}[c]{@{}c@{}}ImageNet-1k~\cite{5206848}\end{tabular}} &
  \multicolumn{1}{c|}{\multirow{2}{*}{CaiT-S24}} &
  83.4\% &
  \multicolumn{1}{c|}{DeiT-T} &
  \begin{tabular}[c]{@{}c@{}}76.5\\ ($\uparrow$ 4.3\%)\end{tabular} \\ \cline{7-9} 
 &
   &
   &
   &
   &
  \multicolumn{1}{c|}{} &
  83.4\% &
  \multicolumn{1}{c|}{DeiT-S} &
  \begin{tabular}[c]{@{}c@{}}82.2\\ ($\uparrow$ 2.3\%)\end{tabular} \\ \cline{6-9} 
 &
   &
   &
   &
   &
  \multicolumn{1}{c|}{Swin-S} &
  83.2\% &
  \multicolumn{1}{c|}{Swin-T} &
  \begin{tabular}[c]{@{}c@{}}82.2\\ ($\uparrow$ 1.0\%)\end{tabular} \\ \cline{2-9} 
 &
  \multirow{3}{*}{AProbe ~\cite{9747484}} &
  \multirow{3}{*}{\begin{tabular}[c]{@{}c@{}}Probe \& Knowledge \\ distillation\end{tabular}} &
  \multirow{3}{*}{\begin{tabular}[c]{@{}c@{}}Probe distillation + \\ Cross-entropy loss\end{tabular}} &
  \begin{tabular}[c]{@{}c@{}}CIFAR-100~\cite{cifar10}\end{tabular} &
  \multicolumn{1}{c|}{\multirow{3}{*}{DeiT-XS}} &
  76.30\% &
  \multicolumn{1}{c|}{\multirow{3}{*}{DeiT-XT}} &
  \begin{tabular}[c]{@{}c@{}}71.8\\ ($\uparrow$ 6.36\%)\end{tabular} \\ \cline{5-5} \cline{7-7} \cline{9-9} 
 &
   &
   &
   &
  \begin{tabular}[c]{@{}c@{}}CIFAR-10~\cite{cifar10}\end{tabular} &
  \multicolumn{1}{c|}{} &
  96.65\% &
  \multicolumn{1}{c|}{} &
  \begin{tabular}[c]{@{}c@{}}93.9\\ ($\uparrow$ 7.64\%)\end{tabular} \\ \cline{5-5} \cline{7-7} \cline{9-9} 
 &
   &
   &
   &
  \begin{tabular}[c]{@{}c@{}}MNIST~\cite{deng2012mnist}\end{tabular} &
  \multicolumn{1}{c|}{} &
  99.39\% &
  \multicolumn{1}{c|}{} &
  \begin{tabular}[c]{@{}c@{}}99.1\\ ($\uparrow$ 0.01\%)\end{tabular} \\ \cline{2-9} 
 &
  \multirow{3}{*}{MiniViT ~\cite{zhang2022minivit}} &
  \multirow{3}{*}{Weight distillation} &
  \multirow{3}{*}{\begin{tabular}[c]{@{}c@{}}Self-attention +\\ Hidden-state + \\ Prediction loss\end{tabular}} &
  \multirow{3}{*}{\begin{tabular}[c]{@{}c@{}}ImageNet-1k~\cite{5206848}\end{tabular}} &
  \multicolumn{1}{c|}{\multirow{3}{*}{\begin{tabular}[c]{@{}c@{}}RegNet-\\ 16GF\end{tabular}}} &
  \multirow{3}{*}{82.9\%} &
  \multicolumn{1}{c|}{\multirow{3}{*}{DeiT-B}} &
  \multirow{3}{*}{\begin{tabular}[c]{@{}c@{}}83.2\\ ($\uparrow$ 1.4\%)\end{tabular}} \\
 &
   &
   &
   &
   &
  \multicolumn{1}{c|}{} &
   &
  \multicolumn{1}{c|}{} &
   \\
 &
   &
   &
   &
   &
  \multicolumn{1}{c|}{} &
   &
  \multicolumn{1}{c|}{} &
   \\ \hline
\multirow{4}{*}{\begin{tabular}[c]{@{}c@{}}Logit\end{tabular}} &
  \multirow{3}{*}{TinyViT ~\cite{wu2022tinyvit}} &
  \multirow{3}{*}{\begin{tabular}[c]{@{}c@{}}Reusing teacher \\ predictions \& data \\ augmentation\end{tabular}} &
  \multirow{3}{*}{Cross entropy loss} &
  \multirow{3}{*}{\begin{tabular}[c]{@{}c@{}}ImageNet-1k~\cite{5206848}\end{tabular}} &
  \multicolumn{1}{c|}{\multirow{3}{*}{\begin{tabular}[c]{@{}c@{}}CLIP-\\ ViT-L\end{tabular}}} &
  \multirow{3}{*}{84.8\%} &
  \multicolumn{1}{c|}{Swin-T} &
  \begin{tabular}[c]{@{}c@{}}83.4\\ ($\uparrow$ 2.2\%)\end{tabular} \\ \cline{8-9} 
 &
   &
   &
   &
   &
  \multicolumn{1}{c|}{} &
   &
  \multicolumn{1}{c|}{\multirow{2}{*}{DeiT-S}} &
  \multirow{2}{*}{\begin{tabular}[c]{@{}c@{}}82.0\\ ($\uparrow$ 2.1\%)\end{tabular}} \\
 &
   &
   &
   &
   &
  \multicolumn{1}{c|}{} &
   &
  \multicolumn{1}{c|}{} &
   \\ \cline{2-9} 
 &
  DearKD ~\cite{chen2022dearkd} &
  Self-generative data &
  \begin{tabular}[c]{@{}c@{}}MSE distillation +\\ Cross entropy+ Intra\\ -divergence loss\end{tabular} &
  \begin{tabular}[c]{@{}c@{}}ImageNet-1k~\cite{5206848}\end{tabular} &
  \multicolumn{1}{c|}{\begin{tabular}[c]{@{}c@{}}ResNet-\\ 101\end{tabular}} &
  77.37\% &
  \multicolumn{1}{c|}{DeiT-Ti} &
  \begin{tabular}[c]{@{}c@{}}71.2\\ ($\downarrow$ 1.0\%)\end{tabular} \\ \hline
\end{tabular}%
}
\begin{tablenotes}
\small
\centering
\item[] \highlight{\textit{Co-advised~\cite{ren2022co}: Transformer-Ti is denoted as \textbf{CiT-Ti} in our table.}}
\end{tablenotes}
\end{table}
\noindent \highlight{\textbf{Logit-based KD} Logit-based KD (see Figure~\ref{fig:logit}) focuses on learning knowledge from the last layer and aims to align the final predictions between the teacher and student.} Touvron et al.~\cite{touvron2021training} \highlight{pioneered a benchmark KD technique} for ViTs by enabling training with substantially smaller datasets. The authors introduced a distillation token, an additional learnable vector used alongside the class token during training. Recently, TinyViT~\cite{wu2022tinyvit} highlighted that smaller ViTs could benefit from larger teacher models trained on extensive datasets, such as distilling the student model on ImageNet-21k and finetuning on ImageNet-1k. This approach reduced the computation significantly during the training of the student model. To optimize computational memory, TinyViT introduced a strategy that pre-stores data augmentation details and logits for large teacher models, reducing memory overhead. \highlight{One problem with TinyViT is that teacher models still use large datasets.} To solve this issue, DearKD ~\cite{chen2022dearkd} proposed the KD methods on self-generative data and used representational KD on intermediate features with response-based KD. The proposed paper used mean square error (MSE) distillation loss for hidden features, CE loss for hard label distillation, and intra-divergence distillation loss function to calculate the loss. It was noteworthy that DearKD surpassed the performance of the baseline ViT model trained with ImageNet-21K datasets, even though it only used 50\% of the data. \highlight{However, logit-based KD only transfers output-level predictions, neglecting intermediate representations such as features or attention. This shallow supervision limits its effectiveness, especially when the student and teacher architectures differ significantly.} \\

\renewcommand{\arraystretch}{1.2}
\begin{table}[htb]
\centering
\caption{Results of KD techniques applied to ViT models for object detection tasks.}
\label{tab:KD_object}
\resizebox{0.8\columnwidth}{!}{%
\begin{tabular}{c|c|c|cccc}
\hline
\multirow{3}{*}{\textbf{Algorithm}} &
  \multirow{3}{*}{\textbf{Pretrained Dataset}} &
  \multirow{3}{*}{\textbf{Dataset}} &
  \multicolumn{4}{c}{\textbf{Students}} \\ \cline{4-7} 
 &
   &
   &
  \multicolumn{2}{c|}{\textbf{Without KD}} &
  \multicolumn{2}{c}{\textbf{With KD}} \\ \cline{4-7} 
 &
   &
   &
  \multicolumn{1}{c|}{\textbf{Models}} &
  \multicolumn{1}{c|}{\textbf{AP\textsuperscript{box}}} &
  \multicolumn{1}{c|}{\textbf{Models}} &
  \textbf{AP\textsuperscript{box}} \\ \hline
Manifold ~\cite{hao2022learning} &
  ImageNet-1k~\cite{5206848} &
  COCO-2017 &
  \multicolumn{1}{c|}{Swin-T} &
  \multicolumn{1}{c|}{43.7} &
  \multicolumn{1}{c|}{Swin-T} &
  44.7($\uparrow$1.0) \\ \hline
MiniViT ~\cite{zhang2022minivit} &
  - &
  COCO-2017 &
  \multicolumn{1}{c|}{Swin-T} &
  \multicolumn{1}{c|}{48.1} &
  \multicolumn{1}{c|}{Swin-T} &
  48.6($\uparrow$0.5) \\ \hline
\end{tabular}%
}
\end{table}
\noindent \highlight{\textbf{Feature-based KD} As we mentioned above, logit-based KD neglects intermediate-level supervision for complete guidance. To solve this problem, feature-based KD (see Figure~\ref{fig:feature}) focuses on exploring intermediate feature information, such as feature maps and their refined information.} Hao et al.~\cite{hao2022learning} introduced a fine-grained manifold distillation technique, \highlight{which considered ViT as feature projectors that map image patches into a sequence of high-dimensional manifold spaces.} The authors then teach the student layers to generate output features having the same patch-level manifold structure as the teacher layer for manually selected teacher-student layers. These output features are normalized and reshaped to compute a manifold relation map, a representation of the manifold structure of the features. However, the manifold relation map computation is resource-consuming and needs to be simplified. To address this computational issue, the authors proposed a decomposition strategy that separates the manifold relation map into three components: \highlight{intra-image, inter-image, and randomly sampled relation maps.} A dedicated manifold distillation loss (MD Loss) is then applied by aggregating the losses computed from each decoupled manifold relation map. \highlight{However, the proposed method only relies on prediction-level distillation.} To consider the other level (e.g., attention level), Minivit~\cite{zhang2022minivit} considered both attention level and hidden-state distillation, guiding a small model to replicate the behavior of the teacher model. The key idea of the proposed method is to multiplex the weights of consecutive transformer blocks by sharing the weights of the MSA and MLP layers. Besides, weight distillation over self-attention transfers knowledge from a large teacher model to a compact student. \highlight{To ensure effective knowledge transfer, MiniViT introduced a multi-level distillation loss that combines three components: (i) a prediction-logit loss that aligns the student's output with the teacher's softened logits, (ii) an attention distillation loss that encourages similarity in self-attention, and (iii) a hidden-state distillation loss that preserves structural relationships. Furthermore, Wang et al.~\cite{9747484} introduced another feature-based  KD technique for ViT named attention probes (AProbes), which streamline ViTs using unlabeled data. The proposed method operates in two stages: first, an attention-based probing mechanism identifies informative samples from diverse unlabeled datasets. Next, a probe-guided distillation process transfers knowledge to the student model by aligning the final predictions and intermediate features with the teacher. Both stages are achieved through cross-entropy loss and a dedicated probe distillation loss. However, feature-based KD methods only focus on individual sample representations despite leveraging intermediate features and often overlook the underlying dependencies across layers. This limitation restricts their ability to capture the holistic knowledge embedded in the teacher model fully. To overcome this limitation, relation-based KD aims to model and transfer relational information.}\\

\noindent \highlight{\textbf{Relation-based KD} Relation-based KD (see Figure~\ref{fig:relation}) is often modeled as an instance graph, where nodes represent sample embeddings and edges are weighted by similarity metrics. One notable example is Co-advise~\cite{ren2022co}, which introduced a multi-teacher framework where lightweight teacher models with diverse inductive biases (e.g., CNNs) collaboratively guide a student ViT model. The key insight is that the inductive bias of a teacher has a more significant impact on student performance than their standalone accuracy. By distilling complementary inductive biases, the student gains a more comprehensive representation of the data. A token-level alignment mechanism is also proposed to match the inductive bias of student tokens with their respective teachers.} Another method, target-aware transformer (TAT)~\cite{lin2022knowledge}, addressed spatial misalignment in distillation. It transferred each pixel in the teacher feature map to all spatial locations in the student using a transformer-based similarity mechanism, enabling fine-grained relational alignment. \highlight{While Co-advise emphasizes architectural diversity and inductive bias, and TAT focuses on spatial adaptability, both exemplify the strength of relation-based KD in enhancing knowledge transfer beyond isolated feature matching.}

\renewcommand{\arraystretch}{1.2}
\begin{table}[htb]
\centering
\caption{Results of KD techniques applied to ViT models for segmentation tasks.}
\label{tab:semantic_segmentation}
\resizebox{0.83\columnwidth}{!}{%
\begin{tabular}{c|c|c|cc|cc|cc}
\hline
\multirow{2}{*}{\textbf{Algorithm}} &
  \multirow{2}{*}{\textbf{Dataset}} &
  \multirow{2}{*}{\textbf{Metrics}} &
  \multicolumn{2}{c|}{\textbf{Teachers}} &
  \multicolumn{2}{c|}{\textbf{Students}} &
  \multicolumn{2}{c}{\textbf{Proposed}} \\ \cline{4-9} 
 &
   &
   &
  \multicolumn{1}{c|}{\textbf{Models}} &
  \textbf{Result} &
  \multicolumn{1}{c|}{\textbf{Models}} &
  \textbf{Result} &
  \multicolumn{1}{c|}{\textbf{Models}} &
  \textbf{Result} \\ \hline
\begin{tabular}[c]{@{}c@{}}Manifold~\cite{hao2022learning}\end{tabular} &
  ADE20K~\cite{zhou2019semantic} &
  mIoU &
  \multicolumn{1}{c|}{\begin{tabular}[c]{@{}c@{}}Swin-S + \\ UPerNet\end{tabular}} &
  47.64 &
  \multicolumn{1}{c|}{\begin{tabular}[c]{@{}c@{}}Swin-S + \\ UPerNet\end{tabular}} &
  44.51 &
  \multicolumn{1}{c|}{\begin{tabular}[c]{@{}c@{}}Swin-T + \\ UPerNet\end{tabular}} &
  \begin{tabular}[c]{@{}c@{}}45.66\\ ($\uparrow$2.58\%)\end{tabular} \\ \hline
\multirow{2}{*}{\begin{tabular}[c]{@{}c@{}}TAT~\cite{lin2022knowledge}\end{tabular}} &
  \begin{tabular}[c]{@{}c@{}}COCO-\\ Stuff10k~\cite{caesar2018cvpr}\end{tabular} &
  mIoU &
  \multicolumn{1}{c|}{ResNet18} &
  33.10 &
  \multicolumn{1}{c|}{ResNet18} &
  26.33 &
  \multicolumn{1}{c|}{ResNet18} &
  \begin{tabular}[c]{@{}c@{}}28.75\\ ($\uparrow$9.09\%)\end{tabular} \\ \cline{2-9} 
 &
  \begin{tabular}[c]{@{}c@{}}Pascal \\ VOC~\cite{pascal}\end{tabular} &
  mIoU &
  \multicolumn{1}{c|}{ResNet18} &
  78.43 &
  \multicolumn{1}{c|}{ResNet18} &
  72.07 &
  \multicolumn{1}{c|}{ResNet18} &
  \begin{tabular}[c]{@{}c@{}}75.76\\ ($\uparrow$9.28\%)\end{tabular} \\ \hline
\end{tabular}%
}
\end{table}
\subsubsection{Discussion}\hfill\\
A key strength of ViT models lies in their scalability to high parametric complexity; however, this demands significant computational resources and incurs substantial costs. KD offers a way to transfer knowledge into more compact student models, yet challenges remain, particularly in the vision domain. 

\highlight{From the results in Tables (\ref{tab:KD_result_classification}–\ref{tab:semantic_segmentation}), we observe a consistent trend: KD techniques can significantly improve the performance of student ViT models. In Table~\ref{tab:KD_result_classification}, relation-based KD techniques such as Co-advise~\cite{ren2022co} demonstrate noticeable top-1 accuracy improvements on the ImageNet-1k dataset, emphasizing the effectiveness of capturing inter-sample relationships. Feature-based techniques like MiniViT~\cite{zhang2022minivit} achieve considerably higher top-1 accuracy than manifold distillation~\cite{hao2022learning} by leveraging intermediate-level information such as attention and hidden states. In contrast, while logit-based methods offer more straightforward supervision through output alignment, their performance improvements tend to be more subtle and, in some cases (e.g., DearKD~\cite{chen2022dearkd}) demonstrate slight accuracy degradation.}

\highlight{In object detection tasks (see Table~\ref{tab:KD_object}), KD techniques yield consistent gains in AP\textsuperscript{box}, indicating successful knowledge transfer in detection pipelines. Although the improvements are smaller than in classification tasks, they still affirm the benefit of distillation in dense prediction contexts. For segmentation tasks (see Table~\ref{tab:semantic_segmentation}), the application of KD leads to substantial mean intersection over union (mIoU) improvements. Additionally, we observe that the majority of existing KD approaches for ViT models have been proposed for classification tasks. At the same time, significantly fewer works have explored their applicability to object detection and segmentation tasks. This highlights a promising avenue for future research, particularly as vision-driven domains such as autonomous driving and medical imaging continue to gain momentum and demand high-performance yet efficient models.} 

\subsection{Quantization}\label{quan}
Quantization is used to reduce the bit-width of the data flowing through a NN model. So, it is used primarily for memory saving, faster inference times, and simplifying the operations for compute acceleration. That makes quantization essential for deploying NN on edge devices with limited computational capabilities. Quantization can be applied to different aspects of techniques. We organize our quantization discussion into two subsections. Firstly, we categorize different quantization techniques applied according to different aspects in Section~\ref{types_quanti}. Lastly, we discuss different quantization techniques in ViT in Section~\ref{vit_quanti}.

\subsubsection{Taxonomy of Quantization Methods} \label{types_quanti}
\hfill\\
\highlight{Table~\ref{table:quan_classification} categorizes quantization techniques based on their role in the quantization pipeline, including when quantization is applied (timing), how it is performed (scheme), how scaling factors are chosen (calibration), the granularity of quantization, and precision Strategies strategies.}\\

\noindent \textbf{Quantization Schemes} Quantization schemes are broadly categorized into uniform and non-uniform techniques. Any weight or activation values in an NN can follow either a uniform or non-uniform distribution. Uniform quantization maps continuous weight and activation values to discrete levels with equal spacing between quantized values. However, non-uniform quantization sets different quantization steps for different parts of the data based on their distribution and importance to the final performance of the model. Non-uniform quantization can improve accuracy but is often more complex to implement in hardware. \\
\renewcommand{\arraystretch}{1.1}
\begin{table}[htb]
\centering
\caption{\highlight{Categorization of quantization techniques across five key aspects: schemes, approaches, calibration methods, granularity levels, and deployment strategies.}}
\label{table:quan_classification}
\resizebox{0.8\columnwidth}{!}{%
\begin{tabular}{c|c|c|c}
\hline
\textbf{\begin{tabular}[c]{@{}c@{}}Aspects of\\ quantization\end{tabular}} &
  \textbf{Techniques} &
  \textbf{Advantages} &
  \textbf{Limitation} \\ \hline
\multirow{2}{*}{\begin{tabular}[c]{@{}c@{}}Quantization \\ schemes\end{tabular}} &
  \begin{tabular}[c]{@{}c@{}}Uniform\\ ~\cite{yuan2022ptq4vit,fang2020post}\end{tabular} &
  \begin{tabular}[c]{@{}c@{}}Simple implementation \& \\ fast inference\end{tabular} &
  \begin{tabular}[c]{@{}c@{}}Inflexible to \\ data distribution\end{tabular} \\ \cline{2-4} 
 &
  \begin{tabular}[c]{@{}c@{}}Non-uniform \\ ~\cite{oh2022non,jeon2022mr}\end{tabular} &
  \begin{tabular}[c]{@{}c@{}}High accuracy \& \\ Efficient bandwidth\end{tabular} &
  \begin{tabular}[c]{@{}c@{}}Complex implementation \&\\  data-dependent\end{tabular} \\ \hline
\multirow{2}{*}{\begin{tabular}[c]{@{}c@{}}Quantization \\ approaches\end{tabular}} &
  \begin{tabular}[c]{@{}c@{}}PTQ\\ ~\cite{lin2021fq,liu2021post}\end{tabular} &
  \begin{tabular}[c]{@{}c@{}}No retraining required \&\\ low overhead\end{tabular} &
  Accuracy drop \\ \cline{2-4} 
 &
  \begin{tabular}[c]{@{}c@{}}QAT\\ ~\cite{9992209, zhang2023qd,li2022q}\end{tabular} &
  \begin{tabular}[c]{@{}c@{}}Preserves accuracy \&\\ robust\end{tabular} &
  \begin{tabular}[c]{@{}c@{}}Requires retraining \& \\ computational intensive\end{tabular} \\ \hline
\multirow{2}{*}{\begin{tabular}[c]{@{}c@{}}Calibration \\ methods\end{tabular}} &
  \begin{tabular}[c]{@{}c@{}}Static\\ ~\cite{fan2019static,liu2021improving}\end{tabular} &
  Lightweight &
  Accuracy drop \\ \cline{2-4} 
 &
  \begin{tabular}[c]{@{}c@{}}Dynamic\\ ~\cite{liu2022instance,huang2023structured}\end{tabular} &
  \begin{tabular}[c]{@{}c@{}}Adaptive to \\ input variations\end{tabular} &
  Runtime overhead \\ \hline
\multirow{2}{*}{Granularity} &
  \begin{tabular}[c]{@{}c@{}}Layer-wise \\ ~\cite{li2019fully,chu2019mixed,ranjan2024lrp}\end{tabular} &
  Simple implementation &
  \begin{tabular}[c]{@{}c@{}}Suboptimal precision\\  for sensitive layers\end{tabular} \\ \cline{2-4} 
 &
  \begin{tabular}[c]{@{}c@{}}Channel-wise\\ ~\cite{zhang2022mffnet,li2019fully,xie2023joint}\end{tabular} &
  Fine-grained control &
  Computational intensive \\ \hline
\multirow{2}{*}{Precision} &
  \begin{tabular}[c]{@{}c@{}}Mixed-Precision\\ ~\cite{chu2019mixed,chu2021mixed,xu2023q,wang2019haq}\end{tabular} &
  \begin{tabular}[c]{@{}c@{}}Balances accuracy \& \\ efficiency\end{tabular} &
  Not hardware-friendly \\ \cline{2-4} 
 &
  \begin{tabular}[c]{@{}c@{}}Hardware-aware\\ ~\cite{wang2019haq,dong2021hao}\end{tabular} &
  \begin{tabular}[c]{@{}c@{}}Hardware specific \\ optimization\end{tabular} &
  Platform-specific \\ \hline
\end{tabular}%
}
\end{table}

\noindent \textbf{Quantization Approaches} Quantization approaches can be broadly classified based on whether they require retraining. Quantization-aware training (QAT) incorporates quantization into both forward and backward passes during training, allowing the model to adapt to lower-precision representations. However, QAT is resource-intensive due to the need for retraining. In contrast, post-training quantization (PTQ) is a more efficient approach that applies quantization after a model has been fully trained in floating point (FP) precision, reducing the precision of weights and activations without additional training. While PTQ is less resource-intensive, it typically results in a higher accuracy drop compared to QAT~\cite{nagel2021white}.\\

\noindent \textbf{Calibration Methods} Calibration is a needed process of determining the appropriate scaling factors during finetuning methods that map the continuous range of FP values to discrete integer values. Calibration mainly ensures that the range of the quantized values matches the range of the original FP values as closely as possible. There are two types when choosing the range. One is dynamic quantization, and the other is static quantization. The weights are quantized statically in dynamic quantization, but activations are quantized dynamically at runtime. Static quantization is quantized post-training. Unlike dynamic quantization, static quantization applies to weights and activations before deploying the model.\\

\noindent \textbf{Granularity} Another aspect of quantization techniques is the granularity of the clipping range of an NN. Layer-wise quantization is one of the granularity techniques where all weights and activations within a layer are quantized using the same scale. Channel-wise quantization, also referred to as per-channel quantization, is another granularity technique applied during the quantization of NN. Different scaling factors are computed for each channel of the weights in channel-wise quantization, meaning different layers can use different quantization parameters. \\

\noindent \textbf{Precision Strategies} \highlight{The precision strategies can be divided into three types:} fixed precision, mixed-precision, and hardware-aware precision. \highlight{In fixed-precision quantization, the entire model is quantized using a uniform bit-width (e.g., INT8), simplifying implementation but potentially leading to suboptimal accuracy for layers more sensitive to quantization. To address this limitation, mixed-precision during quantization has been introduced as a more flexible alternative.}. Mixed-precision uses different parts (e.g., channels, layers) of the model that are quantized to different numerical precisions. \highlight{Unlike fixed precision,} where the entire model is quantized to the same bit-width (like INT8), mixed-precision involves carefully choosing the bit-width for each layer or even each channel within a layer based on their sensitivity and contribution to the final performance of the model. \highlight{Despite these advantages, mixed-precision quantization is often not hardware-friendly due to the overhead of managing multiple precision levels within the same model.} Hardware-aware quantization has emerged as a solution that tailors a neural network's precision reduction process according to the hardware requirement. This technique optimizes the model for the target hardware by adjusting the quantization parameters to match the hardware's operations~\cite{wang2019haq}, such as latency and throughput.
\begin{figure}[htb]
    \centering
    \includegraphics[scale=0.5]{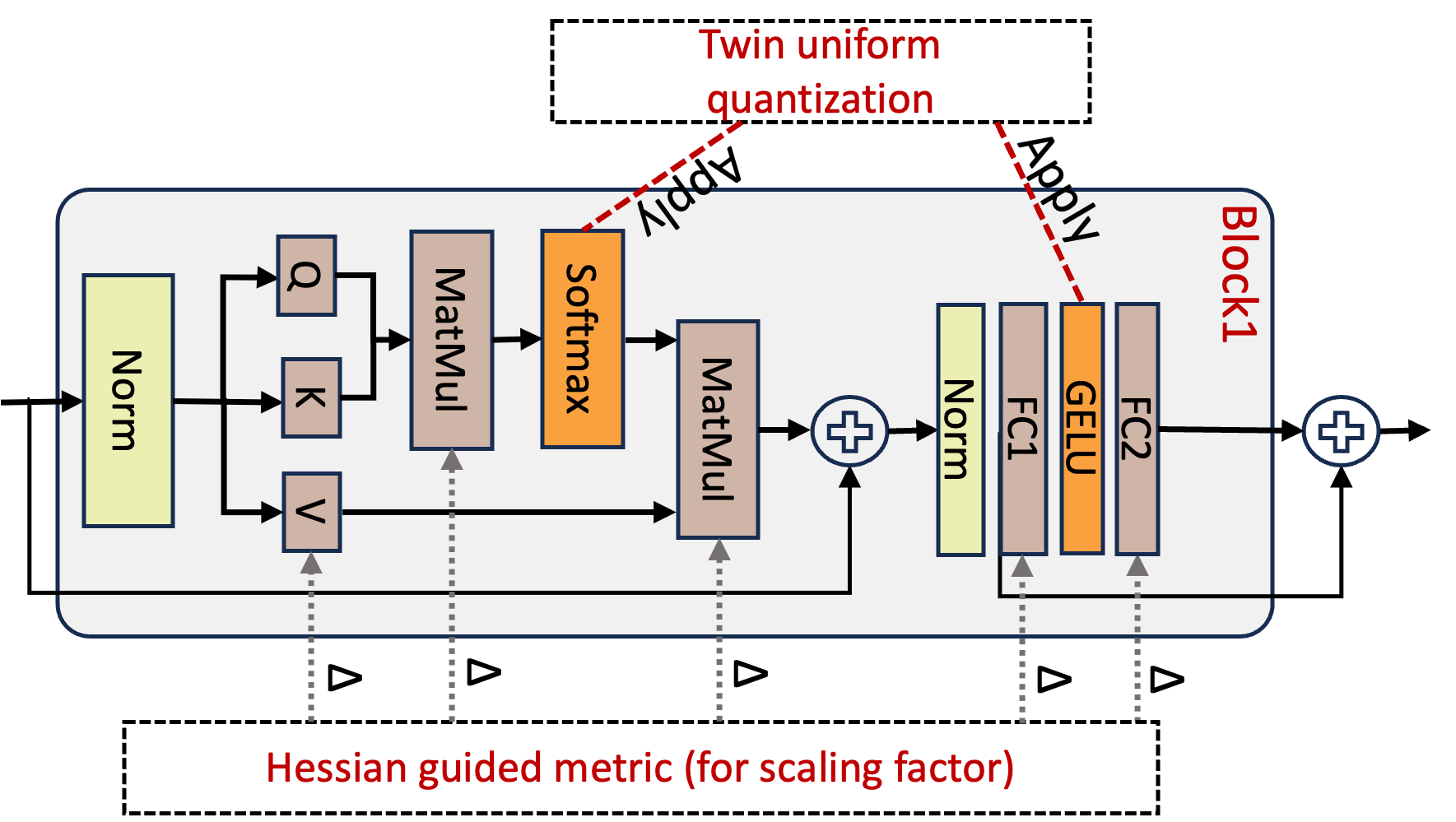}
    \caption{\highlight{An overview of the quantization framework in PTQ4ViT~\cite{yuan2022ptq4vit}, which integrates twin uniform quantization and a Hessian-guided metric for determining scaling factors. Building on this, APQ-ViT~\cite{ding2022towards} addresses the critical quantization error caused by Hessian-based scaling through a block-wise evaluation.}}
    \label{fig:ptq4vit}
\end{figure}
\subsubsection{Quantization Techniques for Vision Transformer} \label{vit_quanti}\hfill\\
Applying quantization in ViT models is quite new in the neural network sector. ViT consists of multiple layers, including a self-attention mechanism and a feedforward neural network. As data passes through these layers, different layers learn to focus on different features of the input data. Applying quantization to ViT can be challenging due to its complexity. Additionally, the loss from quantization can significantly impact the self-attention mechanisms, potentially reducing the model's overall performance. Quantization methods on ViT can be broadly categorized into two main approaches based on their reliance on training or finetuning: PTQ and QAT.
\noindent \paragraph{PTQ Techniques for Vision Transformer}\hfill \\In the current scenario, PTQ is widely used for ViT because it offers an efficient way to meet the computational requirements on edge without additional training or finetuning, making it ideal for resource-constrained deployments. PTQ works can be divided into two categories~\cite{niu2023improving,zheng2022leveraging}: statistic-based PTQ and learning-based PTQ. Statistic-based PTQ methods focus on finding optimal quantization parameters to reduce quantization errors. In contrast, learning-based PTQ methods involve finetuning both the model weights and quantization parameters for improved performance~\cite{lv2024ptq4sam}. \\

\noindent \textbf{Statistic-Based PTQ methods} Most of the current PTQ works on ViT follow. \textbf{statistic-based methods}. \highlight{As illustrated in Figure~\ref{fig:ptq4vit}}, PTQ4ViT~\cite{yuan2022ptq4vit} addressed two key issues with base-PTQ for ViTs: 1) Unbalanced distributions after softmax and asymmetric distributions after GELU. 2) Traditional metrics are ineffective for determining quantization parameters. To solve the first problem, the authors proposed twin uniform quantization, which quantized values into two separate ranges. Additionally, to solve the second problem, they introduced a Hessian-guided metric for improved accuracy instead of \highlight{MSE} and cosine distance. \highlight{While effective in general cases, the Hessian-guided metric struggles in ultra low-bit scenarios (e.g., 2-bit or 4-bit), often missing key quantization errors. APQ-ViT, proposed by Ding et al.~\cite{ding2022towards}, addresses this limitation through refined error modeling.} The authors first proposed a unified bottom-elimination block-wise calibration for extremely low-bit representation. This block-wise calibration scheme enables a more precise evaluation of quantization errors by focusing on block-level disturbances that impact the final output. \highlight{For the second challenge, they observed the "matthew-effect" in the softmax distribution, where smaller values shrink further, and larger values dominate. Existing quantizers often overlook this issue, leading to information loss from the larger values. To address this, the authors proposed matthew-effect preserving quantization (MPQ) for softmax to retain its power-law characteristics and ensure more balanced information retention during quantization.}\\ 

\noindent \textbf{\textit{\highlight{Quantization Error with Noisy Bias}}} Liu et al. proposed NoisyQuant~\cite{liu2023noisyquant}, where they focused on adding a noisy bias to each layer to modify the input activation distribution to reduce the quantization error. The noisy bias is a single vector sampled from a uniform distribution. The authors removed the impact of noisy bias after the activation-weight multiplication in the linear layer with a denoising bias so that the method could retrieve the correct output. Surprisingly, the experiment showed that adding a noisy bias improved top-1 accuracy compared to the PTQ4ViT~\cite{yuan2022ptq4vit} on the ViT-B, DeiT-B, and Swin-S model.\\
\begin{figure}[htb]
  \centering
  \includegraphics[scale=0.25]{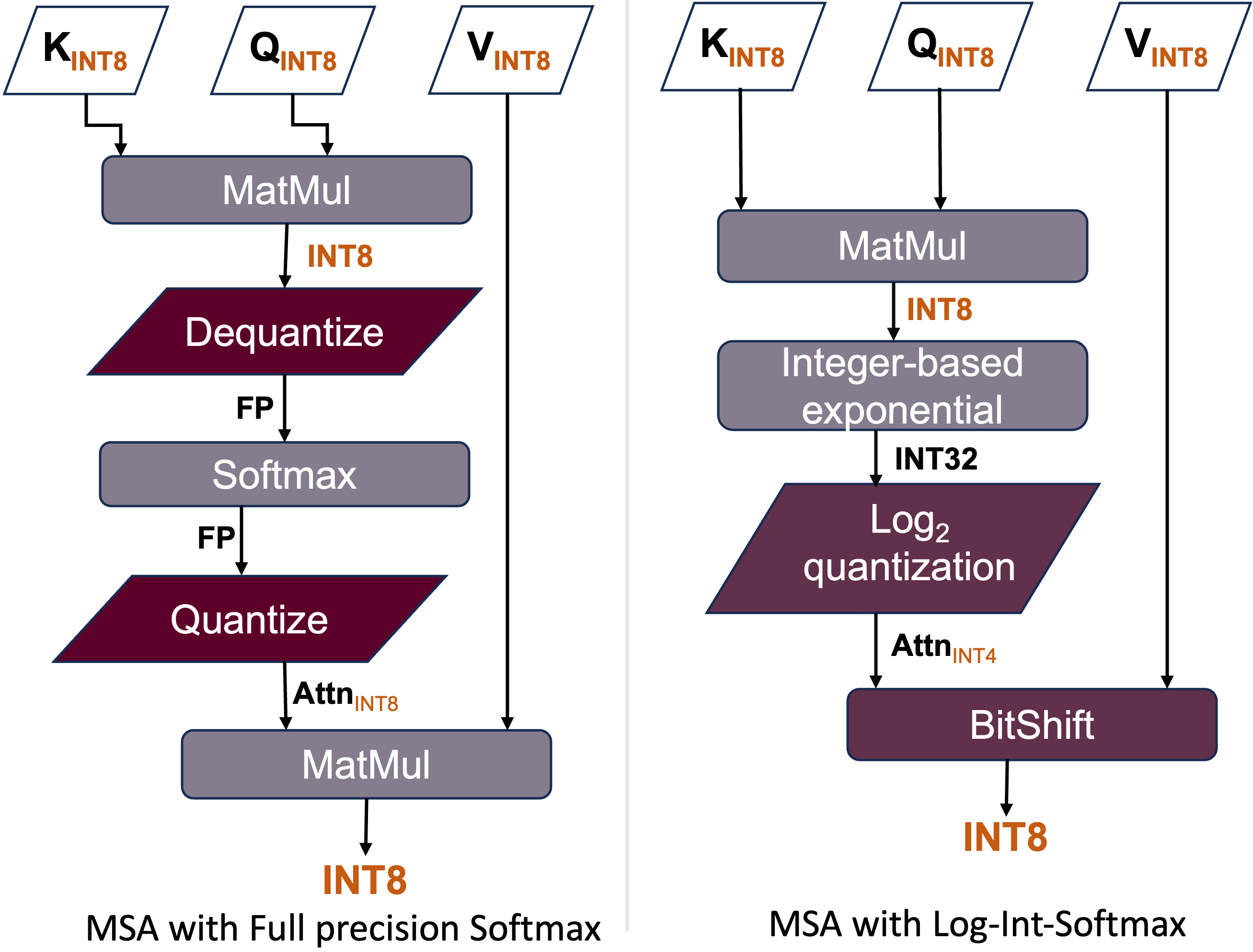}
  \caption{Comparison of using full precision Softmax and log-int-softmax in quantized MSA inference in FQ-ViT~\cite{lin2021fq}.}
  \label{fig:quanwsoft}
\end{figure}

\noindent \textbf{\highlight{\textit{Integer-only PTQ Methods}}} \highlight{Unlike earlier PTQ approaches that rely on FP32 precision (via dequantization) during inference, recent methods have introduced fully integer-only quantization for ViTs~\cite{lin2021fq,li2022vit,li2023repq}. These advancements eliminate floating-point dependencies, reduce computational overhead, and significantly enhance the suitability of ViTs for deployment on resource-constrained edge devices.} Lin et al. first introduced fully quantized PTQ techniques named  FQ-ViT~\cite{lin2021fq}, leveraging the power-of-two factor (PTF) method to minimize performance loss and inference complexity. To solve the non-uniform distribution in attention maps and avoid dequantizing to FP32 before softmax, they proposed that \highlight{log-int-softmax (LIS) be replaced by softmax. Additionally, the authors further streamlined inference by employing 4-bit quantization combined with bit-shift operations.} Figure~\ref{fig:quanwsoft} (left) illustrate\highlight{s} the traditional approach where the traditional approach dequantizes INT8 query (Q) and key (K) matrices to FP before softmax, re-quantizing afterward for attention computations. In contrast, the proposed method (Figure~\ref{fig:quanwsoft} (right)) introduced matrix multiplication followed by integer-based exponential (i-exp). The authors then utilized $Log_{2}$ quantization scale in the softmax function and converted the MatMul to BitShift between the quantized attention map and values (V). This \highlight{integer-only} workflow, including LIS in INT4 format, significantly reduces memory usage while maintaining precision. Extended from FQ-ViT, Li et al. introduced I-ViT~\cite{li2022vit}, the first integer-only PTQ framework for ViT, enabling inference entirely with integer arithmetic and bit-shifting, eliminating FP operations. In this framework, the authors utilized an integer-only pipeline named dyadic anthemic for non-linear functions such as dense layers. In contrast, non-linear functions, including softmax, GELU, and LayerNorm, were approximated with lightweight integer-based methods. The key contribution of this work is that Shiftmax and ShiftGEU replicated the behavior of their FP counterparts using integer bit-shifting. Despite I-ViT's reduction in bit-precision for parameters and its emphasis on integer-only inference, it retained its accuracy. For example, when I-ViT applied to DeiT-B, it achieved 81.74\% top-1 accuracy with 8-bit integer-only inference, outperforming I-BERT~\cite{kim2021bert} by 0.95\% (see Table~\ref{quantization_tech_classification}).\\
\begin{algorithm}[htb]
\small
\caption{\highlight{Fast Progressive Combining Search (FPCS)}}
\label{alg:fpcs}
\begin{algorithmic}[1]
\STATE \textbf{Input:} Coefficients $x, y, z_1, z_2, k, p$; a pretrained full-precision model; a set of calibration data $\mathcal{D}_{\text{calib}}$; and the $l$-th layer to be quantized $\phi_l$.
\STATE \textbf{Output:} Quantization hyperparameters $a^*, b^*$
\STATE \textcolor{blue}{\# The initialization step:}
\STATE Generate the raw input $X_l$ and output $O_l$ by $\phi_l$ based on $\mathcal{D}_{\text{calib}}$, and compute the percentiles $pct_0, pct_{0.1}, pct_{0.9}$ and $pct_1$ by \cite{li2019fully}.
\STATE Compute the uniform partition of the first and second hyperparameters as $\mathcal{A} = \{pct_{0.1} + i \cdot \tau_A | i = 0, \cdots, x\}$ and $\mathcal{B} = \{pct_{0.9} + j \cdot \tau_B | j = 0, \cdots, y\}$ with $\tau_A = (pct_0 - pct_{0.1}) / x$ and $\tau_B = (pct_1 - pct_{0.9}) / y$.
\STATE Generate the candidate set $C_0$ as the Cartesian product of $\mathcal{A}$ and $\mathcal{B}$: $C_0 = \mathcal{A} \times \mathcal{B}$.
\STATE \textcolor{blue}{\# The progressive searching step:}
\FOR{$i = 0, \cdots, p$}
  \STATE \textcolor{blue}{\# The coarse searching step:}
  \STATE Construct the subset $C' \subset C_i$ by selecting the partitions that have the top-$k$ smallest quantization loss.
  \STATE \textcolor{blue}{\# The expanding step:}
  \STATE Update the intervals for fine partitions: $\tau_A := \tau_A / (2 \cdot z_1)$, $\tau_B := \tau_B / (2 \cdot z_2)$.
  \STATE Update the candidate set with fine partitions: $C_{i+1} = \{(a + i \cdot \tau_A, b + j \cdot \tau_B) | (a, b) \in C', i = -z_1, \cdots, z_1; j = -z_2, \cdots, z_2\}$.
\ENDFOR
\STATE \textbf{return} $(a^*, b^*) \in C_p$ with the smallest quantization loss.
\end{algorithmic}
\end{algorithm}

\noindent \textbf{\highlight{\textit{Decoupling Quantization for Edge-Efficient Inference}}} The current studies consider quantizers and hardware standards \highlight{to be always} antagonistic, which is partially true. RepQ-ViT~\cite{li2023repq} decouples the quantization and inference process to explicitly bridge via scale reparameterization between these two steps. The authors applied channel-wise quantization for the post-LayerNorm activations to solve the inter-channel variations and $\log_{\sqrt{2}}$ quantization for the post-softmax activations. \highlight{During inference, the proposed method efficiently reparameterized layer-wise $\log_{2}$-based quantization to reduce the computational overhead associated with activation processing.} Using integer-only quantization for all layers lessened the computational cost dramatically and made them highly suitable for edge devices.\\

\noindent \textbf{\highlight{\textit{Mixed-precision PTQ methods}}} Recent studies have further advanced statistics-based PTQ for ViTs by incorporating \textbf{mixed-precision techniques}. Liu et al.~\cite{liu2021post} first explored a mixed precision PTQ scheme for ViT architectures to reduce the memory and computational requirements. The authors estimated optimal low-bit quantization intervals for weights and inputs, used ranking loss to preserve self-attention order, and analyzed layer-wise quantization loss to study mixed precision using the L1-norm~\cite{wu2018l1} of attention maps and outputs. Using calibration datasets from CIFAR-10~\cite{cifar10}, ImageNet-1k~\cite{5206848}, and COCO2017~\cite{lin2015microsoft}, their method outperformed percentile-based techniques\cite{li2019fully} by 3.35\% on CIFAR-10 with ViT-B model. Recently. Tai et al.~\cite{tai2024mptq} and Ranjan et al.~\cite{ranjan2024lrp} both extended the mixed precision PTQ techniques on ViT. MPTQ-ViT~\cite{tai2024mptq} utilized the smoothQuant~\cite{xiao2023smoothquant} with bias term (SQ-b) to address the asymmetry in activations, reducing clamping loss and improving quantization performance. The authors proposed a search-based scaling factor ratio (OPT-m) to determine the quantization parameters. Later, they incorporate SQ-b and OPT-m to propose greedy mixed precision PTQ techniques for ViT by allocating layer-wise bit-width. Additionally, Ranjan et al.~\cite{ranjan2024lrp} proposed LRP-QViT~\cite{ranjan2024lrp}, an explainability-based approach by assessing each layer's contribution to the model's predictions, guiding the assignment of mixed-precision bit allocations based on layer importance. The authors also clipped the channel-wise quantization to eliminate the outliers from post-LayerNorm activations, mitigating severe inter-channel variations and enhancing quantization robustness. Zhong et al. proposed ERQ~\cite{zhong2024erq} to mitigate the error arising during quantization from weight and activation quantization separately. The authors introduced activation quantization error reduction to reduce the activation error, which is like a ridge regression problem. The authors also proposed weight quantization error reduction in an interactive approach by rounding directions of quantized weight\highlight{s} coupled with a ridge regression solver.\\

\noindent \textbf{Learning-Based PTQ Methods} While most current PTQ methods for ViTs are statistic-based, there are only a few that utilize learning-based approaches. Existing PTQ methods for ViTs face challenges with inflexible quantization of post-softmax and post-GELU activations, which follow power-law-like distributions. \highlight{One of the major challenges in learning-based approaches is efficiently determining the optimal quantization parameters, such as scaling factors.} Adalog~\cite{wu2025adalog} introduced a fast progressive combining search (FPCS) strategy, which is designed to search the hyperparameter space effectively while maintaining linear complexity. \highlight{The FPCS algorithm comprises two stages: initialization and progressive search. The latter performs a coarse search to select top candidate pairs, followed by an expansion step that locally refines the search. The complete procedure is outlined in Algorithm~\ref{alg:fpcs}.} On the other hand, CLAMP-ViT~\cite{ramachandran2025clamp} leveraged \highlight{contrastive} learning in layer-wise evolutionary search to identify optimal quantization parameters. 

\begin{figure}[ht]
    \centering
    \begin{minipage}{0.42\textwidth}
        \centering
        \includegraphics[scale=0.5]{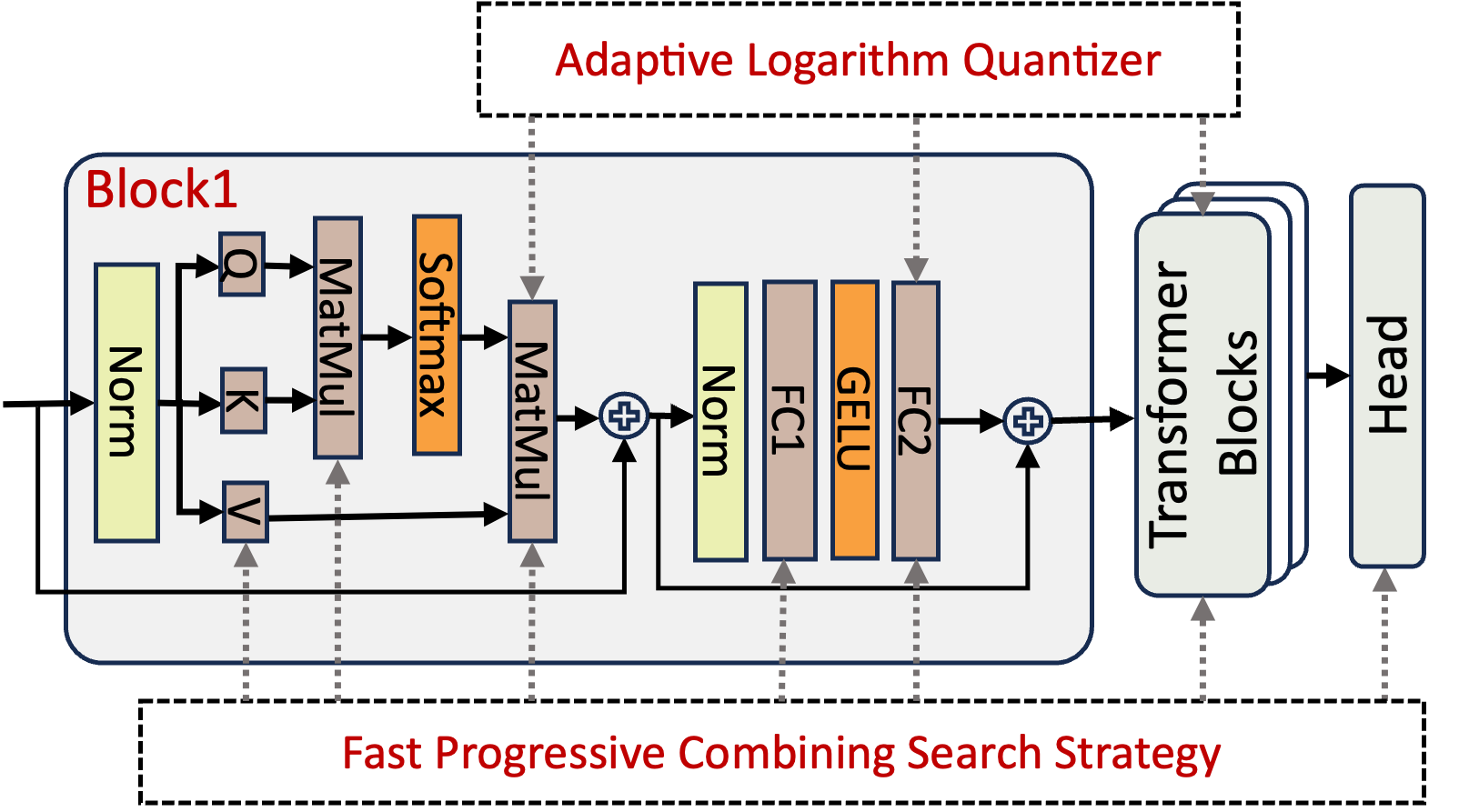}
        \caption{\highlight{The Overview of the AdaLog quantization method. AdaLog is applied to post-Softmax and post-GELU activations, with bias parameterization and a search strategy~\cite{wu2025adalog}.}}
        \label{fig:adalog}
    \end{minipage}
    \hfill
    \begin{minipage}{0.56\textwidth}
        
        \centering
        \includegraphics[scale=0.43]{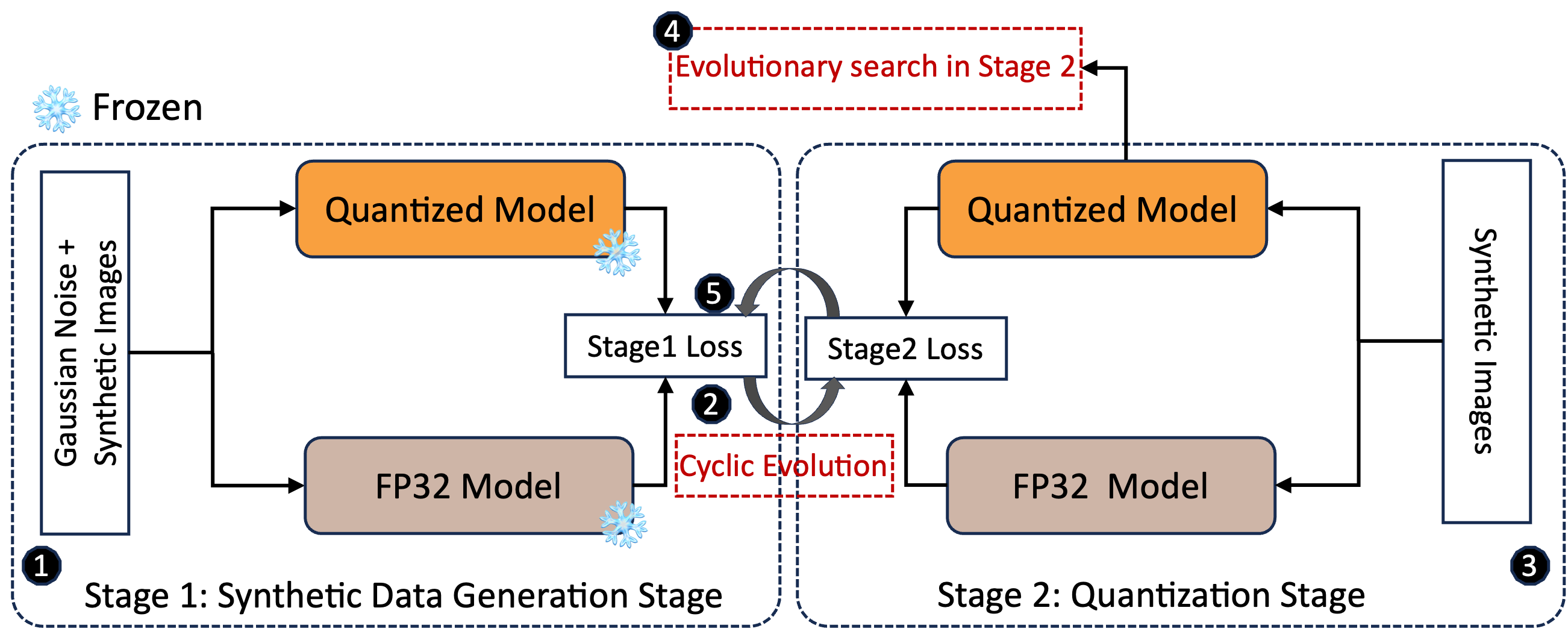}
        \caption{\highlight{The Overview of the Clamp-ViT~\cite{ramachandran2025clamp}. Stage~1 corresponds to steps \textcircled{1}–\textcircled{2}, Stage~2 corresponds to steps \textcircled{3}–\textcircled{4}, and the iterative refinement process is denoted by step \textcircled{5}.}}
        \label{fig:clampvit}
    \end{minipage}
\end{figure}

\highlight{As illustrated in Figure~\ref{fig:adalog}, Adalog explored the adaptive $\log_{2}$ and $\log_{\sqrt{2}}$ to better align} with the power-law distribution of activations while ensuring hardware-friendly quantization. The adaptive logarithm quantizers are applied to post-softmax and post-GELU activations through bias reparameterization. CLAMP-ViT~\cite{ramachandran2025clamp} adopted a two-stage approach between data generation and model quantization. \highlight{As illustrated in Figure~\ref{fig:clampvit}, stage 1 is to generate semantically rich and meaningful images. The stage 1 loss is used to adaptively refine the synthetic dataset in a self-supervised manner, forming a feedback loop with the generator. The refined synthetic images generated in Stage 1 are passed to Stage 2. Here, an evolutionary search is conducted on the quantized model to identify optimal layer-wise quantization parameters (e.g., bit-widths, scale factors). This search aims to minimize the Stage 2 loss, which reflects the fitness of quantized representations compared to the FP32 model. However, learning-based PTQ remains largely unexplored for low-bit quantization. Incorporating NAS to discover optimal quantization parameters automatically presents a promising future research direction instead of an evolutionary search.}

\noindent \paragraph{QAT Techniques for Vision Transformer}\hfill \\ Compared to PTQ techniques, QAT methods for ViTs remain relatively underexplored. Existing QAT approaches can be broadly classified into two categories: leveraging KD to optimize the quantized model and standalone independent frameworks. \\
\renewcommand{\arraystretch}{1.15}
\begin{table}[]
\centering
\caption{\highlight{Results of different post-training quantization(classification) techniques proposed for ViTs. \textbf{MP} denotes mixed precision; \textbf{W-bit} refers to weight bit-widths and \textbf{A-bit} refers to activation bit-widths. Here, Baseline refers to the closest comparable results for classification tasks.}}
\label{quantization_tech_classification}
\resizebox{\columnwidth}{!}{%
\begin{tabular}{c|c|c|c|c|cccc}
\hline
\multirow{2}{*}{\textbf{Algorithm}} &
  \multirow{2}{*}{\textbf{Code}} &
  \multirow{2}{*}{\textbf{Key point}} &
  \multirow{2}{*}{\textbf{Backbone}} &
  \multirow{2}{*}{\textbf{Dataset}} &
  \multicolumn{4}{c}{\textbf{Results}} \\ \cline{6-9} 
 &
   &
   &
   &
   &
  \multicolumn{1}{c|}{\textbf{Baseline}} &
  \multicolumn{1}{c|}{\textbf{W-bit}} &
  \multicolumn{1}{c|}{\textbf{A-bit}} &
  \textbf{\begin{tabular}[c]{@{}c@{}}Top-1\\ accuracy\end{tabular}} \\ \hline
\multirow{7}{*}{\begin{tabular}[c]{@{}c@{}}PTQ\\ ~\cite{liu2021post}\end{tabular}} &
  \multirow{7}{*}{\xmark} &
  \multirow{7}{*}{\begin{tabular}[c]{@{}c@{}}Introduce a mixed-precision \\ weights strategy\end{tabular}} &
  \multirow{6}{*}{ViT-B} &
  \multirow{2}{*}{CIFAR-10} &
  \multicolumn{1}{c|}{\multirow{2}{*}{Percentile}} &
  \multicolumn{1}{c|}{6 MP} &
  \multicolumn{1}{c|}{6 MP} &
  96.83 ($+3.35$) \\ \cline{7-9} 
 &
   &
   &
   &
   &
  \multicolumn{1}{c|}{} &
  \multicolumn{1}{c|}{8 MP} &
  \multicolumn{1}{c|}{8 MP} &
  97.79 ($+2.03$) \\ \cline{5-9} 
 &
   &
   &
   &
  \multirow{2}{*}{CIFAR-100} &
  \multicolumn{1}{c|}{\multirow{2}{*}{Percentile}} &
  \multicolumn{1}{c|}{6 MP} &
  \multicolumn{1}{c|}{6 MP} &
  83.99 ($+3.15$) \\ \cline{7-9} 
 &
   &
   &
   &
   &
  \multicolumn{1}{c|}{} &
  \multicolumn{1}{c|}{8 MP} &
  \multicolumn{1}{c|}{8 MP} &
  85.76 ($+2.48$) \\ \cline{5-9} 
 &
   &
   &
   &
  \multirow{2}{*}{\begin{tabular}[c]{@{}c@{}}ImageNet-1k\\ ~\cite{5206848}\end{tabular}} &
  \multicolumn{1}{c|}{\multirow{2}{*}{Percentile}} &
  \multicolumn{1}{c|}{6 MP} &
  \multicolumn{1}{c|}{6 MP} &
  75.26 ($+3.68$) \\ \cline{7-9} 
 &
   &
   &
   &
   &
  \multicolumn{1}{c|}{} &
  \multicolumn{1}{c|}{8 MP} &
  \multicolumn{1}{c|}{8 MP} &
  76.98 ($+2.88$) \\ \cline{4-9} 
 &
   &
   &
  DeiT-B &
  \begin{tabular}[c]{@{}c@{}}ImageNet-1k\\ ~\cite{5206848}\end{tabular} &
  \multicolumn{1}{c|}{Bit-Split} &
  \multicolumn{1}{c|}{6 MP} &
  \multicolumn{1}{c|}{6 MP} &
  74.58 ($+0.54$) \\ \hline
\multirow{4}{*}{\begin{tabular}[c]{@{}c@{}}PTQ4ViT\\ ~\cite{yuan2022ptq4vit}\end{tabular}} &
  \multirow{4}{*}{\cmark} &
  \multirow{4}{*}{\begin{tabular}[c]{@{}c@{}}Twin-uniform \\ quantization with \\ Hessian-guided calibration\end{tabular}} &
  ViT-B &
  \multirow{4}{*}{\begin{tabular}[c]{@{}c@{}}ImageNet-1k\\ ~\cite{5206848}\end{tabular}} &
  \multicolumn{1}{c|}{\multirow{4}{*}{Base-PTQ}} &
  \multicolumn{1}{c|}{8} &
  \multicolumn{1}{c|}{8} &
  85.82 ($+0.52$) \\ \cline{4-4} \cline{7-9} 
 &
   &
   &
  DeiT-B &
   &
  \multicolumn{1}{c|}{} &
  \multicolumn{1}{c|}{8} &
  \multicolumn{1}{c|}{8} &
  82.97 ($+.64$) \\ \cline{4-4} \cline{7-9} 
 &
   &
   &
  \multirow{2}{*}{Swin-B} &
   &
  \multicolumn{1}{c|}{} &
  \multicolumn{1}{c|}{\multirow{2}{*}{8}} &
  \multicolumn{1}{c|}{\multirow{2}{*}{8}} &
  \multirow{2}{*}{86.39 ($+0.23$)} \\
 &
   &
   &
   &
   &
  \multicolumn{1}{c|}{} &
  \multicolumn{1}{c|}{} &
  \multicolumn{1}{c|}{} &
   \\ \hline
\multirow{6}{*}{\begin{tabular}[c]{@{}c@{}}APQ-ViT\\ ~\cite{ding2022towards}\end{tabular}} &
  \multirow{6}{*}{\xmark} &
  \multirow{6}{*}{\begin{tabular}[c]{@{}c@{}}Propose a Matthew-effect \\ preserving scheme for low\\  bit quantization\end{tabular}} &
  \multirow{2}{*}{ViT-B} &
  \multirow{6}{*}{\begin{tabular}[c]{@{}c@{}}ImageNet-1k\\ ~\cite{5206848}\end{tabular}} &
  \multicolumn{1}{c|}{\multirow{6}{*}{\begin{tabular}[c]{@{}c@{}}PTQ4ViT\\ ~\cite{yuan2022ptq4vit}\end{tabular}}} &
  \multicolumn{1}{c|}{6} &
  \multicolumn{1}{c|}{6} &
  82.21 ($+0.56$) \\ \cline{7-9} 
 &
   &
   &
   &
   &
  \multicolumn{1}{c|}{} &
  \multicolumn{1}{c|}{4} &
  \multicolumn{1}{c|}{4} &
  41.41 ($+10.72$) \\ \cline{4-4} \cline{7-9} 
 &
   &
   &
  \multirow{2}{*}{Swin-B/384} &
   &
  \multicolumn{1}{c|}{} &
  \multicolumn{1}{c|}{6} &
  \multicolumn{1}{c|}{6} &
  85.60 ($+0.16$) \\ \cline{7-9} 
 &
   &
   &
   &
   &
  \multicolumn{1}{c|}{} &
  \multicolumn{1}{c|}{4} &
  \multicolumn{1}{c|}{4} &
  80.84 ($+2.0$) \\ \cline{4-4} \cline{7-9} 
 &
   &
   &
  \multirow{2}{*}{DeiT-B} &
   &
  \multicolumn{1}{c|}{} &
  \multicolumn{1}{c|}{6} &
  \multicolumn{1}{c|}{6} &
  80.42 ($+0.17$) \\ \cline{7-9} 
 &
   &
   &
   &
   &
  \multicolumn{1}{c|}{} &
  \multicolumn{1}{c|}{4} &
  \multicolumn{1}{c|}{4} &
  67.48 ($+3.09$) \\ \hline
\multirow{6}{*}{\begin{tabular}[c]{@{}c@{}}NoisyQuant\\ ~\cite{liu2023noisyquant}\end{tabular}} &
  \multirow{6}{*}{\cmark} &
  \multirow{6}{*}{\begin{tabular}[c]{@{}c@{}}Add a fixed noisy factor \\ to the date distribution\end{tabular}} &
  \multirow{2}{*}{ViT-B} &
  \multirow{6}{*}{\begin{tabular}[c]{@{}c@{}}ImageNet-1k\\ ~\cite{5206848}\end{tabular}} &
  \multicolumn{1}{c|}{\multirow{6}{*}{\begin{tabular}[c]{@{}c@{}}PTQ4ViT\\ ~\cite{yuan2022ptq4vit}\end{tabular}}} &
  \multicolumn{1}{c|}{6} &
  \multicolumn{1}{c|}{6} &
  81.90 ($+6.24$) \\ \cline{7-9} 
 &
   &
   &
   &
   &
  \multicolumn{1}{c|}{} &
  \multicolumn{1}{c|}{8} &
  \multicolumn{1}{c|}{8} &
  84.10 ($+0.71$) \\ \cline{4-4} \cline{7-9} 
 &
   &
   &
  \multirow{2}{*}{DeiT-B} &
   &
  \multicolumn{1}{c|}{} &
  \multicolumn{1}{c|}{6} &
  \multicolumn{1}{c|}{6} &
  79.77 ($+.99$) \\ \cline{7-9} 
 &
   &
   &
   &
   &
  \multicolumn{1}{c|}{} &
  \multicolumn{1}{c|}{8} &
  \multicolumn{1}{c|}{8} &
  81.30 ($+0.36$) \\ \cline{4-4} \cline{7-9} 
 &
   &
   &
  \multirow{2}{*}{Swin-S} &
   &
  \multicolumn{1}{c|}{} &
  \multicolumn{1}{c|}{6} &
  \multicolumn{1}{c|}{6} &
  84.57 ($+1.22$) \\ \cline{7-9} 
 &
   &
   &
   &
   &
  \multicolumn{1}{c|}{} &
  \multicolumn{1}{c|}{8} &
  \multicolumn{1}{c|}{8} &
  85.11 ($+0.32$) \\ \hline
\multirow{4}{*}{\begin{tabular}[c]{@{}c@{}}FQ-Vit\\ ~\cite{lin2021fq}\end{tabular}} &
  \multirow{4}{*}{\cmark} &
  \multirow{4}{*}{\begin{tabular}[c]{@{}c@{}}Power-of-two for LayerNorm \\ \& replace Softmax with\\  Log-Int-Softmax\end{tabular}} &
  DeiT-T &
  \multirow{4}{*}{\begin{tabular}[c]{@{}c@{}}ImageNet-1k\\ ~\cite{5206848}\end{tabular}} &
  \multicolumn{1}{c|}{\multirow{4}{*}{Percentile}} &
  \multicolumn{1}{c|}{8} &
  \multicolumn{1}{c|}{8} &
  71.61 ($+0.14$) \\ \cline{4-4} \cline{7-9} 
 &
   &
   &
  DeiT-S &
   &
  \multicolumn{1}{c|}{} &
  \multicolumn{1}{c|}{8} &
  \multicolumn{1}{c|}{8} &
  79.17 ($+2.6$) \\ \cline{4-4} \cline{7-9} 
 &
   &
   &
  DeiT-B &
   &
  \multicolumn{1}{c|}{} &
  \multicolumn{1}{c|}{8} &
  \multicolumn{1}{c|}{8} &
  81.20 ($+1.83$) \\ \cline{4-4} \cline{7-9} 
 &
   &
   &
  Swin-B &
   &
  \multicolumn{1}{c|}{} &
  \multicolumn{1}{c|}{8} &
  \multicolumn{1}{c|}{8} &
  82.97 ($+42.04$) \\ \hline
\multirow{4}{*}{\begin{tabular}[c]{@{}c@{}}I-ViT\\ ~\cite{li2022vit}\end{tabular}} &
  \multirow{4}{*}{\cmark} &
  \multirow{4}{*}{Integer-only inference} &
  ViT-B &
  \multirow{4}{*}{\begin{tabular}[c]{@{}c@{}}ImageNet-1k\\ ~\cite{5206848}\end{tabular}} &
  \multicolumn{1}{c|}{\multirow{4}{*}{\begin{tabular}[c]{@{}c@{}}I-BERT\\ ~\cite{kim2021bert}\end{tabular}}} &
  \multicolumn{1}{c|}{-} &
  \multicolumn{1}{c|}{-} &
  84.76 ($+1.06$) \\ \cline{4-4} \cline{7-9} 
 &
   &
   &
  DeiT-B &
   &
  \multicolumn{1}{c|}{} &
  \multicolumn{1}{c|}{-} &
  \multicolumn{1}{c|}{-} &
  81.74 ($+0.95$) \\ \cline{4-4} \cline{7-9} 
 &
   &
   &
  \multirow{2}{*}{Swin-S} &
   &
  \multicolumn{1}{c|}{} &
  \multicolumn{1}{c|}{\multirow{2}{*}{-}} &
  \multicolumn{1}{c|}{\multirow{2}{*}{-}} &
  \multirow{2}{*}{83.01 ($+1.15$)} \\
 &
   &
   &
   &
   &
  \multicolumn{1}{c|}{} &
  \multicolumn{1}{c|}{} &
  \multicolumn{1}{c|}{} &
   \\ \hline
\multirow{3}{*}{\begin{tabular}[c]{@{}c@{}}RepQ-ViT\\ ~\cite{li2023repq}\end{tabular}} &
  \multirow{3}{*}{\cmark} &
  \multirow{3}{*}{\begin{tabular}[c]{@{}c@{}}$\log_{\sqrt{2}}$ reparametrized to $\log_{2}$\\  during quantization \& inference\end{tabular}} &
  ViT-B &
  \multirow{3}{*}{\begin{tabular}[c]{@{}c@{}}ImageNet-1k\\ ~\cite{5206848}\end{tabular}} &
  \multicolumn{1}{c|}{\multirow{3}{*}{\begin{tabular}[c]{@{}c@{}}APQ-ViT\\ ~\cite{ding2022towards}\end{tabular}}} &
  \multicolumn{1}{c|}{4} &
  \multicolumn{1}{c|}{4} &
  83.62 ($+1.41$) \\ \cline{4-4} \cline{7-9} 
 &
   &
   &
  DeiT-B &
   &
  \multicolumn{1}{c|}{} &
  \multicolumn{1}{c|}{4} &
  \multicolumn{1}{c|}{4} &
  81.27 ($+0.85$) \\ \cline{4-4} \cline{7-9} 
 &
   &
   &
  Swin-S &
   &
  \multicolumn{1}{c|}{} &
  \multicolumn{1}{c|}{4} &
  \multicolumn{1}{c|}{4} &
  82.79 ($+0.12$) \\ \hline
\multirow{2}{*}{\begin{tabular}[c]{@{}c@{}}MPTQ-ViT\\ ~\cite{tai2024mptq}\end{tabular}} &
  \multirow{2}{*}{\xmark} &
  \multirow{2}{*}{\begin{tabular}[c]{@{}c@{}}Introduce SmoothQuant~\cite{xiao2023smoothquant} \\ with bias term\end{tabular}} &
  ViT-B &
  \multirow{2}{*}{\begin{tabular}[c]{@{}c@{}}ImageNet-1k\\ ~\cite{5206848}\end{tabular}} &
  \multicolumn{1}{c|}{\multirow{2}{*}{\begin{tabular}[c]{@{}c@{}}TSPTQ-ViT\\ ~\cite{10096817}\end{tabular}}} &
  \multicolumn{1}{c|}{6} &
  \multicolumn{1}{c|}{6} &
  82.70 ($+0.41$) \\ \cline{4-4} \cline{7-9} 
 &
   &
   &
  DeiT-B &
   &
  \multicolumn{1}{c|}{} &
  \multicolumn{1}{c|}{6} &
  \multicolumn{1}{c|}{6} &
  81.25 ($+0.64$) \\ \hline
\multirow{3}{*}{\begin{tabular}[c]{@{}c@{}}LRP-QViT\\ ~\cite{ranjan2024lrp}\end{tabular}} &
  \multirow{3}{*}{\xmark} &
  \multirow{3}{*}{\begin{tabular}[c]{@{}c@{}}Importance-based layer\\  bit precision\end{tabular}} &
  ViT-B &
  \multirow{3}{*}{\begin{tabular}[c]{@{}c@{}}ImageNet-1k\\ ~\cite{5206848}\end{tabular}} &
  \multicolumn{1}{c|}{\multirow{3}{*}{\begin{tabular}[c]{@{}c@{}}RepQ-ViT\\ ~\cite{li2023repq}\end{tabular}}} &
  \multicolumn{1}{c|}{6 MP} &
  \multicolumn{1}{c|}{6 MP} &
  83.87 ($+0.25$) \\ \cline{4-4} \cline{7-9} 
 &
   &
   &
  DeiT-B &
   &
  \multicolumn{1}{c|}{} &
  \multicolumn{1}{c|}{6 MP} &
  \multicolumn{1}{c|}{6 MP} &
  81.44 ($+0.17$) \\ \cline{4-4} \cline{7-9} 
 &
   &
   &
  Swin-S &
   &
  \multicolumn{1}{c|}{} &
  \multicolumn{1}{c|}{6 MP} &
  \multicolumn{1}{c|}{6 MP} &
  82.86 ($+0.07$) \\ \hline
\multirow{3}{*}{\begin{tabular}[c]{@{}c@{}}ERQ\\ ~\cite{zhong2024erq}\end{tabular}} &
  \multirow{3}{*}{\cmark} &
  \multirow{3}{*}{\begin{tabular}[c]{@{}c@{}}Introduced weight quantization \\ error reduction metrics\end{tabular}} &
  ViT-B &
  \multirow{3}{*}{\begin{tabular}[c]{@{}c@{}}ImageNet-1k\\ ~\cite{5206848}\end{tabular}} &
  \multicolumn{1}{c|}{\multirow{3}{*}{\begin{tabular}[c]{@{}c@{}}AdaRound\\ ~\cite{nagel2020up}\end{tabular}}} &
  \multicolumn{1}{c|}{5} &
  \multicolumn{1}{c|}{5} &
  82.81 ($+0.81$) \\ \cline{4-4} \cline{7-9} 
 &
   &
   &
  DeiT-B &
   &
  \multicolumn{1}{c|}{} &
  \multicolumn{1}{c|}{5} &
  \multicolumn{1}{c|}{5} &
  80.65 ($+0.47$) \\ \cline{4-4} \cline{7-9} 
 &
   &
   &
  Swin-S &
   &
  \multicolumn{1}{c|}{} &
  \multicolumn{1}{c|}{5} &
  \multicolumn{1}{c|}{5} &
  82.44 ($+0.32$) \\ \hline
\multirow{3}{*}{\begin{tabular}[c]{@{}c@{}}Adalog\\ ~\cite{wu2025adalog}\end{tabular}} &
  \multirow{3}{*}{\cmark} &
  \multirow{3}{*}{\begin{tabular}[c]{@{}c@{}}Proposed adaptive log-based \\ quantization for post-Softmax\\ \& post GELU activations\end{tabular}} &
  ViT-B &
  \multirow{3}{*}{\begin{tabular}[c]{@{}c@{}}ImageNet-1k\\ ~\cite{5206848}\end{tabular}} &
  \multicolumn{1}{c|}{\multirow{3}{*}{\begin{tabular}[c]{@{}c@{}}RepQ-ViT\\ ~\cite{li2023repq}\end{tabular}}} &
  \multicolumn{1}{c|}{6} &
  \multicolumn{1}{c|}{6} &
  84.80($+1.18$) \\ \cline{4-4} \cline{7-9} 
 &
   &
   &
  DeiT-B &
   &
  \multicolumn{1}{c|}{} &
  \multicolumn{1}{c|}{6} &
  \multicolumn{1}{c|}{6} &
  81.55 ($+0.28$) \\ \cline{4-4} \cline{7-9} 
 &
   &
   &
  Swin-S &
   &
  \multicolumn{1}{c|}{} &
  \multicolumn{1}{c|}{6} &
  \multicolumn{1}{c|}{6} &
  83.19 ($+0.40$) \\ \hline
\multirow{2}{*}{\begin{tabular}[c]{@{}c@{}}CLAMP-ViT\\ ~\cite{ramachandran2025clamp}\end{tabular}} &
  \multirow{2}{*}{\cmark} &
  \multirow{2}{*}{\begin{tabular}[c]{@{}c@{}}Evolutionary search for \\ optimal quantization parameters\end{tabular}} &
  DeiT-S &
  \multirow{2}{*}{\begin{tabular}[c]{@{}c@{}}ImageNet-1k\\ ~\cite{5206848}\end{tabular}} &
  \multicolumn{1}{c|}{\multirow{2}{*}{\begin{tabular}[c]{@{}c@{}}LRP-QViT\\ ~\cite{ranjan2024lrp}\end{tabular}}} &
  \multicolumn{1}{c|}{6} &
  \multicolumn{1}{c|}{6} &
  79.43 ($+0.40$) \\ \cline{4-4} \cline{7-9} 
 &
   &
   &
  Swin-S &
   &
  \multicolumn{1}{c|}{} &
  \multicolumn{1}{c|}{6} &
  \multicolumn{1}{c|}{6} &
  82.86 ($+0.00$) \\ \hline
\end{tabular}%
}
\end{table}

\noindent \textbf{Leveraging KD in QAT} Q-Vit~\cite{li2022q}  first proposed an information rectification module based on information theory to resolve the convergence issue during joint training of quantization. The authors then proposed distributed guided distillation by taking appropriate activities and utilizing the knowledge from similar matrices in distillation to perform the optimization perfectly. However, Q-ViT lacks other CV tasks, such as object detection. Another recent work, Q-DETR~\cite{xu2023q}, is introduced to solve the information distortion problem. The authors explored the low-bit quantization of DETR and proposed a bi-level optimization framework based on the information bottleneck principle. However, Q-DETR failed to keep the attention activations less than 4 bits and resulted in mixed-precision quantization, which is hardware-inefficient in the current scenario. Both Q-ViT and Q-DETR explored the lightweight version of DETR apart from modifying the \highlight{MSA}. AQ-DETR~\cite{aqvit} focused on solving the problem that exists for low bits of DETR in previous studies. The authors introduced an auxiliary query module and a layer-by-layer distillation module to reduce the quantization error between the quantized attention and its full-precision counterpart. All the previously discussed works are heavily dependent on the data. Li et al. ~\cite{li2023psaq} proposed PSAQ-ViT, aiming to achieve a data-free quantization framework by utilizing the property of KD. The authors introduced an adaptive teacher-student strategy enabling cyclic interaction between generated samples and the quantized model under the supervision of the full-precision model, significantly improving accuracy. The framework leverages task- and model-independent prior information, making it universal across various vision tasks such as classification and object detection.\\
\renewcommand{\arraystretch}{1.2}
\begin{table}[]
\centering
\caption{Results of different quantization (object detection) techniques proposed for ViTs. The algorithms are experimented on COCO 2017~\cite{lin2014microsoft} datasets for object detection tasks. Here, Baseline refers to the closest comparable results for object detection. \textbf{MP} denotes
mixed precision.}
\label{tab:quantization_object}
\resizebox{0.9\columnwidth}{!}{%
\begin{tabular}{c|c|ccccc}
\hline
\multirow{2}{*}{\textbf{Algorithm}} &
  \multirow{2}{*}{\textbf{Backbone}} &
  \multicolumn{5}{c}{\textbf{Results}} \\ \cline{3-7} 
 &
   &
  \multicolumn{1}{c|}{\textbf{Baseline}} &
  \multicolumn{1}{c|}{\textbf{W-bit}} &
  \multicolumn{1}{c|}{\textbf{A-bit}} &
  \multicolumn{1}{c|}{\textbf{mAP}} &
  \textbf{AP\textsuperscript{box}} \\ \hline
\multirow{2}{*}{PTQ~\cite{liu2021post}} &
  \multirow{2}{*}{DETR} &
  \multicolumn{1}{c|}{\multirow{2}{*}{Easyquant~\cite{wu2020easyquant}}} &
  \multicolumn{1}{c|}{6 MP} &
  \multicolumn{1}{c|}{6 MP} &
  \multicolumn{1}{c|}{40.5($+1.5$)} &
  - \\
 &
   &
  \multicolumn{1}{c|}{} &
  \multicolumn{1}{c|}{8 MP} &
  \multicolumn{1}{c|}{8 MP} &
  \multicolumn{1}{c|}{41.7($+1.3$)} &
  - \\ \hline
\multirow{2}{*}{APQ-ViT~\cite{ding2022towards}} &
  \multirow{2}{*}{Mask-RCNN+Swin-T} &
  \multicolumn{1}{c|}{\multirow{2}{*}{PTQ4ViT~\cite{yuan2022ptq4vit}}} &
  \multicolumn{1}{c|}{6} &
  \multicolumn{1}{c|}{6} &
  \multicolumn{1}{c|}{-} &
  45.4 ($+39.6$) \\
 &
   &
  \multicolumn{1}{c|}{} &
  \multicolumn{1}{c|}{4} &
  \multicolumn{1}{c|}{4} &
  \multicolumn{1}{c|}{-} &
  23.7 ($+16.8$) \\ \hline
FQ-ViT~\cite{lin2021fq} &
  Mask-RCNN+Swin-S &
  \multicolumn{1}{c|}{OMSE~\cite{choukroun2019low}} &
  \multicolumn{1}{c|}{8} &
  \multicolumn{1}{c|}{8} &
  \multicolumn{1}{c|}{47.8(+5.2)} &
  - \\ \hline
NoisyQuant~\cite{liu2023noisyquant} &
  DETR &
  \multicolumn{1}{c|}{PTQ~\cite{liu2021post}} &
  \multicolumn{1}{c|}{8} &
  \multicolumn{1}{c|}{8} &
  \multicolumn{1}{c|}{41.4($+0.2$)} &
  - \\ \hline
\multirow{2}{*}{RepQ-ViT~\cite{li2023repq}} &
  Mask-RCNN+Swin-S &
  \multicolumn{1}{c|}{\multirow{2}{*}{APQ-ViT~\cite{ding2022towards}}} &
  \multicolumn{1}{c|}{6} &
  \multicolumn{1}{c|}{6} &
  \multicolumn{1}{c|}{-} &
  47.8 ($+0.1$) \\ \cline{2-2} \cline{4-7} 
 &
  Cascade Mask-RCNN+Swin-S &
  \multicolumn{1}{c|}{} &
  \multicolumn{1}{c|}{6} &
  \multicolumn{1}{c|}{6} &
  \multicolumn{1}{c|}{-} &
  44.6 ($+0.1$) \\ \hline
\multirow{2}{*}{LRP-QViT~\cite{ranjan2024lrp}} &
  Mask-RCNN+Swin-S &
  \multicolumn{1}{c|}{\multirow{2}{*}{RepQ-ViT~\cite{li2023repq}}} &
  \multicolumn{1}{c|}{6 MP} &
  \multicolumn{1}{c|}{6 MP} &
  \multicolumn{1}{c|}{} &
  48.1 ($+0.3$) \\ \cline{2-2} \cline{4-7} 
 &
  Cascade Mask-RCNN+Swin-S &
  \multicolumn{1}{c|}{} &
  \multicolumn{1}{c|}{6 MP} &
  \multicolumn{1}{c|}{6 MP} &
  \multicolumn{1}{c|}{} &
  51.4 ($+0.0$) \\ \hline
\end{tabular}%
}

\end{table}

\noindent \textbf{Standalone QAT techniques} Although most works utilized the KD in QAT works, there are limited standalone studies without KD. PackQViT~\cite{dong2024packqvit} proposed activation-aware sub-8-bit QAT techniques for mobile devices. The authors leveraged $\log_{2}$ quantization or clamping to address the long-tailed distribution and outlier-aware training to handle the channel-wise outliers. Furthermore, the authors utilized int-2\textsuperscript{n}-softmax, int-LayerNorm, and int-GELU to enable integer-only computation. They designed a SIMD-based 4-bit packed multiplier to achieve end-to-end ViT acceleration on mobile devices. Another recent study named QD-BEV~\cite{zhang2023qd} explored QAT on BEVFormer~\cite{li2022bevformer} by leveraging 
image and BEV features. The authors identified that applying quantization directly in BEV tasks makes the training unstable, which leads to performance degradation. The authors proposed view-guided distillation to stabilize the QAT by conducting a systematic analysis of quantizing BEV networks. This work will open a new direction for autonomous vehicle research, applying QAT to reduce computational costs.

\subsubsection{Discussion}\hfill \\ Tables~\ref{quantization_tech_classification} and~\ref{tab:quantization_object} provide an overview of PTQ techniques for image classification and object detection on ViT models. A notable trend is the dominance of \highlight{PTQ} methods, with fewer studies exploring training-phase QAT. Table~\ref{quantization_tech_classification} highlights the top-1 accuracy improvements achieved by PTQ methods on ViT architectures for the classification tasks, while Table~\ref{tab:quantization_object} summarizes the mean average precision (mAP) and AP\textsuperscript{box} improvement achieved from the baseline techniques for the object detection task. Moreover, Table~\ref{quantization_qat_classification} summarizes QAT techniques for image classification and object detection. Interestingly, most QAT methods for ViTs, such as those leveraging KD~\cite{li2022q,xu2023q,aqvit,li2023psaq}, focus on optimizing quantized models but lack experimental validation on edge devices. In contrast, approaches like PackQViT~\cite{dong2024packqvit} have experimented with their QAT frameworks on mobile devices, pushing the boundaries of practical deployment. However, these quantization techniques have been broadly experimented on datasets like ImageNet-1k and COCO 2017, raising questions about their generalizability to specialized domains such as medical imaging or autonomous driving. This gap underscores the need for future research to explore versatile quantization strategies that cater to diverse application areas and resource-constrained edge devices.

\highlight{It is important to note that the impact of model compression techniques such as pruning, quantization, and KD extends beyond theoretical efficiency. These methods directly influence how ViTs can be deployed on edge devices, especially when paired with hardware-aware acceleration strategies discussed in Section~\ref{acce_tech}. Moreover, the practical effectiveness of these compression techniques often depends on the capabilities of software toolchains, which provide essential backend support for optimized deployment, as explored in Section~\ref{tools_for_edge}.}
\renewcommand{\arraystretch}{1.2}
\begin{table}[htb]
\centering
\caption{\highlight{Results of different QAT techniques proposed for ViTs. The results of QD-BEV~\cite{zhang2023qd} are the mean average precision (mAP) rather than AP\textsuperscript{box}.}}
\label{quantization_qat_classification}
\resizebox{\columnwidth}{!}{%
\begin{tabular}{c|c|c|c|cccccc}
\hline
\multirow{2}{*}{\textbf{Algorithm}} &
  \multirow{2}{*}{\textbf{Code}} &
  \multirow{2}{*}{\textbf{Key point}} &
  \multirow{2}{*}{\textbf{Dataset}} &
  \multicolumn{6}{c}{\textbf{Results}} \\ \cline{5-10} 
 &
   &
   &
   &
  \multicolumn{1}{c|}{\textbf{Baseline}} &
  \multicolumn{1}{c|}{\textbf{Model}} &
  \multicolumn{1}{c|}{\textbf{W-bit}} &
  \multicolumn{1}{c|}{\textbf{A-bit}} &
  \multicolumn{1}{c|}{\textbf{\begin{tabular}[c]{@{}c@{}}Top-1\\ (\%)\end{tabular}}} &
  \textbf{AP\textsuperscript{box}} \\ \hline
\multirow{4}{*}{\begin{tabular}[c]{@{}c@{}}Q-ViT\\ ~\cite{li2022q}\end{tabular}} &
  \multirow{4}{*}{\cmark} &
  \multirow{4}{*}{\begin{tabular}[c]{@{}c@{}}Enables 3-bit ViT \\ QAT.\end{tabular}} &
  \multirow{4}{*}{\begin{tabular}[c]{@{}c@{}}ImageNet-1k\\ ~\cite{5206848}\end{tabular}} &
  \multicolumn{1}{c|}{\multirow{4}{*}{\begin{tabular}[c]{@{}c@{}}LSQ\\ ~\cite{esser2019learned}\end{tabular}}} &
  \multicolumn{1}{c|}{\multirow{2}{*}{DeiT-B}} &
  \multicolumn{1}{c|}{4} &
  \multicolumn{1}{c|}{4} &
  \multicolumn{1}{c|}{83.0 ($+2.1$)} &
  - \\
 &
   &
   &
   &
  \multicolumn{1}{c|}{} &
  \multicolumn{1}{c|}{} &
  \multicolumn{1}{c|}{2} &
  \multicolumn{1}{c|}{2} &
  \multicolumn{1}{c|}{74.2 ($+3.9$)} &
  - \\ \cline{6-10} 
 &
   &
   &
   &
  \multicolumn{1}{c|}{} &
  \multicolumn{1}{c|}{\multirow{2}{*}{Swin-S}} &
  \multicolumn{1}{c|}{4} &
  \multicolumn{1}{c|}{4} &
  \multicolumn{1}{c|}{84.4 ($+1.9$)} &
  - \\
 &
   &
   &
   &
  \multicolumn{1}{c|}{} &
  \multicolumn{1}{c|}{} &
  \multicolumn{1}{c|}{2} &
  \multicolumn{1}{c|}{2} &
  \multicolumn{1}{c|}{76.9 ($+4.5$)} &
  - \\ \hline
\multirow{4}{*}{\begin{tabular}[c]{@{}c@{}}Q-DETR\\ ~\cite{xu2023q}\end{tabular}} &
  \multirow{4}{*}{\cmark} &
  \multirow{4}{*}{\begin{tabular}[c]{@{}c@{}}KD to enhance\\  quantized DETR's \\ representation\end{tabular}} &
  \multirow{2}{*}{\begin{tabular}[c]{@{}c@{}}PASCAL VOC\\ ~\cite{everingham2010pascal}\end{tabular}} &
  \multicolumn{1}{c|}{\multirow{4}{*}{\begin{tabular}[c]{@{}c@{}}LSQ\\ ~\cite{esser2019learned}\end{tabular}}} &
  \multicolumn{1}{c|}{TR50} &
  \multicolumn{1}{c|}{2} &
  \multicolumn{1}{c|}{2} &
  \multicolumn{1}{c|}{-} &
  50.7 ($+8.1$) \\ \cline{6-10} 
 &
   &
   &
   &
  \multicolumn{1}{c|}{} &
  \multicolumn{1}{c|}{SM-TR50} &
  \multicolumn{1}{c|}{2} &
  \multicolumn{1}{c|}{2} &
  \multicolumn{1}{c|}{-} &
  50.2 ($+7.9$) \\ \cline{4-4} \cline{6-10} 
 &
   &
   &
  \multirow{2}{*}{\begin{tabular}[c]{@{}c@{}}COCO 2017\\ ~\cite{lin2014microsoft}\end{tabular}} &
  \multicolumn{1}{c|}{} &
  \multicolumn{1}{c|}{TR50} &
  \multicolumn{1}{c|}{4} &
  \multicolumn{1}{c|}{4} &
  \multicolumn{1}{c|}{-} &
  39.4 ($+6.1$) \\ \cline{6-10} 
 &
   &
   &
   &
  \multicolumn{1}{c|}{} &
  \multicolumn{1}{c|}{SM-TR50} &
  \multicolumn{1}{c|}{4} &
  \multicolumn{1}{c|}{4} &
  \multicolumn{1}{c|}{-} &
  38.3 ($+4.4$) \\ \hline
\multirow{4}{*}{\begin{tabular}[c]{@{}c@{}}AQ-DETR\\ ~\cite{aqvit}\end{tabular}} &
  \multirow{4}{*}{\xmark} &
  \multirow{4}{*}{\begin{tabular}[c]{@{}c@{}}QAT  based on \\ auxiliary queries \\ for DETR\end{tabular}} &
  \multirow{2}{*}{\begin{tabular}[c]{@{}c@{}}PASCAL VOC\\ ~\cite{everingham2010pascal}\end{tabular}} &
  \multicolumn{1}{c|}{\multirow{4}{*}{\begin{tabular}[c]{@{}c@{}}Q-DETR\\ ~\cite{xu2023q}\end{tabular}}} &
  \multicolumn{1}{c|}{TR50} &
  \multicolumn{1}{c|}{4} &
  \multicolumn{1}{c|}{4} &
  \multicolumn{1}{c|}{-} &
  53.7 ($+3.3$) \\ \cline{6-10} 
 &
   &
   &
   &
  \multicolumn{1}{c|}{} &
  \multicolumn{1}{c|}{DE TR50} &
  \multicolumn{1}{c|}{4} &
  \multicolumn{1}{c|}{4} &
  \multicolumn{1}{c|}{-} &
  63.1 ($+2.0$) \\ \cline{4-4} \cline{6-10} 
 &
   &
   &
  \multirow{2}{*}{\begin{tabular}[c]{@{}c@{}}COCO 2017\\ ~\cite{lin2014microsoft}\end{tabular}} &
  \multicolumn{1}{c|}{} &
  \multicolumn{1}{c|}{TR50} &
  \multicolumn{1}{c|}{4} &
  \multicolumn{1}{c|}{4} &
  \multicolumn{1}{c|}{-} &
  40.2 ($+2.8$) \\ \cline{6-10} 
 &
   &
   &
   &
  \multicolumn{1}{c|}{} &
  \multicolumn{1}{c|}{DE TR50} &
  \multicolumn{1}{c|}{4} &
  \multicolumn{1}{c|}{4} &
  \multicolumn{1}{c|}{-} &
  44.1 ($+3.4$) \\ \hline
\multirow{4}{*}{\begin{tabular}[c]{@{}c@{}}PSAQ-ViT \\ V2~\cite{li2023psaq}\end{tabular}} &
  \multirow{4}{*}{\cmark} &
  \multirow{4}{*}{\begin{tabular}[c]{@{}c@{}}Data-free QAT \\ using adaptive KD\end{tabular}} &
  \multirow{2}{*}{\begin{tabular}[c]{@{}c@{}}ImageNet-1k\\ ~\cite{5206848}\end{tabular}} &
  \multicolumn{1}{c|}{\multirow{2}{*}{\begin{tabular}[c]{@{}c@{}}PSAQ-ViT\\ ~\cite{li2022psaqvit}\end{tabular}}} &
  \multicolumn{1}{c|}{DeiT-B} &
  \multicolumn{1}{c|}{8} &
  \multicolumn{1}{c|}{8} &
  \multicolumn{1}{c|}{81.5 ($+2.4$)} &
  - \\ \cline{6-10} 
 &
   &
   &
   &
  \multicolumn{1}{c|}{} &
  \multicolumn{1}{c|}{Swin-S} &
  \multicolumn{1}{c|}{8} &
  \multicolumn{1}{c|}{8} &
  \multicolumn{1}{c|}{82.1 ($+5.5$)} &
  - \\ \cline{4-10} 
 &
   &
   &
  \multirow{2}{*}{\begin{tabular}[c]{@{}c@{}}COCO 2017\\ ~\cite{lin2014microsoft}\end{tabular}} &
  \multicolumn{1}{c|}{\multirow{2}{*}{Standard V2}} &
  \multicolumn{1}{c|}{DC} &
  \multicolumn{1}{c|}{8} &
  \multicolumn{1}{c|}{8} &
  \multicolumn{1}{c|}{-} &
  44.8 ($+.3$) \\ \cline{6-10} 
 &
   &
   &
   &
  \multicolumn{1}{c|}{} &
  \multicolumn{1}{c|}{SC} &
  \multicolumn{1}{c|}{8} &
  \multicolumn{1}{c|}{8} &
  \multicolumn{1}{c|}{-} &
  50.9 ($+0.6$) \\ \hline
\multirow{3}{*}{\begin{tabular}[c]{@{}c@{}}PackQViT\\ ~\cite{dong2024packqvit}\end{tabular}} &
  \multirow{3}{*}{\cmark} &
  \multirow{3}{*}{\begin{tabular}[c]{@{}c@{}}8-bit QAT for \\ mobile devices\end{tabular}} &
  \multirow{2}{*}{\begin{tabular}[c]{@{}c@{}}ImageNet-1k\\ ~\cite{5206848}\end{tabular}} &
  \multicolumn{1}{c|}{\multirow{2}{*}{\begin{tabular}[c]{@{}c@{}}Q-ViT\\ ~\cite{li2022q}\end{tabular}}} &
  \multicolumn{1}{c|}{DeiT-B} &
  \multicolumn{1}{c|}{8} &
  \multicolumn{1}{c|}{8} &
  \multicolumn{1}{c|}{82.9 ($+0.5$)} &
  - \\ \cline{6-10} 
 &
   &
   &
   &
  \multicolumn{1}{c|}{} &
  \multicolumn{1}{c|}{Swin-S} &
  \multicolumn{1}{c|}{8} &
  \multicolumn{1}{c|}{8} &
  \multicolumn{1}{c|}{84.1 ($+0.5$)} &
  - \\ \cline{4-10} 
 &
   &
   &
  \begin{tabular}[c]{@{}c@{}}COCO 2017\\ ~\cite{lin2014microsoft}\end{tabular} &
  \multicolumn{1}{c|}{\begin{tabular}[c]{@{}c@{}}PTQ\\ ~\cite{liu2021post}\end{tabular}} &
  \multicolumn{1}{c|}{TR50} &
  \multicolumn{1}{c|}{8} &
  \multicolumn{1}{c|}{8} &
  \multicolumn{1}{c|}{-} &
  60.0 ($-3.1$) \\ \hline
\multirow{3}{*}{\begin{tabular}[c]{@{}c@{}}QD-BEV\\ ~\cite{zhang2023qd}\end{tabular}} &
  \multirow{3}{*}{\xmark} &
  \multirow{3}{*}{\begin{tabular}[c]{@{}c@{}}View-guided loss \\ image and \\ BEV features.\end{tabular}} &
  \multirow{3}{*}{\begin{tabular}[c]{@{}c@{}}NuScenes\\ ~\cite{caesar2020nuscenes}\end{tabular}} &
  \multicolumn{1}{c|}{\begin{tabular}[c]{@{}c@{}}BEV-DFQ\\ ~\cite{nagel2019data}\end{tabular}} &
  \multicolumn{1}{c|}{\multirow{3}{*}{BEV}} &
  \multicolumn{1}{c|}{8} &
  \multicolumn{1}{c|}{8} &
  \multicolumn{1}{c|}{-} &
  40.6 ($+2.2$) \\ \cline{5-5} \cline{7-10} 
 &
   &
   &
   &
  \multicolumn{1}{c|}{\begin{tabular}[c]{@{}c@{}}BEV-HAWQv3\\ ~\cite{yao2021hawq}\end{tabular}} &
  \multicolumn{1}{c|}{} &
  \multicolumn{1}{c|}{8} &
  \multicolumn{1}{c|}{8} &
  \multicolumn{1}{c|}{-} &
  40.6 ($+3.0$) \\ \cline{5-5} \cline{7-10} 
 &
   &
   &
   &
  \multicolumn{1}{c|}{\begin{tabular}[c]{@{}c@{}}BEV-PACT\\ ~\cite{choi2018pact}\end{tabular}} &
  \multicolumn{1}{c|}{} &
  \multicolumn{1}{c|}{8} &
  \multicolumn{1}{c|}{8} &
  \multicolumn{1}{c|}{-} &
  40.6 ($+3.2$) \\ \hline
\end{tabular}%
}

\begin{tablenotes}
\centering
\footnotesize
\item \textit{Note: Model names are shortened to save space in the table (SM TR50: SMCA-DeiT-R50).\\
\textbf{TR50:} DETR-R50;\quad \textbf{DE:} Deformable;\quad \textbf{DC:} DeiT-S with Cascade Mask R-CNN;\\
\textbf{SC:} Swin-S with Cascade Mask R-CNN;\quad \textbf{SM:} SMCA; \quad \textbf{BEV:} BEVFormer.}
\end{tablenotes}
\end{table}
\section{Tools for Efficient Edge Deployment}\label{tools_for_edge}
Efficient edge deployment of ViT requires a combination of software tools, evaluation tools, and advanced optimization techniques for different hardware architectures. Software tools streamline model deployment by providing optimized libraries and frameworks tailored for edge environments. Optimization techniques, such as memory optimization and pipeline parallelism, enhance performance by leveraging hardware-specific optimizations. Finally, heterogeneous platforms, including CPUs, GPUs, FPGAs, and custom accelerators, offer the flexibility to balance power, performance, and cost for various applications. In this section, we explore these essential pillars of edge deployment.


\subsection{Software Tools}\label{sof}
Deploying deep learning models on heterogeneous platforms demands specialized software tools that bridge the gap between cutting-edge artificial intelligence (AI) research and real-world applications. These tools empower developers to optimize, accelerate, and seamlessly integrate AI models across different hardware architectures. \highlight{Table~\ref{softwaretools} compares widely used deployment tools across four functional dimensions: model optimization (MO), model compilation (MC), high-level synthesis (HLS), and inference runtime (IR), with their hardware compatibility spanning FPGA, GPU, and CPU platforms.} 

As FPGAs offer highly parallel and reconfigurable hardware capabilities,  deploying AI models on FPGAs requires specialized software tools for efficient hardware mapping, optimization, and deployment. Both Vivado Design Suite~\footnote{\label{vivado}Vivado design suite. Retrieved January 18, 2025, from \url{https://www.amd.com/en/products/software/adaptive-socs-and-fpgas/vivado.html}} (Xilinx) and the Quartus Prime Design Software~\footnote{\label{quartus}Intel Quartus Prime. Retrieved January 18, 2025, from \url{https://www.intel.com/content/www/us/en/products/details/fpga/development-tools/quartus-prime.html}} (Intel) offer advanced synthesis, converting high-level languages to hardware description language (HDL)  and preoptimized AI accelerators IP cores (such as Xilinx DPU or Intel AI Suite) that help to accelerate the inference task. 

Vitis AI~\footnote{\label{vitis}Vitis AI. Retrieved January 18, 2025, from \url{https://www.xilinx.com/products/design-tools/vitis/vitis-ai.html}} (for Xilinx) and OpenVINO~\footnote{\label{Openvinotoolkit}Openvinotoolkit. Retrieved January 18, 2025, from \url{https://github.com/openvinotoolkit/openvino}}  (for Intel) \highlight{provide support for partial model compression, hardware-optimized operator libraries, and integrated runtime environments. However, both frameworks are tightly coupled to their respective hardware ecosystems, which limits cross-platform portability. Each serves as a comprehensive deployment platform for FPGA-based applications, leveraging their corresponding hardware toolchains for synthesis, optimization, and implementation.} Besides those two software tools, Hls4ml~\footnote{\label{hls4ml}Hls4ml. Retrieved January 18, 2025, from \url{https://fastmachinelearning.org/hls4ml/index.html} } and FINN~\footnote{\label{finn}FINN. Retrieved January 18, 2025, from \url{https://xilinx.github.io/finn/} } designed to explore deep neural network inference on FPGAs efficiently and swiftly. However, both these libraries are still in the experimental phase.

In the GPU domain, TensorRT~\footnote{\label{tensorrt}TensorRT. Retrieved January 18, 2025, from \url{https://developer.nvidia.com/tensorrt}} uses as a popular inference optimizer and accelerating library for NVIDIA GPU-based edge devices. In a broader manner, Triton Inference Server~\footnote{\label{triton}NVIDIA Triton Inference Server. Retrieved January 18, 2025, from \url{https://github.com/triton-inference-server/server}} enables the scalable, production-ready serving of multiple deep learning models on NVIDIA GPUs, x86, and Arm CPUs, integrating seamlessly with frameworks like TensorFlow, Keras, and PyTorch. \highlight{However, neither supports FPGAs.} On the other hand, oneDNN~\footnote{\label{onednn}oneDNN. Retrieved January 18, 2025, from \url{https://github.com/oneapi-src/oneDNN}} is a widely adopted, open-source performance library for deep learning applications, offering cross-platform support. It is highly optimized for Intel CPUs, integrated GPUs, and ARM-based processors, with ongoing experimental support for NVIDIA GPUs, AMD GPUs, and RISC-V architectures. ONNX Runtime~\footnote{\label{onnxrun}ONNX Runtime. Retrieved January 18, 2025, from \url{https://onnxruntime.ai/}} stands out as a versatile, hardware-agnostic inference engine, supporting deployment across CPUs, GPUs, and FPGAs via backend execution providers. It balances broad compatibility with modular optimization and is often preferred in cross-platform production environments.

\highlight{In practice, the choice of tool depends on the target device, optimization needs, and ecosystem. For example, TensorRT and Triton Inference Server are ideal for GPU deployments, Vitis AI excels in customized FPGA acceleration, while ONNX Runtime is well-suited for flexible and portable edge inference across multiple hardware types. In conclusion, unified toolkits for multiple hardware platforms might be an interesting research direction in the future.}
 \renewcommand{\arraystretch}{1.2}
\begin{table}[htb]
\centering
\caption{\highlight{The overview of popular software tools for deploying deep learning models on different hardware architectures.}}
\label{softwaretools}
\resizebox{\columnwidth}{!}{%
\begin{tabular}{c|c|cccc|ccc}
\hline
\multirow{2}{*}{\textbf{Tool}} &
  \multirow{2}{*}{\textbf{Type}} &
  \multicolumn{4}{c|}{\textbf{Key Features}} &
  \multicolumn{3}{c}{\textbf{Supported Edge Devices}} \\ \cline{3-9} 
 &
   &
  \multicolumn{1}{c|}{\textbf{MO}} &
  \multicolumn{1}{c|}{\textbf{MC}} &
  \multicolumn{1}{c|}{\textbf{HLS}} &
  \textbf{IR} &
  \multicolumn{1}{c|}{\textbf{FPGA}} &
  \multicolumn{1}{c|}{\textbf{GPU}} &
  \textbf{CPU} \\ \hline
Xilinx Vivado design suite~\footref{vivado} &
  Toolkit &
  \multicolumn{1}{c|}{\xmark} &
  \multicolumn{1}{c|}{\cmark} &
  \multicolumn{1}{c|}{\cmark} &
  \xmark &
  \multicolumn{1}{c|}{\begin{tabular}[c]{@{}c@{}}Xilinx \\ FPGA\end{tabular}} &
  \multicolumn{1}{c|}{No} &
  No \\ \hline
Intel Quartus Prime~\footref{quartus} &
  Toolkit &
  \multicolumn{1}{c|}{\xmark} &
  \multicolumn{1}{c|}{\cmark} &
  \multicolumn{1}{c|}{\cmark} &
  \xmark &
  \multicolumn{1}{c|}{\begin{tabular}[c]{@{}c@{}}Intel\\ FPGA\end{tabular}} &
  \multicolumn{1}{c|}{No} &
  No \\ \hline
OpenVINO~\footref{Openvinotoolkit} &
  Toolkit &
  \multicolumn{1}{c|}{\cmark} &
  \multicolumn{1}{c|}{\xmark} &
  \multicolumn{1}{c|}{\xmark} &
  \cmark &
  \multicolumn{1}{c|}{\begin{tabular}[c]{@{}c@{}}Intel\\ FPGA\end{tabular}} &
  \multicolumn{1}{c|}{Yes} &
  Yes \\ \hline
Vitis AI~\footref{vitis} &
  Engine &
  \multicolumn{1}{c|}{\cmark} &
  \multicolumn{1}{c|}{\cmark} &
  \multicolumn{1}{c|}{\xmark} &
  \xmark &
  \multicolumn{1}{c|}{\begin{tabular}[c]{@{}c@{}}Xilinx \\ FPGA\end{tabular}} &
  \multicolumn{1}{c|}{No} &
  No \\ \hline
FINN~\footref{finn} &
  Engine &
  \multicolumn{1}{c|}{\cmark} &
  \multicolumn{1}{c|}{\cmark} &
  \multicolumn{1}{c|}{\xmark} &
  \cmark &
  \multicolumn{1}{c|}{\begin{tabular}[c]{@{}c@{}}Xilinx \\ FPGA\end{tabular}} &
  \multicolumn{1}{c|}{No} &
  No \\ \hline
TensorRT~\footref{tensorrt} &
  Engine &
  \multicolumn{1}{c|}{\xmark} &
  \multicolumn{1}{c|}{\xmark} &
  \multicolumn{1}{c|}{\xmark} &
  \cmark &
  \multicolumn{1}{c|}{No} &
  \multicolumn{1}{c|}{\begin{tabular}[c]{@{}c@{}}NVIDIA\\ GPU\end{tabular}} &
  No \\ \hline
ONNX Runtime~\footref{onnxrun} &
  Engine &
  \multicolumn{1}{c|}{\xmark} &
  \multicolumn{1}{c|}{\xmark} &
  \multicolumn{1}{c|}{\xmark} &
  \cmark &
  \multicolumn{1}{c|}{Yes} &
  \multicolumn{1}{c|}{Yes} &
  Yes \\ \hline
Hls4ml~\footref{hls4ml} &
  Library &
  \multicolumn{1}{c|}{\xmark} &
  \multicolumn{1}{c|}{\xmark} &
  \multicolumn{1}{c|}{\cmark} &
  \xmark &
  \multicolumn{1}{c|}{Yes} &
  \multicolumn{1}{c|}{No} &
  No \\ \hline
oneDNN~\footref{onednn} &
  Library &
  \multicolumn{1}{c|}{\cmark} &
  \multicolumn{1}{c|}{\xmark} &
  \multicolumn{1}{c|}{\xmark} &
  \cmark &
  \multicolumn{1}{c|}{No} &
  \multicolumn{1}{c|}{Experimental} &
  Intel CPU, ARM \\ \hline
\begin{tabular}[c]{@{}c@{}}NVIDIA Triton \\ Inference Server~\footref{triton}\end{tabular} &
  Inference Server &
  \multicolumn{1}{c|}{\xmark} &
  \multicolumn{1}{c|}{\xmark} &
  \multicolumn{1}{c|}{\xmark} &
  \cmark &
  \multicolumn{1}{c|}{No} &
  \multicolumn{1}{c|}{\begin{tabular}[c]{@{}c@{}}NVIDIA\\ GPU\end{tabular}} &
  x86, ARM \\ \hline
\end{tabular}%
}
\begin{tablenotes}
\centering
\footnotesize
\item \textit{
\textbf{MO:} Model Optimization;\quad \textbf{MC:} Model Compilation;\\
\textbf{HLS:} High-Level Synthesis;\quad \textbf{IR:} Inference Runtime}
\end{tablenotes}
\end{table}
\subsection{Evaluation Tools}\label{evaluation_m}
Evaluating the performance of ViT acceleration techniques on edge platforms requires specialized tools and metrics to evaluate power consumption, energy efficiency, accuracy, and latency. Fortunately, most hardware vendors offer built-in tools and libraries to facilitate precise measurement of these key performance indicators.
\subsubsection{Latency}\hfill\\ Latency and frame per second (FPS) can be calculated as follows:
\[
\text{FPS} = \frac{1}{Latency}
\]
For GPU-based evaluations, PyTorch provides \texttt{torch.cuda.Event(enable\_timing=True)} for GPU-based evaluations, which accurately measures latency during inference. On NVIDIA EdgeGPU platforms, the \textbf{TensorRT Profiler} offers a detailed latency breakdown for Jetson boards. For AMD FPGAs, the \textbf{Vitis AI Profiler~\footnote{\label{vitisAI}Vitis AI Profiler. Retrieved February 10, 2025, from \url{https://github.com/Xilinx/Vitis-AI/tree/master/examples/vai_profiler}}} enables profiling during deployment, ensuring optimized execution. Additionally, \textbf{Intel's OpenVINO benchmark tool~\footref{Openvinotoolkit}} supports latency and throughput measurements across Intel CPUs and FPGAs, providing a standardized evaluation framework.

\subsubsection{Power}\hfill\\ Measuring power consumption is critical yet challenging in evaluating ViT acceleration techniques. Standard tools for general-purpose platforms (GPPs) like CPUs and GPUs include \textbf{Intel Power Gadget} for Intel CPUs and \textbf{NVIDIA-SMI} for NVIDIA GPUs. Power can be measured on edge GPU platforms, such as NVIDIA Jetson boards, using \textbf{tegraStats}, which provides real-time power monitoring, GPU utilization, and temperature. For FPGAs and ACAPs, AMD Xilinx offers \textbf{Xilinx Power Estimator (XPE)~\footnote{\label{xpe}Xilinx Power Estimator. Retrieved January 18, 2025, from \url{https://www.amd.com/en/products/adaptive-socs-and-fpgas/technologies/power-efficiency/power-estimator.html}}} for power estimation based on hardware configurations, while \textbf{Vaitrace} enables runtime power profiling for FPGA and adaptive compute acceleration platforms (ACAP). These tools provide essential insights into power efficiency, thermal behavior, and overall performance trade-offs across edge hardware platforms.
\subsubsection{Energy}\hfill\\ Energy consumption in ViT acceleration can be estimated through throughput per joule (GOP/J) and FPS per watt (FPS/W). However, accurately measuring energy on general-purpose platforms (GPPs) is challenging due to background processes affecting power readings. In contrast, edge devices provide a more controlled environment where only one primary task is executed simultaneously, making energy estimation more reliable. Several tools facilitate energy measurement: \textbf{Xilinx Vivado Power Analyzer~\footref{vivado}} estimates energy efficiency for FPGAs by profiling dynamic power, while \textbf{RAPL} tracks CPU-level energy consumption on x86 architectures. Additionally, energy efficiency can be derived using power measurements combined with latency, enabling a deeper evaluation of acceleration techniques.

\subsubsection{Resosurce Utlization}\hfill\\ Resource utilization is primarily analyzed in FPGA-based acceleration techniques to optimize hardware efficiency and minimize resource usage. Currently, Intel and AMD are two FPGA vendors. AMD offers \textbf{Vivado Design suite~\footref{vivado}} for synthesis evaluation of the FPGA before deployment. Similarly, Intel provides the \textbf{Quartus Prime~\footref{quartus}} software, which facilitates FPGA synthesis, resource utilization monitoring, and performance evaluation. Both vendors offer additional AI optimization frameworks—AMD’s \textbf{Vitis AI~\footref{vitis}} and Intel’s \textbf{FPGA SDK for OpenCL (AOCL)}—to enhance the efficiency of ViT acceleration on FPGA platforms.
\subsection{Common Optimization Techniques}
\textbf{Memory Optimization} Techniques like Huffman coding can be used to compress the weights. On-chip memory utilization efficiently uses the FPGA's Block RAMs (BRAMs) to store weights and intermediate feature maps, reducing the need for off-chip memory accesses, which can be slow and power-hungry.

\noindent \textbf{Pipeline Parallelism} The technique splits the model into stages and simultaneously processes different inputs at each stage, which helps in maximizing the throughput.

\noindent \textbf{Loop Unrolling} This FPGA-specific optimization involves unrolling loops in the FPGA design to speed up the processing. For instance, when performing matrix multiplications in the transformer layers.

\noindent \textbf{Layer Fusion} Layer fusion combines multiple layers into a single computational unit, reducing memory access between layers and improving the overall throughput.

\noindent \textbf{Hardware-friendly Activation Functions} Replace complex activation functions with simpler, hardware-friendly alternatives. For example, using piecewise linear approximations for non-linearities.
 
\noindent \textbf{Optimized Matrix Operations} ViT involves many matrix multiplications (in the attention mechanisms). Optimizing these matrix operations for FPGA leads to significant speed-ups. Techniques like systolic arrays or optimized linear algebra cores are employed.

\noindent \textbf{Dynamic Precision} In recent studies, some work uses mixed precision computations where certain parts of the model use lower precision (e.g., 8-bit). In comparison, other parts use higher precision (e.g., 16-bit or 32-bit). There are some works in which the authors introduced fixed point and PoT precision and optimally balanced accuracy and performance.

\highlight{Although software frameworks enable optimized ViT execution with hardware-aware features, achieving real-time performance on edge devices often requires low-level, hardware-specific acceleration techniques. Section~\ref{acce_tech} explores dedicated acceleration methods for deploying ViTs on edge platforms.}
\section{Accelerating Strategies For ViT on Edge}\label{acce_tech}
This section explores acceleration strategies for non-linear operations, discusses SOTA ViT acceleration techniques, and provides a comprehensive performance analysis in terms of both hardware efficiency and accuracy.
\subsection{Accelerating Non-linear operations} ViT models in CV can mainly be split into two types of operations: linear and non-linear. Optimizing non-linear operations in quantized ViT is as crucial as optimizing linear operations. While low-bit computing units significantly reduce computational complexity and memory footprint, \highlight{they} are primarily designed for linear operations such as matrix multiplications and convolutions. However, non-linear functions, including softmax, GELU, and LayerNorm, are the essential components of ViT architectures yet \highlight{largely unexplored} in these low-bit computing environments. The lack of support for non-linear operations on hardware creates computational bottlenecks during the edge deployment, as non-linear functions often require FP32 operations. This results in latency and increases power and energy consumption, ultimately lessening the full benefits \highlight{of quantization} on edge deployment. Moreover, during inference on quantized ViT models on edge devices, frequent quantization and dequantization operations surrounding non-linear layers add further inefficiencies, slowing inference and reducing throughput~\cite{stevens2021softermax}. Researchers developed integer-based approximations for non-linear operations to solve these issues, eliminating the need for frequent FP32 computations in those layers.

As illustrated in Table~\ref{nonlinearoperation}, we introduce integer-only approximations for different hardware platforms to enhance ViT inference efficiency. As we discussed details in section~\ref{quan}, FQ-ViT~\cite{lin2021fq} utilized LIS for integer-only variants of the softmax that approximate the exponential component using \highlight{a} second-order polynomial coupled with $\log 2$ quantization. For LayerNorm, the authors applied PTF to shift the quantized activations and later computed mean and variance using integer arithmetic. However, this approach seems hardware efficient; their methods of practical deployment on edge platforms are unclear. Additionally, I-ViT~\cite{li2022vit} calculates the square root in LayerNorm using an integer-based iterative method. The authors then introduced ShiftGELU, which used sigmoid-based approximations for GELU approximations. I-ViT utilized NVIDIA RTX 2080 Ti GPU as a hardware platform to evaluate their method. 

PackQViT~\cite{dong2024packqvit} is an extended version of the FQ-ViT concept that also leverages second-order polynomial approximations more straightforwardly, replacing the usual constant $e$ with 2 in the softmax. Although PackQViT requires training, the simplification ensures no accuracy loss. EdgeKernel~\cite{zhang2023practical} addressed precision challenges in softmax computations by optimizing the selection of the bit shift parameter on Apple A13 and M1 chips, ensuring high accuracy while minimizing significant bit truncations. Additionally, it employs asymmetric quantization for LayerNorm inputs, converting them to a uint16 format to enhance computational efficiency while maintaining data integrity.

SOLE~\cite{wang2023sole} and SwiftTron~\cite{marchisio2023swifttron} both are utilized application-specific integrated circuit (ASIC) platforms. SOLE optimizes the software perspectives, introducing E2Softmax with \highlight{$\log_{2}$} quantization to avoid traditional FP32 precision in softmax layers. Additionally, SOLE designed a two-stage LayerNorm unit using PTF factors. However, SwiftTron focused on \highlight{developing} customized hardware for ASIC to efficiently use non-linear operations, even in FP32, which accounts for diverse scaling factors in performing correct computations.
\renewcommand{\arraystretch}{1.2}
\begin{table}[]
\centering
\caption{Overview of ViT models utilizing integer approximations for non-linear operations to enhance inference efficiency and avoid dequantization.}
\label{nonlinearoperation}
\resizebox{0.8\columnwidth}{!}{%
\begin{tabular}{c|c|ccc|c}
\hline
\multirow{2}{*}{\textbf{Model}} &
  \multirow{2}{*}{\textbf{\begin{tabular}[c]{@{}c@{}}Experiment\\ Hardware\end{tabular}}} &
  \multicolumn{3}{c|}{\textbf{Non-linear Operations}} &
  \multirow{2}{*}{\textbf{Retrain}} \\ \cline{3-5}
 &
   &
  \multicolumn{1}{c|}{\textbf{Softmax}} &
  \multicolumn{1}{c|}{\textbf{GELU}} &
  \textbf{LayerNorm} &
   \\ \hline
FQ-ViT~\cite{lin2021fq} &
  - &
  \multicolumn{1}{c|}{\cmark} &
  \multicolumn{1}{c|}{\xmark} &
  \cmark &
  \xmark \\ \hline
I-ViT~\cite{li2022vit} &
  RTX 2080 Ti GPU &
  \multicolumn{1}{c|}{\cmark} &
  \multicolumn{1}{c|}{\cmark} &
  \cmark &
  \cmark \\ \hline
EdgeKernel~\cite{zhang2023practical} &
  Apple A13 and M1 chips &
  \multicolumn{1}{c|}{\cmark} &
  \multicolumn{1}{c|}{\cmark} &
  \cmark &
  \xmark \\ \hline
PackQViT~\cite{dong2024packqvit} &
  Snapdragon 870 SoC (Mobile Phone) &
  \multicolumn{1}{c|}{\cmark} &
  \multicolumn{1}{c|}{\cmark} &
  \cmark &
  \cmark \\ \hline
SOLE~\cite{wang2023sole} &
  ASIC 28nm &
  \multicolumn{1}{c|}{\cmark} &
  \multicolumn{1}{c|}{\cmark} &
  \cmark &
  \xmark \\ \hline
SwiftTron~\cite{marchisio2023swifttron} &
  ASIC 65 nm &
  \multicolumn{1}{c|}{\cmark} &
  \multicolumn{1}{c|}{\cmark} &
  \cmark &
  \xmark \\ \hline
\end{tabular}%
}
\vspace{-3mm}
\end{table}
\subsection{Current Accelerating Techniques on ViT}
Efficient acceleration techniques for seamless deployment on edge devices are essential where computational and energy constraints limit performance. Various techniques have been developed to optimize ViT execution, balancing throughput, latency, and energy efficiency while ensuring minimal loss in accuracy. These techniques can be broadly classified into SW-HW co-design and hardware-only acceleration. SW-HW co-design integrates algorithmic optimizations with hardware-aware modifications to ensure efficient deployment of ViTs on edge devices, enhancing throughput and energy efficiency. On the other hand, hardware-only approaches push efficiency even further by designing architectures specifically optimized for transformer workloads, utilizing systolic arrays, spatial computing, and near-memory processing to eliminate bottlenecks caused by data movement and external memory access. Additionally, alternative approaches, such as distributing multiple tiny edge devices by partitioning the model into submodels, offer promising directions to rethink ViT acceleration from a fundamentally different perspective. 
\renewcommand{\arraystretch}{1.2}
\begin{table}[!htb]
\centering
\caption{The overview of current accelerating techniques for ViT on edge devices. 
In the baseline models, \textbf{B} denotes Base; \textbf{S} denotes Small, and \textbf{T} denotes Tiny versions. 
Five types of hardware devices are used in existing SW-HW co-design frameworks: 
GPU ($\spadesuit$), EdgeGPU ($\blacktriangle$), CPU ($\bigstar$), FPGA ($\clubsuit$), and AMD Versal Adaptive Compute Acceleration (ACAP) ($\blacklozenge$).}
\label{hwswco}
\resizebox{1\columnwidth}{!}{%
\begin{tabular}{c|c|c|c|cc|cc}
\hline
\multirow{2}{*}{\textbf{Approaches}} &
  \multirow{2}{*}{\textbf{Framework}} &
  \multirow{2}{*}{\textbf{Retrain}} &
  \multirow{2}{*}{\textbf{\begin{tabular}[c]{@{}c@{}}Baseline\\ Models\end{tabular}}} &
  \multicolumn{2}{c|}{\textbf{Hardware device}} &
  \multicolumn{2}{c}{\textbf{Key optimization}} \\ \cline{5-8} 
 &
   &
   &
   &
  \multicolumn{1}{c|}{\textbf{Baseline}} &
  \textbf{Experiment} &
  \multicolumn{1}{c|}{\textbf{Software}} &
  \textbf{Hardware} \\ \hline
\multirow{13}{*}{SW-HW co-design} &
  \multirow{2}{*}{VAQF~\cite{sun2022vaqf}} &
  \multirow{2}{*}{\xmark} &
  \multirow{2}{*}{DeiT-B/S/T} &
  \multicolumn{1}{c|}{Intel i7-9800X\textsuperscript{$\bigstar$}} &
  \multirow{2}{*}{ZCU102\textsuperscript{$\clubsuit$}} &
  \multicolumn{1}{c|}{\multirow{2}{*}{Quantization}} &
  \multirow{2}{*}{\begin{tabular}[c]{@{}c@{}}Traditional\\ resources\end{tabular}} \\
 &
   &
   &
   &
  \multicolumn{1}{c|}{TITAN RTX\textsuperscript{$\spadesuit$}} &
   &
  \multicolumn{1}{c|}{} &
   \\ \cline{2-8} 
 &
  Auto-ViT-Acc ~\cite{lit2022auto} &
  \xmark &
  DeiT-B/S/T &
  \multicolumn{1}{c|}{A100\textsuperscript{$\spadesuit$}} &
  ZCU102 \textsuperscript{$\clubsuit$} &
  \multicolumn{1}{c|}{\begin{tabular}[c]{@{}c@{}}Mixed \\ quantization\end{tabular}} &
  \begin{tabular}[c]{@{}c@{}}Traditional\\ resources\end{tabular} \\ \cline{2-8} 
&
  HeatViT~\cite{dong2023heatvit} &
  \cmark &
  DeiT-B/S/T &
  \multicolumn{1}{c|}{A100\textsuperscript{$\spadesuit$}} &
  ZCU102 \textsuperscript{$\clubsuit$} &
  \multicolumn{1}{c|}{\begin{tabular}[c]{@{}c@{}}Adaptive \\ token pruning\end{tabular}} &
  \begin{tabular}[c]{@{}c@{}}Motivated from\\ ~\cite{lit2022auto}\end{tabular} \\ \cline{2-8} 
 &
  \multirow{4}{*}{EQ-ViT~\cite{10745859}} &
  \multirow{4}{*}{\cmark} &
  \multirow{4}{*}{DeiT-T} &
  \multicolumn{1}{c|}{ZCU102\textsuperscript{$\clubsuit$}} &
  \multirow{2}{*}{VCK190\textsuperscript{$\blacklozenge$}} &
  \multicolumn{1}{c|}{\multirow{4}{*}{\begin{tabular}[c]{@{}c@{}}Kernel-level\\ profiling\end{tabular}}} &
  \multirow{4}{*}{\begin{tabular}[c]{@{}c@{}}Spatial \& heterogeneous \\ accelerators\end{tabular}} \\
 &
   &
   &
   &
  \multicolumn{1}{c|}{U250\textsuperscript{$\clubsuit$}} &
   &
  \multicolumn{1}{c|}{} &
   \\
   &
   &
   &
   &
  \multicolumn{1}{c|}{AGX Orin\textsuperscript{$\blacktriangle$}} &
  \multirow{2}{*}{VEK280\textsuperscript{$\blacklozenge$}} &
  \multicolumn{1}{c|}{} &
   \\
 &
   &
   &
   &
  \multicolumn{1}{c|}{A100\textsuperscript{$\spadesuit$}} &
   &
  \multicolumn{1}{c|}{} &
   \\ \cline{2-8} 
 &
  \multirow{2}{*}{M\textsuperscript{3}ViT~\cite{fan2022m3vit}} &
  \multirow{2}{*}{\cmark} &
  \multirow{2}{*}{ViT-S/T} &
  \multicolumn{1}{c|}{\multirow{2}{*}{RTX 8000\textsuperscript{$\spadesuit$}}} &
  \multirow{2}{*}{ZCU102\textsuperscript{$\clubsuit$}} &
  \multicolumn{1}{c|}{\multirow{2}{*}{\begin{tabular}[c]{@{}c@{}}Mixture of expert\\ (MoE)\end{tabular}}} &
  \multirow{2}{*}{\begin{tabular}[c]{@{}c@{}}Computing MoE\\ expert-by-expert\end{tabular}} \\
 &
   &
   &
   &
  \multicolumn{1}{c|}{} &
   &
  \multicolumn{1}{c|}{} &
   \\ \cline{2-8} 
 &
  \multirow{2}{*}{ViTCoD~\cite{you2023vitcod}} &
  \multirow{2}{*}{\cmark} &
  \multirow{2}{*}{DeiT-B/S/T} &
  \multicolumn{1}{c|}{Jetson Xavier\textsuperscript{$\blacktriangle$}} &
  \multirow{2}{*}{ASIC (28nm)} &
  \multicolumn{1}{c|}{\multirow{2}{*}{\begin{tabular}[c]{@{}c@{}}Prunes \& polarizes \\ the attention maps\end{tabular}}} &
  \multirow{2}{*}{\begin{tabular}[c]{@{}c@{}}On-chip encoder \\ \& decoder engines\end{tabular}} \\
 &
   &
   &
   &
  \multicolumn{1}{c|}{CPU\textsuperscript{$\bigstar$}} &
   &
  \multicolumn{1}{c|}{} &
   \\ \cline{2-8} 
 &
  SOLE~\cite{wang2023sole} &
  \xmark &
  DeiT-T &
  \multicolumn{1}{c|}{2080Ti\textsuperscript{$\spadesuit$}} &
  ASIC (28nm) &
  \multicolumn{1}{c|}{\begin{tabular}[c]{@{}c@{}}E2Softmax \& \\ AILayerNorm\end{tabular}} &
  \begin{tabular}[c]{@{}c@{}}Custom hardware\\ unit\end{tabular} \\ \hline
\multirow{3}{*}{Pure HW} &
  \multirow{2}{*}{ViA~\cite{wang2022via}} &
  \multirow{2}{*}{\xmark} &
  \multirow{2}{*}{Swin-T} &
  \multicolumn{1}{c|}{Intel i7-5930X\textsuperscript{$\bigstar$}} &
  \multirow{2}{*}{Alveo U50\textsuperscript{$\clubsuit$}} &
  \multicolumn{1}{c|}{\multirow{2}{*}{-}} &
  \multirow{2}{*}{\begin{tabular}[c]{@{}c@{}}Multi kernel parallelism\\ with half mapping method\end{tabular}} \\
 &
   &
   &
   &
  \multicolumn{1}{c|}{V100\textsuperscript{$\spadesuit$}} &
   &
  \multicolumn{1}{c|}{} &
   \\ \cline{2-8} 
 &
  ViTA ~\cite{nag2023vita} &
  \xmark &
  DeiT-B/S/T &
  \multicolumn{1}{c|}{ASIC (40nm)} &
  Zynq ZC7020\textsuperscript{$\clubsuit$} &
  \multicolumn{1}{c|}{-} &
  \begin{tabular}[c]{@{}c@{}}Head level pipeline\\ \& Inter-layer MLP\end{tabular} \\ \hline
\multirow{3}{*}{Other Techniques} &
  ED-ViT~\cite{liu2024ed} &
  \cmark &
  ViT-B/S/T &
  \multicolumn{1}{c|}{-} &
  \begin{tabular}[c]{@{}c@{}}Raspberry Pi-4B\\ RTX 4090\textsuperscript{$\spadesuit$}\end{tabular} &
  \multicolumn{1}{c|}{-} &
  \begin{tabular}[c]{@{}c@{}}Distributed edge devices\\ for deploying submodels\end{tabular} \\ \cline{2-8} 
 &
  \multirow{2}{*}{COSA Plus~\cite{10612833}} &
  \multirow{2}{*}{\xmark} &
  \multirow{2}{*}{ViT-B} &
  \multicolumn{1}{c|}{RTX 3090\textsuperscript{$\spadesuit$}} &
  \multirow{2}{*}{XCVU13P \textsuperscript{$\clubsuit$}} &
  \multicolumn{1}{c|}{\multirow{2}{*}{-}} &
  \multirow{2}{*}{\begin{tabular}[c]{@{}c@{}}Systolic array with\\ optimized dataflow\end{tabular}} \\ \cline{5-5}
   &
   &
   &
   &
  \multicolumn{1}{c|}{6226R server CPU\textsuperscript{$\bigstar$}} &
   &
  \multicolumn{1}{c|}{} &
   \\ \hline
\end{tabular}
}
\vspace{-3mm}
\end{table}
\subsubsection{Software Optimization in \highlight{SW-HW} Co-design} \label{sw_hw_s}\hfill \\
Recent advancements in software part in SW-HW co-design for ViT acceleration encompass a range of optimization techniques, including quantization-based acceleration~\cite{sun2022vaqf,lit2022auto,10745859,dong2023heatvit}, sparse and adaptive attention mechanisms~\cite{you2023vitcod,dong2023heatvit}, adding mixture of experts (MoE) layers~\cite{fan2022m3vit}, analyzing kernel profiling and execution scheduling~\cite{10745859}, custom hardware softmax and LayerNorm to replace traditional softmax and LayerNorm~\cite{wang2023sole,stevens2021softermax}.\\ 

\noindent \textbf{Quantization-Based Software Acceleration} VAQF~\cite{sun2022vaqf} utilized binary quantization for weights and low-precision for the activations in the software part. VAQF automatically outputs the efficient quantization parameters based on the model structure and the expected frame per second (FPS) to meet the hardware specifications. The primary purpose of this work was to achieve high throughput on hardware while maintaining model accuracy. Moreover, Li et al.~\cite{lit2022auto} used a mixed-scheme (fixed+PoT) ViT quantization algorithm that can fully leverage heterogeneous FPGA resources for a target FPS. Both frameworks utilized targeted FPS as their input to achieve maximum hardware efficiency during inference. Additionally, both VAQF and AutoViT-Acc~\cite{lit2022auto} utilized PTQ methods as quantization. EQ-ViT~\cite{10745859} combined latency and accuracy requirements to decide the final quantization strategy \highlight{,} leveraging the QAT approach. Additionally, EQ-ViT implemented a profiling-based execution scheduler that dynamically allocates workloads across hardware accelerators.\\

\noindent \textbf{Pruning-Based Software Acceleration} ViT's self-attention has quadratic complexity concerning input sequence length, leading to high memory bandwidth consumption. Sparse attention mechanisms aim to reduce redundant computation by focusing on pruning. For instance, ViTCoD~\cite{you2023vitcod} efficiently employed structured \highlight{pruning} and polarized the attention maps to remove redundant attention scores, creating a more memory-efficient execution. Moreover,  HeatViT~\cite{dong2023heatvit} utilized image-adaptive token pruning, identifying and removing unimportant tokens before transformer blocks using a multi-token selector, dynamically reducing computational complexity. The proposed method is highly inspired by SP-ViT ~\cite{kong2022spvit} and uses a similar token selector like SP-ViT. \\

\noindent \textbf{Other Approaches} Additionally, MoE improves model efficiency by activating only the most relevant expert networks per input instance, reducing unnecessary computations. One of the first studies named M\textsuperscript{3}ViT~\cite{fan2022m3vit} implemented MoE layers where a router dynamically selects the appropriate experts for processing. In this work, the authors conveyed training dynamics to balance large capacity and efficiency by selecting only a subset of experts using the MoE router. Beyond traditional quantization and attention optimizations, some co-design approaches target the computational bottlenecks of softmax and LayerNorm. SOLE~\cite{wang2023sole} proposed E2Softmax and AILayerNorm, hardware-aware modifications for non-linear operations that replace FP32 with integer-only approximation. This integer-only computation improved the latency significantly during the inference.
\subsubsection{Hardware Optimization in \highlight{SW-HW} Co-design}\label{s_hw_sw}\hfill \\
Hardware optimization plays a crucial role in the \highlight{SW-HW} co-design of ViTs, ensuring efficient execution across different accelerators. Key techniques include C++ based hardware descriptions, high-level synthesis (HLS), and accelerator bitstream generation~\cite{sun2022vaqf,lit2022auto,dong2023heatvit}. Additionally, frameworks leverage AI engine (AIE) kernels~\cite{10745859} and custom hardware units (e.g., SOLE) to optimize execution for edge deployment.\\

\noindent \textbf{\highlight{ASIC-Based Hardware Acceleration}} \highlight{ASIC-based accelerators have become a cornerstone in deploying deep learning models on edge devices, with the majority of advancements driven by leading technology companies. Notable examples include Google’s TPU~\cite{akopyan2015truenorth}, IBM’s TrueNorth~\cite{akopyan2015truenorth}, and NVIDIA’s open-source NVDLA~\cite{NVDLA2021}. Academic research has also contributed with efficient designs such as Origami~\cite{cavigelli2016origami} and ZASCA~\cite{ardakani2019fast}, which are tailored for low-power convolutional inference.}

\highlight{Google’s TPUs have released several versions~\cite{jouppi2018domain}. TPUv1, released in 2016, featured a 256$\times$256 systolic array of 8-bit fixed-point multipliers, delivering up to 92 TOPS at 75 W—well-suited for accelerating matrix-heavy operations like those found in ViT inference. TPUv2 and TPUv3 introduced support for FP operations and higher throughput. For resource-constrained environments, Google developed the EdgeTPU, a lightweight variant tailored for edge AI workloads.}

\highlight{As ViTs grow in popularity, their unique architectural demands have motivated the development of ViT-specific ASIC accelerators. Unlike CNN-based accelerators, ViT inference requires efficient handling of attention mechanisms and normalization layers. Recent efforts such as ViTCoD~\cite{you2023vitcod} and SOLE\cite{wang2023sole} addressed these challenges through SW-HW co-design strategies and have demonstrated promising ASIC implementations for ViT acceleration at the edge.}\\

\begin{figure}[]
  \centering
  \begin{minipage}[b]{0.47\linewidth}
    \includegraphics[width=\linewidth]{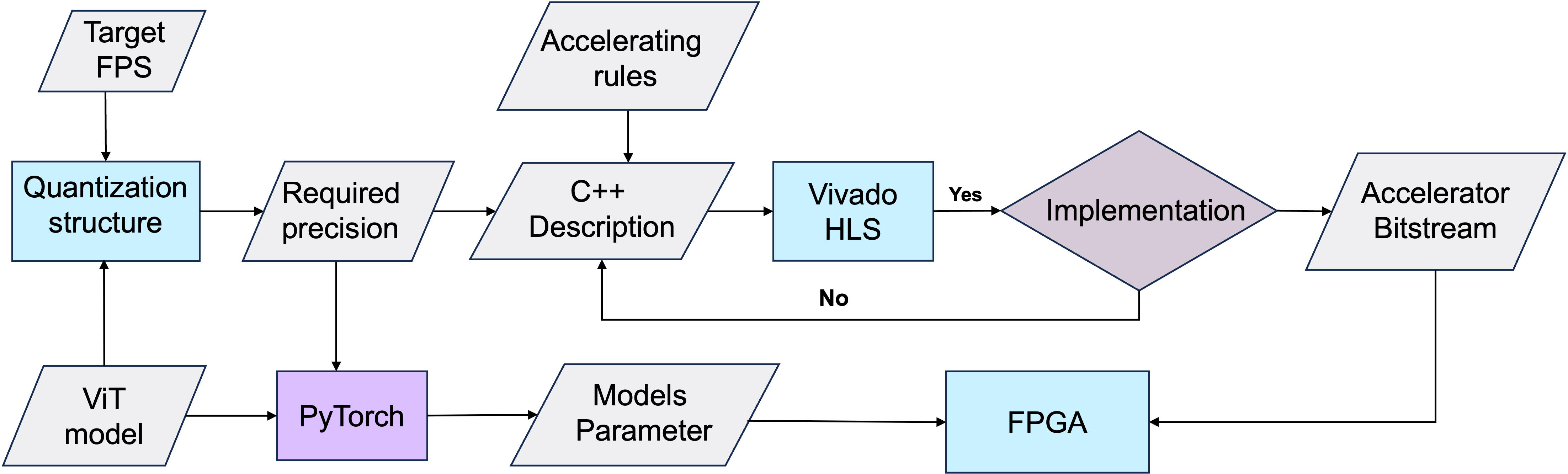}
   \caption{Overflow of VAQF acclerator~\cite{sun2022vaqf}. Using different colors in the architecture differentiates between types of processes within the overall workflow. The light gray boxes represent settings that are input/output to the process. The light blue boxes denote active processing steps or software tools within the workflow, like Vivado HLS. The lavender box signifies a platform/library used in the process, like "PyTorch". The light purple box indicates decision-making points or critical stages in the architecture.}
  \label{fig:vaqf}
  \end{minipage}
  \hspace{0.5cm} 
  \begin{minipage}[b]{0.47\linewidth}
    \includegraphics[width=\linewidth]{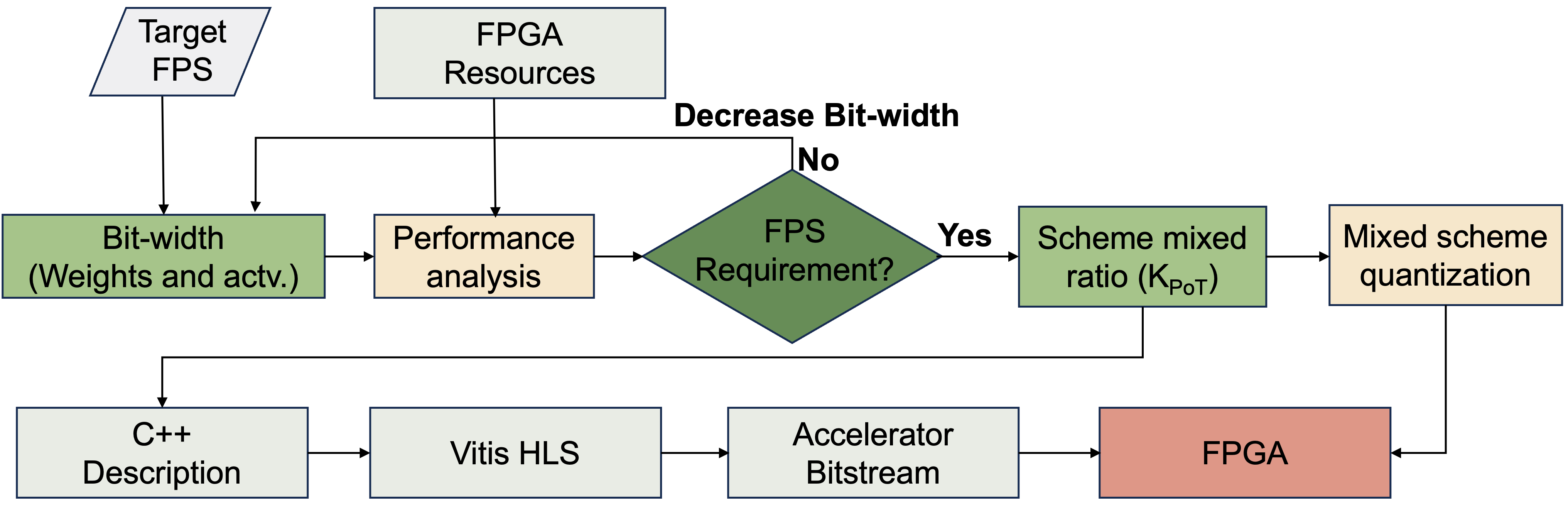}
  \caption{The overview of Auto-ViT-Acc framework ~\cite{lit2022auto}. The "FPGA resource utilization modeling" was utilized for performance analysis and the estimated FPS rate for ViT accelerator with assigned bit-width for mixed schemes and lessening the bit-width until achieving the target FPS. The proposed mixed-scheme quantization then utilized mixed ratio (k\textsubscript{pot}) results to implement on FPGA through "C++ Description for accelerator," "Vitis HLS" and "Accelerator bitstream"}
  \label{fig:aut-vcc}
  \end{minipage}
\end{figure}

\noindent \textbf{FPGA-Based Hardware Acceleration} FPGA is the pioneer for hardware accelerating strategies because of its reconfigurable characteristics. VAQF~\cite{sun2022vaqf} adopted the quantization schemes from the software part and applied them to the accelerator on the hardware side. Figure~\ref{fig:vaqf} illustrates the overview of the VAQF framework. The accelerator's C++ description was synthesized using the Vivado HLS tool. Initial accelerator parameters focused on maximizing parallelism, but there were adjustments due to Vivado's placement or routing problems. Successful implementations produced a bitstream file for FPGA deployment. Moreover, as illustrated in Figure~\ref{fig:aut-vcc}, Auto-ViT Acc~\cite{lit2022auto} first used the "FPGA Resource Utilization Modeling" module to give performance analysis and calculate the FPS of the FPGA ViT accelerator in \highlight{the} software part. Inspired by VAQF, the authors implemented the FPGA accelerator using a C++ hardware description, synthesized through Vitis HLS to generate the final accelerator bitstream. 

In ViTs, the computational bottleneck often arises from General Matrix Multiply (GEMM) operations, which form the core of self-attention and feed-forward layers. HeatViT~\cite{dong2023heatvit} optimizes ViT execution by dynamically selecting tokens and loading each layer from off-chip DDR memory to on-chip buffers before processing via the GEMM engine. This approach, inspired by \cite{lit2022auto}, minimizes redundant computations and improves memory efficiency. The proposed HeatViT addressed two main challenges of hardware implementation in their proposed architecture, as follows.
\begin{enumerate}
\item The GEMM loop tiling must be adjusted to factor in an extra dimension from multi-head parallelism.
\item ViTs have more non-linear operations than CNNs; these must be optimized for better quantization and efficient hardware execution while maintaining accuracy.
\end{enumerate}

\noindent \textbf{MoE Execution for Hardware Efficiency} While MoE optimization techniques dynamically select experts, leading to an unpredictable computing pattern that makes hardware execution difficult. M\textsuperscript{3}ViT~\cite{fan2022m3vit} reordered computations to process tokens expert-by-expert rather than token-by-token, reducing irregular memory access and improving hardware parallelism. However, frequent off-chip memory accesses in MoE layers introduce a latency bottleneck. The authors utilized a ping-pong buffering technique for continuous processing without memory stalls, where one buffer fetches expert weights while another buffer performs computations. Additionally, the per-expert token queueing system groups tokens per expert, limiting the underutilization of compute units due to various expert demands.\\

\noindent \textbf{Optimizing Non-Linear Operations with Custom Hardware Units} Recent studies, such as EQ-ViT, separated matrix multiply (MM) and non-MM by efficiently mapping batch MM (BMM) and convolutions to AIE vector cores. Memory-bounded and non-linear operations are executed within the FPGA's programmable element. Additionally, EQ-ViT leveraged fine-grained pipeline execution to overlap computation with memory transfers,  maximizing resource utilization. A key strength of this framework lies in its hardware mapping methodology, where execution is formulated as a mixed-integer programming (MIP) optimization problem, ensuring that latency and resource constraints are satisfied while maximizing throughput. Likewise, SOLE~\cite{10745859} designed separate units for the proposed E2Softmax  and AILayerNorm to perform non-linear operations efficiently. The E2Softmax unit included Log2Exp and an approximate Log-based divider that is implemented in a LUT-free and multiplication-free manner. Additionally, the AILayerNorm unit operates in two stages: the first stage performs statistical calculations, while the second stage applies the affine transformation. Similar to M\textsuperscript{3}ViT~\cite{fan2022m3vit}\highlight{, SOLE} also utilized ping pong buffer to pipeline the AILayerNorm unit. \\

\noindent \textbf{Pure Hardware Accelerators} Although most ViT acceleration techniques rely on SW-HW co-design—where software optimization plays a crucial role in achieving high efficiency and accuracy—the potential of pure hardware optimization has gained attention in recent studies. A recent study named ViA~\cite{wang2022via} addressed issues during data flow through the layers in ViT. The authors utilized a partitioning strategy to reduce the impact of data locality in the image and enhance the efficiency of computation and memory access. Additionally, by examining the computing flow of the ViT, the authors also utilized the half-layer mapping and throughput analysis to lessen the effects of path dependency due to the shortcut mechanism and to maximize the use of hardware resources for efficient transformer execution. The study developed two reuse processing engines with an internal stream, distinguishing them from previous overlaps or stream design patterns from the optimization strategies.

Moreover, ViTA~\cite{nag2023vita} used two sets of MAC units to minimize the off-chip memory accesses. The first set of MAC units representing the hidden layers was broadcast to the second set of MAC units through a non-linear activation function. That broadcasting approach helped to compute the partial products corresponding to the output layer. The authors allocated these resources to maintain the pipeline technique as if the hidden layer value computations and the output layer partial product computation took equal time. This approach enabled the integration of several mainstream ViT models by only adjusting the configuration.\\

\noindent \textbf{Sparse Attention Optimization for ViTs}  Recent studies such as ViTCoD~\cite{you2023vitcod} introduced a sparser engine to process the sparse attention metrics. The authors structured the multiply-accumulate (MAC) into the encoder and decoder MAC lines to optimize the matrix multiplications.

\begin{table}[]
\centering
\caption{Compatibilities comparison of SOTA SW-HW co-design accelerating techniques. Here, GPP denotes general-purpose platforms such as CPU and GPU.}
\label{performance_analysis}
\renewcommand{\arraystretch}{1.2}
\resizebox{0.8\columnwidth}{!}{%
\begin{tabular}{c|c|c|c|c|c}
\hline
\multirow{2}{*}{\textbf{Framework}} &
  \multirow{2}{*}{\textbf{Baseline}} &
  \multirow{2}{*}{\textbf{\begin{tabular}[c]{@{}c@{}}Effort\\ Modulation\end{tabular}}} &
  \multirow{2}{*}{\textbf{\begin{tabular}[c]{@{}c@{}}Prediction\\ Mechanism\end{tabular}}} &
  \multirow{2}{*}{\textbf{\begin{tabular}[c]{@{}c@{}}Accuracy\\ Top1(\%)\end{tabular}}} &
  \multirow{2}{*}{\textbf{\begin{tabular}[c]{@{}c@{}}GPP\\ Compatible\end{tabular}}} \\
                               &        &                 &                &      &        \\ \hline
ViTCoD~\cite{you2023vitcod}    & DeiT-S & Constant        & Norm Score     & 78.1 & \xmark \\ \hline
HeatViT~\cite{dong2023heatvit} & DeiT-S & Constant        & Head level     & 79.1 & \xmark \\ \hline
PIVOT ~\cite{moitra2024pivot}  & DeiT-S & Input-aware     & Entropy Metric & 79.4 & \cmark \\ \hline
VAQF~\cite{sun2022vaqf}        & DeiT-S & Hardware-driven & Hardware-aware & 79.5 & \xmark \\ \hline
\end{tabular}%
}
\vspace{-3mm}
\end{table}
\subsubsection{Other Techniques}\hfill \\
 ED-ViT~\cite{liu2024ed} utilized distributed workloads approaches to deploy the ViT \highlight{model} utilizing multiple tiny edge devices such as Raspberry Pi-4B. The authors partitioned the model into multiple submodels, mapping each submodel to a separate edge device. This distributed execution strategy enables ViTs to achieve efficiency comparable to single powerful edge accelerators like EdgeGPUs or FPGAs while leveraging cost-effective and scalable edge computing resources. Similarly, COSA Plus, proposed by Wang et al.~\cite{10612833}, capitalized on high inherent parallelism within ViT models by implementing a runtime-configurable hybrid dataflow strategy. This method dynamically switches between weight-stationary and output-stationary dataflows in a systolic array, optimizing the computational efficiency for matrix multiplications within the attention mechanism. COSA Plus enhances processing element (PE) utilization by adapting data movement patterns to workload variations.
\subsubsection{Discussion}\hfill \\
From our observations, most of the SW-HW co-design proposed for FPGA or ACAP design due to their reconfigurable nature where the authors baselined GPU, CPU, or other FPGA platforms. Table~\ref{hwswco} provides a comprehensive overview of the SOTA SW-HW co-design strategies for ViTs on various edge platforms. The analysis focuses on the hardware devices and optimization techniques. A key observation from this table is that different hardware platforms support distinct optimization levels. While GPUs and EdgeGPUs are widely used due to their parallel processing capabilities, FPGAs and ACAP platforms provide customized acceleration, often resulting in lower latency and energy-efficient execution. 

VAQF and Auto-ViTAcc apply PTQ optimization to reduce precision while maintaining accuracy, while HeatViT and EQ-ViT optimize workload distribution through kernel-level techniques and adaptive token pruning. \highlight{On the other hand,} MoE introduces irregular memory access patterns, leading to suboptimal hardware utilization. While ping-pong buffering is a well-established technique to mitigate memory stalls, further optimizations are required to exploit hardware parallelism and memory efficiency.

Additionally, Table~\ref{hwswco} shows that ASIC designs dominate speedup, but FPGA offers better energy efficiency. This suggests that ASIC accelerators provide raw computational power, while FPGA solutions are more power-efficient but slightly lower in absolute speedup.

Most current acceleration techniques for ViTs are highly customized for specialized hardware, such as FPGAs and ASICs, limiting their deployment flexibility \highlight{to other platforms}. Table~\ref{performance_analysis} highlights the trade-offs between accuracy, adaptability, and hardware compatibility in SW-HW co-design techniques for ViTs. \highlight{However, limited support for cross-platform compatibility poses challenges for extending these methods to a wide range of edge devices.} Among the surveyed techniques, PIVOT~\cite{moitra2024pivot} emerges as the most flexible solution, as it maintains high accuracy while being compatible with GPPs, unlike ViTCoD, HeatViT, and VAQF, which are hardware-specialized acceleration techniques. Additionally, Table~\ref{performance_analysis} highlights how different acceleration techniques prioritize computations, offering insights into their adaptability across various deployment scenarios.
\subsection{Performance Analysis for Accelerating Techniques}
This section provides a performance analysis of state-of-the-art (SOTA) accelerating techniques, focusing on key metrics such as power consumption, energy efficiency, resource utilization, and throughput.
\renewcommand{\arraystretch}{1.3}
\begin{table}[]
\centering
\caption{A comparative analysis of hardware performance for various ViT acceleration techniques, deployed on FPGA (\textbf{$\clubsuit$}) and ACAP (\textbf{$\blacklozenge$}) platforms. Performance metrics include Frames Per Second (FPS), Energy Efficiency, and Throughput. The notation \textbf{$\intercal$} represents FPS measured as images per second, while \textbf{$\lozenge$} denotes energy efficiency in GOP/J. Additionally, \textbf{*} indicates power consumption measured in Watt-Seconds (W·S), providing deeper insights into the trade-offs between computation speed and energy usage.}
\label{Hardware_performance}
\resizebox{\columnwidth}{!}{%
\begin{tabular}{c|c|c|c|cccc|c|c|c|c|c}
\hline
\multirow{2}{*}{\textbf{Framework}} &
  \multirow{2}{*}{\textbf{Device}} &
  \multirow{2}{*}{\textbf{Precision}} &
  \multirow{2}{*}{\textbf{\begin{tabular}[c]{@{}c@{}}Frequency\\ (MHZ)\end{tabular}}} &
  \multicolumn{4}{c|}{\textbf{Resource Utilization}} &
  \multirow{2}{*}{\textbf{FPS}} &
  \multirow{2}{*}{\textbf{\begin{tabular}[c]{@{}c@{}}Power\\ (W)\end{tabular}}} &
  \multirow{2}{*}{\textbf{\begin{tabular}[c]{@{}c@{}}Energy \\ (FPS/W)\end{tabular}}} &
  \multirow{2}{*}{\textbf{\begin{tabular}[c]{@{}c@{}}Throughput \\ (GOPs)\end{tabular}}} &
  \multirow{2}{*}{\textbf{\begin{tabular}[c]{@{}c@{}}Speedup \\ ($\uparrow$)\end{tabular}}} \\ \cline{5-8}
 &
   &
   &
   &
  \multicolumn{1}{c|}{\textbf{BRAM}} &
  \multicolumn{1}{c|}{\textbf{DSP}} &
  \multicolumn{1}{c|}{\textbf{KLUT}} &
  \textbf{KFF} &
   &
   &
   &
   &
   \\ \hline
VAQF~\cite{sun2022vaqf} &
  ZCU102\textsuperscript{$\clubsuit$} &
  W1A8 &
  150 &
  \multicolumn{1}{c|}{565.5} &
  \multicolumn{1}{c|}{1564} &
  \multicolumn{1}{c|}{143} &
  110 &
  24.8 &
  8.7 &
  2.85 &
  861.2 &
  - \\ \hline
ViA~\cite{wang2022via} &
  U50\textsuperscript{$\clubsuit$} &
  FP16 &
  300 &
  \multicolumn{1}{c|}{1002} &
  \multicolumn{1}{c|}{2420} &
  \multicolumn{1}{c|}{258} &
  257 &
  - &
  39 &
  7.94\textsuperscript{$\lozenge$} &
  309.6 &
  59.5$\times$ \\ \hline
ViTA ~\cite{nag2023vita} &
  ZC7020\textsuperscript{$\clubsuit$} &
  INT8 &
  150 &
  \multicolumn{1}{c|}{-} &
  \multicolumn{1}{c|}{-} &
  \multicolumn{1}{c|}{-} &
  - &
  8.7 &
  0.88 &
  3.13 &
  - &
  2$\times$ \\ \hline
Auto-ViT-Acc ~\cite{lit2022auto} &
  ZCU102\textsuperscript{$\clubsuit$} &
  W8A8+W4A8 &
  150 &
  \multicolumn{1}{c|}{-} &
  \multicolumn{1}{c|}{1556} &
  \multicolumn{1}{c|}{186} &
  - &
  34.0 &
  9.40 &
  3.66 &
  1181.5 &
  - \\ \hline
HeatViT ~\cite{dong2023heatvit} &
  ZCU102\textsuperscript{$\clubsuit$} &
  INT8 &
  150 &
  \multicolumn{1}{c|}{528.6} &
  \multicolumn{1}{c|}{2066} &
  \multicolumn{1}{c|}{161.4} &
  101.8 &
  11.4 &
  54.8 &
  4.83 &
  - &
  4.89$\times$ \\ \hline
EQ-ViT~\cite{10745859} &
  VCK190\textsuperscript{$\blacklozenge$} &
  W4A8 &
  - &
  \multicolumn{1}{c|}{16\textsuperscript{$\phi$}} &
  \multicolumn{1}{c|}{28\textsuperscript{$\phi$}} &
  \multicolumn{1}{c|}{6.5\textsuperscript{$\phi$}} &
  - &
  10695\textsuperscript{$\intercal$} &
  - &
  224.7 &
  - &
  - \\ \hline
Zhang et al.~\cite{zhang2024109} &
  XCZU9EG\textsuperscript{$\clubsuit$} &
  W8A8 &
  300 &
  \multicolumn{1}{c|}{283} &
  \multicolumn{1}{c|}{2147} &
  \multicolumn{1}{c|}{118} &
  139 &
  36.4 &
   &
  73.56\textsuperscript{$\lozenge$} &
  2330.2 &
  - \\ \hline
M\textsuperscript{3}ViT~\cite{fan2022m3vit} &
  ZCU104\textsuperscript{$\clubsuit$} &
  INT8 &
  300 &
  \multicolumn{1}{c|}{-} &
  \multicolumn{1}{c|}{-} &
  \multicolumn{1}{c|}{-} &
  - &
  84 &
  10 &
  0.690\textsuperscript{*} &
  1217.4 &
  - \\ \hline
ViTCoD~\cite{you2023vitcod} &
  ASIC &
  W8A8 &
  500 &
  \multicolumn{1}{c|}{} &
  \multicolumn{1}{c|}{} &
  \multicolumn{1}{c|}{} &
   &
  - &
  - &
  - &
  - &
  5.6$\times$ \\ \hline
SOLE~\cite{wang2023sole} &
  ASIC &
  INT8 + SOLE &
  - &
  \multicolumn{1}{c|}{-} &
  \multicolumn{1}{c|}{-} &
  \multicolumn{1}{c|}{-} &
  - &
  - &
  - &
  - &
  - &
  57.5$\times$ \\ \hline
ME-ViT~\cite{marino2023me} &
  U200\textsuperscript{$\clubsuit$} &
  W8A8 &
  300 &
  \multicolumn{1}{c|}{288} &
  \multicolumn{1}{c|}{1024} &
  \multicolumn{1}{c|}{192} &
  132 &
  94.13 &
  31.8 &
  4.15 &
  - &
  - \\ \hline
\end{tabular}%
}
\end{table}
\subsubsection{Resource Utilization}\hfill \\ 
Resource utilization measures the hardware efficiency of various ViT acceleration techniques in terms of the use of block RAM (BRAM), digital signal processing units (DSP), Kilo lookup tables (KLUT), and Kilo flip-flops (KFF) from available resources. As illustrated in Table~\ref{Hardware_performance}, different acceleration techniques exhibit varying resource consumption patterns, reflecting their optimization strategies and deployment constraints.

From the observation of Table~\ref{Hardware_performance}, HeatViT (2066 DSPs)~\cite{dong2023heatvit} and Zhang et al. (2147 DSPs)~\cite{zhang2024109} demonstrate the highest DSP \highlight{utlization}, indicating their reliance on intensive parallel processing to accelerate transformer computations\highlight{,} although both used different FPGA variants. In terms of on-chip memory usage, ViA (1002 BRAMs)~\cite{wang2022via} and VAQF (565.5 BRAMs)~\cite{sun2022vaqf} exhibit significant BRAM consumption, emphasizing a design strategy that prioritizes data locality to minimize off-chip memory access latency. In contrast, EQ-ViT (16 BRAMs)~\cite{10745859} utilizes remarkably low BRAM, likely due to its two-level optimization kernels, leveraging both single artificial intelligence engines (AIEs) and AIE array levels. Additionally, Auto-ViT-Acc~\cite{lit2022auto} (186 KLUTs) and HeatViT~\cite{dong2023heatvit} (161.4 KLUTs) indicate that they need to perform significant logical operations to implement their mixed precision quantization and token pruning to deploy on edge.
\subsubsection{Energy Efficiency}\hfill \\
Table~\ref{Hardware_performance} also indicates the energy efficiency of the ViT accelerating techniques. In the current studies, energy efficiency was measured in two ways: using throughput (GOP/J) and using FPS (FPS/watt). We also include the power consumption of the accelerating techniques for better accountability. It is perhaps difficult to conclude about energy efficiency when different edge targets are used for the CV task. \highlight{As illustrated in Table~\ref{Hardware_performance}}, VAQF~\cite{sun2022vaqf}, Auto-ViT-Acc~\cite{lit2022auto}, and HeatViT~\cite{dong2023heatvit} utilized the same FPGA AMD ZCU102 board with the same number of resources and frequency. We can observe that VAQF outperforms Auto-ViT-Acc and HeatViT \highlight{in terms of} energy and power usage. However, we were unable to find any energy comparison from the original paper for ASIC-based accelerating techniques (e.g., ViTCoD, SOLE).
\subsubsection{Throughput}\hfill \\
Table~\ref{Hardware_performance} illustrates the throughput of the accelerating techniques. We observe significant variations in throughput, influenced by factors such as hardware architecture, precision, and optimization strategies. Zhang et al. (2330.2 GOPs)\highlight{~\cite{zhang2024109}} achieve the highest throughput, leveraging an FPGA-based implementation with optimized parallel execution. Auto-ViT-Acc (1181.5 GOPs) and M³ViT (1217.4 GOPs) also report high throughput, suggesting effective hardware utilization. However, several studies, such as ViTCoD, do not report the throughput. Additionally, EQ-ViT and VAQF balance throughput with power efficiency, offering a more energy-efficient alternative.
\subsubsection{Accuracy}\hfill \\ Table~\ref{accuracy_performance} presents a comparative accuracy analysis across various ViT acceleration techniques on different edge platforms. To ensure a fair and structured comparison, we categorize our evaluation into four widely used ViT-based models: DeiT-Base, DeiT-Tiny, ViT-Base, and ViT-Small. While most acceleration techniques focus on classification tasks using the ImageNet-1K dataset~\cite{5206848}.

Among the methods analyzed, Auto-ViT-Acc achieves the highest Top-1 accuracy (81.8\%) on DeiT-Base, significantly surpassing VAQF (77.6\%) for classification tasks, demonstrating its effectiveness in preserving model accuracy while accelerating inference. For DeiT-Tiny, accuracy varies significantly across different methods: HeatViT (72.1\%), EQ-ViT (74.5\%), ViTCoD (70.0\%), and SOLE (71.07\%). Notably, EQ-ViT achieves the highest accuracy among these, highlighting the effectiveness of its attention-based optimizations. However, its energy consumption is significantly higher, indicating a trade-off between accuracy and efficiency.

M\textsuperscript{3}ViT\cite{fan2022m3vit} extends its evaluation to segmentation tasks on PASCAL-ContextNYUD-v2\cite{silberman2012indoor}. For segmentation tasks, M\textsuperscript{3}ViT delivers strong performance on PASCAL-Context (72.8 mIoU) but exhibits a noticeable decline on NYUD-v2 (45.6 mIoU), suggesting that its model compression techniques may be dataset-sensitive.

\renewcommand{\arraystretch}{1.2}
\begin{table}[]
\centering
\caption{Comparison of accuracy across different ViT acceleration techniques using DeiT-B (Base), DeiT-T (Tiny), ViT-B (Base), and ViT-S (Small) as baseline models. Results include Top-1 accuracy on ImageNet-1K and mean Intersection over Union (mIoU) for segmentation benchmarks.}
\label{accuracy_performance}
\small
\resizebox{0.7\columnwidth}{!}{%
\begin{tabular}{c|c|c|cc}
\hline
\multirow{2}{*}{\textbf{Framework}} &
  \multirow{2}{*}{\textbf{Baseline}} &
  \multirow{2}{*}{\textbf{Dataset}} &
  \multicolumn{2}{c}{\textbf{Accuracy}} \\ \cline{4-5} 
                                 &        &                                             & \multicolumn{1}{c|}{\textbf{Top-1(\%)}} & \textbf{mIoU} \\ \hline
VAQF~\cite{sun2022vaqf}          & DeiT-B & \multirow{2}{*}{ImageNet-1K~\cite{5206848}} & \multicolumn{1}{c|}{77.6}               & -             \\
Auto-ViT-Acc ~\cite{lit2022auto} & DeiT-B &                                             & \multicolumn{1}{c|}{81.8}               & -             \\ \hline
HeatViT ~\cite{dong2023heatvit}  & DeiT-T & \multirow{4}{*}{ImageNet-1K~\cite{5206848}} & \multicolumn{1}{c|}{72.1}               & -             \\
EQ-ViT~\cite{10745859}           & DeiT-T &                                             & \multicolumn{1}{c|}{74.5}               & -             \\
ViTCoD~\cite{you2023vitcod}      & DeiT-T &                                             & \multicolumn{1}{c|}{70.0}               &               \\
SOLE~\cite{wang2023sole}         & DeiT-T &                                             & \multicolumn{1}{c|}{71.07}              & -             \\ \hline
Zhang et al.~\cite{zhang2024109} & ViT-B  & ImageNet-1K~\cite{5206848}                  & \multicolumn{1}{c|}{83.1}               & -             \\ \hline
\multirow{2}{*}{M\textsuperscript{3}ViT~\cite{fan2022m3vit}} &
  \multirow{2}{*}{ViT-S} &
  PASCAL-Context~\cite{everingham2010pascal} &
  \multicolumn{1}{c|}{\multirow{2}{*}{-}} &
  72.8 \\ \cline{3-3} \cline{5-5} 
                                 &        & NYUD-v2~\cite{silberman2012indoor}          & \multicolumn{1}{c|}{}                   & 45.6          \\ \hline
\end{tabular}%
}
\vspace{-3mm}
\end{table}

\section{Challenges and Future Directions of ViT on Edge Devices}\label{cha_fu}
ViT models are computation-intensive, and their deployment on resource-constrained edge devices has been a big challenge. However, with the advancement of edge AI, this is now changing, and the efficient and cost-effective implementation of ViT models is possible directly on edge hardware. This increases accessibility for end users and reduces reliance on cloud infrastructure, which lowers latency, improves privacy, and reduces operational costs. However, some areas, such as real-world scenarios and \highlight{SW-HW} co-design still need to be explored for ViT on edge devices. In this section, we will discuss the current challenges and future opportunities of ViT on edge devices.

\subsection{\highlight{Advancement in SW-HW} Co-design}\label{h_fu} The SW-HW co-design can reduce the current dilemma between model and hardware architectures. Additionally, different edge hardware platforms (e.g., CPUs, GPUs, and FPGAs) have varying capabilities in handling precision, memory bandwidth, and computational efficiency. Often, accelerators support a uniform bit-width tensor, and this distinct bit-width precision needs zero padding, incurring inefficient memory usage. It is so hard to optimize the ViT for each type of hardware. Leveraging hardware-aware compression techniques can improve the efficiency of edge deployment. Frameworks such as DNNWeaver~\cite{7783720}, VAQF~\cite{sun2022vaqf}, M$^3$ViT~\cite{fan2022m3vit} have been developed for different hardware platforms like FPGA, GPU accelerators for efficient edge inference. However, most of the current frameworks can not handle the sparsity caused by model compression. Therefore, the advancement of reconfigurability of \highlight{SW-HW} co-design for handling sparsity can be a future problem to solve.

\highlight{Additionally, future work should focus on establishing collaborative SW-HW design pipelines that integrate compression-aware modeling with hardware-specific execution constraints. For example, model compression in software could be dynamically adapted based on hardware feedback or user-specific hardware requirements. This requires tighter integration between models and hardware profiling tools. Further, hardware-agnostic intermediate representations and unified compiler toolchains could help bridge the gap between diverse ViT models and heterogeneous edge devices.
}

\subsection{\highlight{Utilizing} NAS for Inference and Compression Parameter Search}\label{u_nas}
\highlight{NAS is widely used to discover efficient model structures. However, its application to ViTs on edge devices remains limited due to high computational costs and strong coupling to specific hardware platforms.} Frameworks such as HAQ~\cite{wang2019haq} and APQ~\cite{wang2020apq} utilized NAS to automate pruning and quantization strategies through reinforcement learning or evolutionary algorithms. However, these works are highly customized for specific hardware (e.g., HAQ for FPGA). ProxylessNAS~\cite{cai2018proxylessnas} is a notable effort in this direction that introduces hardware-aware search and demonstrates strong real-time latency prediction, but it is limited to CNN architectures. Extending such approaches to ViT and its hierarchical, attention-based structure remains an open challenge.

\highlight{Future research should extend NAS to simultaneously search for model-independent compression parameters, such as quantization bit-widths, pruning ratios, and distillation strategies, based on hardware-aware objectives (e.g., latency, memory usage, energy efficiency). Reinforcement learning can be utilized to efficiently guide the search process based on hardware requirements. Additionally,  exploring NAS to FPGA custom design, which requires deep hardware expertise, might be an interesting and challenging research area for the efficient deployment of ViT.}
\subsection{\highlight{Hardware Accelerators} to Handle Sparsity and Mixed Precision}\label{h_acc} Traditional processors such as GPUs, CPUs, or even FPGAs struggle with executing irregular tensor operations efficiently. Current accelerators are not inherently designed to process sparse tensors, as they often fetch zero values from memory to processing elements (PEs), incurring unnecessary latency and energy overhead. \highlight{Another challenge is} the limited support for mixed-precision execution. Although recent techniques have shown that mixed-precision quantization can preserve ViT accuracy while reducing model size, most edge hardware requires uniform bit-widths across operations~\cite{lin2024awq}, leading to compatibility issues or bit padding \highlight{and making it hard to deploy on edge devices.}

\highlight{To address these limitations, future research should focus on developing sparsity-aware accelerator architectures capable of dynamically skipping zero computations, using methods such as compressed sparse row (CSR) encoding, zero-skipping logic~\cite{vestias2019fast}, or sparsity-driven scheduling techniques~\cite{10122694,10824859}. Similarly, designing bit-level reconfigurable compute units could enable native support for mixed-precision execution without introducing significant hardware redundancy. Thus, specialized techniques are needed to optimize the storage and computation of nonzero values in ViT.}

\highlight{Additionally, emerging math-based accelerating techniques such as tensor-train (TT) decomposition offer a promising direction for accelerating ViT models. TT-decomposition can significantly reduce parameter count and computational complexity by representing high-dimensional weight matrices in a compact, low-rank format. Recent studies, including TT-ViT~\cite{pham2022tt} and TT-MLP~\cite{10032168}, demonstrate its potential for transformer and MLP compression, while hardware-aware implementations~\cite{qu2021hardware} further highlight its relevance for deployment. Although integration into edge-specific ViT deployments remains limited, TT-decomposition represents a promising future research avenue for ViT acceleration, especially on FPGA due to its reconfigurable property.}

\subsection{Automated Edge Aware Model Compression} Most of the current model compression techniques require manual adjustment of hyperparameters such as quantization bit width, pruning ratio, or layer-wise sparsity. Compression hyperparameters must be adjusted automatically or adaptively within the resource budget with minimum degradation of accuracy. From our observation, few works explore adjusting the compression parameters automatically. SparseViT~\cite{chen2023sparsevit} is one of the works that effectively reduced computation by targeting less-important regions with dynamically chosen pruning ratios in the images for the ViT model, achieving significant latency reductions. As a result, developing a hardware-efficient automaticity compression technique can be an interesting research domain in the future. 

Another drawback we observe from section~\ref{vit_quanti} is that most of the current work on ViT is post-training. The most interesting reason, perhaps the interactive nature of the training process, is that implementing compression techniques during training requires cost and time. However, QAT techniques on ViT are promising~\cite{li2023psaq,dong2024packqvit,zhang2023qd,aqvit}, but there is limited work on other compression techniques. Thus, the automated exploration of compression techniques during training can improve hardware realization. However, efficient compression techniques for faster convergence during training with reduced computation need to be explored in the future.
\subsection{Developing Benchmarks} Proper benchmark standards to evaluate the performance of the edge devices are essential. The different stages of the model for deploying edge, including compression and accelerators, require a universal and comprehensive set of metrics to compare different proposed solutions. However, benchmarking datasets and models from the system perspective \highlight{is} limited. For example, most of the compression techniques for ViT were evaluated on the ImageNet-1k~\cite{5206848} dataset for classification tasks and \highlight{the} COCO-2017~\cite{lin2015microsoft} datasets for object detection tasks. However, expanding that knowledge \highlight{to} different real-world application areas is still limited due to the lack of datasets and model benchmarks. Thus, more benchmarking datasets and ViT models for evaluating the proposed system/framework need to be developed for different application areas, such as medical imaging and autonomous driving. In addition, making one compression technique universal for different CV tasks is challenging. Making task-independent universal compression techniques can be an interesting research domain in the future.

\subsection{Real-world Case Studies}\label{real_world} The deployment of ViT on edge devices has gained significant progress in recent years. However, most of the compression techniques, frameworks, and accelerators are limited to evaluation in an academic environment. For instance, there are numerous ViT \highlight{models} on medical imaging datasets for different CV tasks such as image classification~\cite{dai2021transmed,raj2023strokevit,gheflati2022vision}, segmentation~\cite{hatamizadeh2021swin,li2023lvit,heidari2023hiformer,he2023h2former,yang2023ept}, and object detection~\cite{shou2022object,leng2023deep,wittmann2022swinfpn,lin2021aanet}. However, few studies have evaluated compression and accelerator techniques on those  ViT-based medical imaging models. Exploring ViT on edge for medical imaging can be challenging because of the unique nature of the data (3D ultrasounds or MRIs).

Maintaining accuracy, latency, and precision is critical in real-world applications, particularly in critical fields like medical imaging. In medical imaging, a significant challenge \highlight{during compression} of the ViT models is maintaining the spatial resolution and feature details since even small degradations in accuracy will affect the diagnostic outcome significantly~\cite{hou2019high}. Compression techniques can reduce excessive feature abstraction, potentially discarding vital low-level details essential for accurate diagnoses. Furthermore, transfer learning in medical imaging adds another layer of complexity—determining which pre-trained layers to retain or modify without losing critical learned representations is a significant challenge~\cite{peng2022rethinking,vrbanvcivc2020transfer}. Therefore, it is an open research direction to achieve a balance between model efficiency and diagnostic reliability for ViT models on edge for real-world scenarios such as healthcare applications.

\subsection{Seamless Model-to-Edge Integration} The conversion from a trained model into a hardware-compatible version for inference requires extensive time, cost, and, most importantly, manual \highlight{labor} in each step. Additionally, there is a high knowledge gap between the research community. For example, training new models requires extensive software knowledge, while compressing and developing accelerating strategies require deep hardware knowledge. It is challenging to find an expert in both directions in the research community. These difficulties create a significant research gap in developing tools that automatically map the models \highlight{to} hardware. However, FPGA can overcome some limitations with the ability to reconfigure new operations and modules. However, the available tools are insufficient for the automatic mapping of models and are more limited for ViT.

The current deep learning frameworks for edge deployment help researchers quickly prototype the models to deploy on edge. However, it lacks support with the rapid growth of different model architectures. Additionally, most of the current frameworks are evaluated for CNN models, while those frameworks are still in the experimental phase for ViT. For instance, Xilinx provides quantization support through the FINN-R framework for inference realization on FPGA~\cite{blott2018finn}, limiting only standard techniques for ViT. Therefore, the automatic mapping from model to edge, precisely a one-click solution for deployment on edge based on the budget or auto-generated compression techniques, can be an interesting domain in the future.
\subsection{Robustness to Diverse Data Modality}In real-world scenarios, data sources change from sensor to sensor or vendor to vendor. For example, medical imaging includes X-rays and other modalities like magnetic resonance imaging (MRI), computed tomography (CT) scans, and ultrasounds. Each modality has its characteristics, and a compression technique effective for one might not be for another. So, using a generic model compression technique for all modalities is always tricky. Such heterogeneity may cause inconsistency in data, imposing a challenge on edge performance. Federated learning, multi-modal fusion, and adaptive data calibration techniques can be promising research directions to mitigate the data inconsistency problem at the edge device.

\section{Conclusion}\label{conclusion} 

\highlight{In this survey, we comprehensively analyze model compression and hardware acceleration techniques on ViT models, spanning techniques ranging from pruning, quantization, and KD, software tools, to SW-HW co-design.} Our analysis indicates that while SOTA ViT compression and acceleration techniques effectively reduce computational overhead and improve inference speed, challenges such as hardware adaptability, memory bottlenecks, and optimal compression strategies remain unexplored. Additionally, \highlight{we} discuss the potential future directions, such as utilizing NAS to find hardware-aware optimization parameters, sparsity-aware accelerators, and efficient cross-platform SW-HW co-design frameworks. 

\highlight{Overall, optimizing ViT for edge devices is an evolving field, marked by both challenges and opportunities. This survey provides a comprehensive overview of the current landscape and aims to inspire continued innovation in this crucial area.}

\section{Acknowledgement}
This work was partly supported by the U.S. National Science
Foundation under Grants CNS-2245729 and Michigan Space
Grant Consortium 80NSSC20M0124. \highlight{The authors would also like to thank Yunge Li for the continuous discussions and insightful feedback that helped shape the direction of this study.}
\bibliographystyle{elsarticle-num} 
\bibliography{reference}






\end{document}